\documentclass[10pt,twocolumn,letterpaper]{article}
\usepackage[pagenumbers]{cvpr}
\usepackage[T1]{fontenc}

\usepackage{algorithm}
\usepackage{algpseudocodex}
\usepackage{amsfonts}
\usepackage{multirow}
\usepackage{placeins}
\usepackage{stfloats}
\newcommand{\vecbold}[1]{\boldsymbol{#1}}
\makeatletter
\algrenewcommand{\LComment}[1]{\Statex \hskip\ALG@thistlm \(\triangleright\) #1}
\makeatother

\definecolor{cvprblue}{rgb}{0.21,0.49,0.74}
\usepackage[breaklinks,colorlinks,allcolors=cvprblue]{hyperref}

\def\confName{CVPR}
\def\confYear{2026}

\title{XMix: Combating Extremely Noisy Labels via Local Smoothness in Self-Supervised Feature Space}
\author{
Chengqi Li \quad Yangdi Lu \quad Zhihao Shi \quad Wenbo He\\
Department of Computing and Software, McMaster University\\
Hamilton, ON, Canada
\and
Chamseddine Talhi \quad Nadjia Kara\\
Département de génie logiciel et TI, École de Technologie Supérieure\\
Montreal, QC, Canada
}

\begin{document}

\maketitle

\begin{abstract}
Supervised deep learning models rely on large, accurately labeled datasets, yet noisy annotations are often unavoidable and can severely degrade performance under high noise levels. Recent state-of-the-art methods tackle this by using sample selection strategies that exploit the memorization effect to filter out clean data for semi-supervised learning. However, these methods struggle with extreme noise, class imbalance, and require careful tuning or prior noise knowledge. To address these limitations, we propose XMix, a novel framework that leverages local smoothness in the self-supervised feature space to systematically enhance all stages of the sample selection process, without dependence on potentially corrupted labels. First, XMix estimates the noise rate using maximum likelihood among self-supervised feature neighbors. Second, these neighbors then help identify additional clean samples and ensure balanced selection across classes during sample selection. Finally, in the semi-supervised learning phase, XMix uses neighboring samples to generate more reliable pseudo-labels. Our empirical results show that XMix substantially outperforms existing methods in extremely noisy environments and maintains superior performance in standard LNL benchmarks.
\end{abstract}

\noindent\textbf{Code:} \url{https://github.com/steveli88/XMix}

\section{Introduction}

The availability of substantial amounts of precisely annotated data is crucial for the performance of supervised deep learning models \cite{deng_imagenet_2009}. However, annotation errors are unavoidable due to the intrinsic limitations of both human annotators \cite{yan_learning_2014} and automatic annotation systems \cite{forsyth_new_2008}. Learning with Noisy Labels (LNL), initially formalized by Natarajan et al. \cite{natarajan_learning_2013}, has emerged as an active area focused on mitigating the negative effects of corrupted labels on deep image classification.

\begin{figure}[t]
\centering
\begin{minipage}[htb]{0.49\linewidth}
    \centering
    {\includegraphics[width=\linewidth]{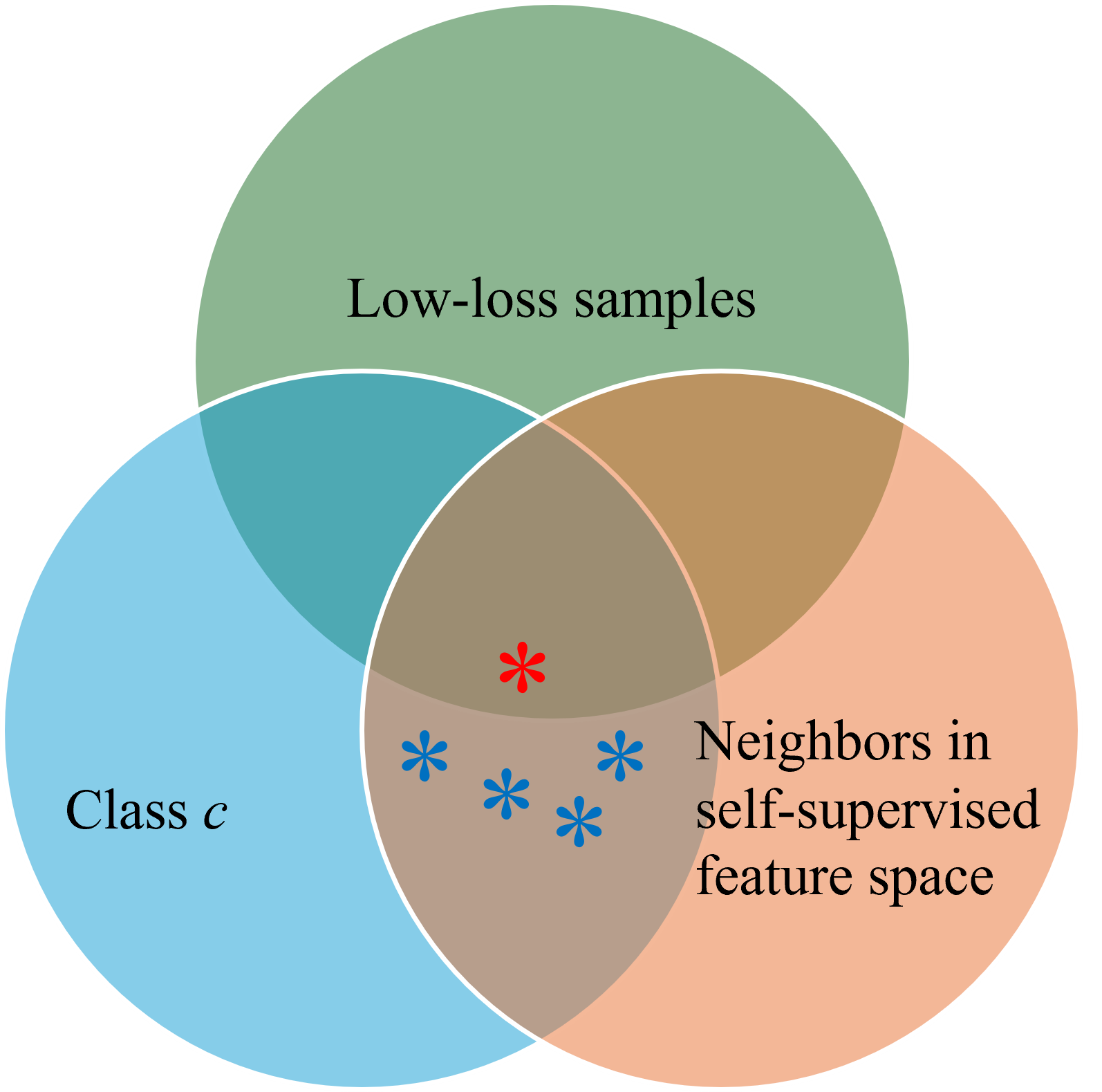}}
    \centerline{(a)}
\end{minipage}
\begin{minipage}[htb]{0.49\linewidth}
    \centering
    {\includegraphics[width=\linewidth]{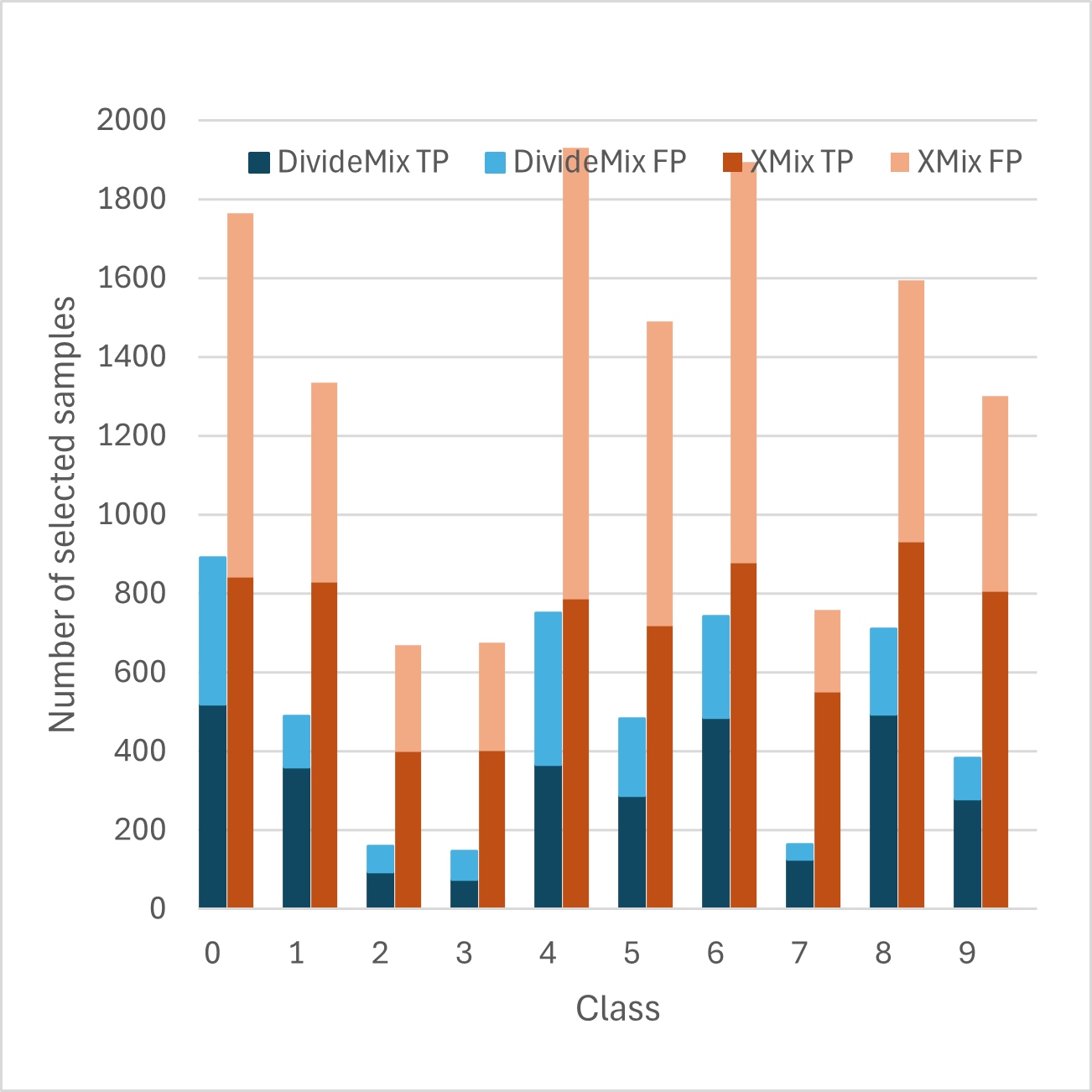}}
    \centerline{(b)}
\end{minipage}
\caption{(a) Expanding clean sample selection via local smoothness in the self-supervised feature space. {\color{red}The red star} denotes a clean sample initially selected by the low-loss criterion, while {\color{blue}blue stars} indicate additional clean samples identified. (b) Selected samples for each class (DivideMix vs XMix) on the CIFAR-10 dataset with $90\%$ symmetric label noise. TP and FP represent true and false clean samples, displayed in light and dark shades, respectively.}
\label{fig:xmix}
\end{figure}

\begin{table}[t]
\centering
\resizebox{\linewidth}{!}{
\begin{tabular}{llllll}
\toprule
Dataset                          & \multicolumn{3}{l}{CIFAR-10} & \multicolumn{2}{l}{CIFAR-100} \\
\midrule
Method\textbackslash{}Noise rate & 92\%   & 95\%   & 98\%   & 92\%     & 95\%    \\
\midrule
DivideMix \cite{li_dividemix_2020}                         & 57.6\%   & 51.3\%  & 17.2\%  & 12.6\%        & 7.7\%   \\
UNICON \cite{karim_unicon_2022}                           & 87.6\%   & 80.8\%  & 50.6\%  & 32.1\%        & 19.1\%  \\
CrossSplit \cite{kim_crosssplit_2023}                       & 80.4\%   & 72.0\%  & 31.3\%  & 46.3\%        & 30.0\%  \\
ProMix \cite{xiao_promix_2023}                           & 79.7\%   & 55.1\%  & 42.0\%  & 29.9\%        & 8.5\%   \\
PLReMix \cite{liu_rlremix_2025}                           & 85.1\%   & 79.3\%  & 40.4\%  & 40.7\%        & 25.7\% \\
\midrule
XMix (DivideMix+) & 90.4\%   & 85.2\%  & 41.4\%  & 28.5\%        & 17.3\%   \\
XMix (ProMix+) & \textbf{93.9\%}   & \textbf{92.7\%}  & \textbf{56.5\%}  & \textbf{64.1\%}        & \textbf{31.4\%} \\
\bottomrule
\end{tabular}
}
\caption{Classification accuracies on CIFAR-10 and CIFAR-100 under extreme label noise. The "+" sign following a method name means the method is enhanced with both balanced \& expanded selection and nearest neighbor pseudo-labeling. The best results are in \textbf{bold}. }
\label{tab:cifar_extreme_noise}
\end{table}

Current LNL methods broadly fall into two categories: loss correction \cite{patrini_making_2017,hendrycks_using_2018,xia_are_2019,yao_dual_2020,li_provably_2021} and sample selection \cite{han_co-teaching_2018,wei_combating_2020,li_dividemix_2020,ortego_multi-objective_2021,xiao_promix_2023}. Loss correction methods estimate a noise transition matrix that adjusts the loss function to account for mislabeled data. While conceptually appealing, these methods become unreliable when facing datasets with many classes or high noise levels, where estimating an accurate transition matrix is particularly challenging. In contrast, sample selection methods use a low-loss criterion \cite{han_co-teaching_2018} based on the memorization effect \cite{zhang_understanding_2017}, treating low-loss samples as labeled and the rest as unlabeled during the subsequent semi-supervised learning phase \cite{berthelot_mixmatch_2019,chen_softmatch_2023}. Despite their empirical success and flexibility, existing sample selection approaches face notable limitations. They often struggle in scenarios with severe label noise, fail to identify a sufficiently large clean subset, and can introduce class imbalance within the selected samples. Moreover, these methods typically require careful hyperparameter tuning or prior knowledge of the underlying noise distribution to remain effective.

Typically, sample selection methods \cite{li_dividemix_2020, xiao_promix_2023} start with a warm-up phase where a neural network is trained on images and their noisy labels for a few epochs. Afterward, samples whose predictions align closely with their labels are treated as clean, leveraging the memorization effect \cite{zhang_understanding_2017}. However, these methods rely on the joint distribution of images and noisy labels, often neglecting rich, noise-agnostic information intrinsic to the images themselves. For instance, objects sharing similar colors or structures are more likely to belong to the same class, providing valuable prior information that could further enhance sample selection and improve resilience to severe noise.

Building on the smoothness assumption \cite{van_engelen_survey_2020} that samples close to each other in feature space are likely to share the same label, we first estimate the class-wise noise rate within these local feature neighborhoods using maximum likelihood estimation. This estimated noise rate guides hyperparameter selection for the baseline sample selection model, making it adaptable to varying noise levels without requiring prior knowledge of the noise. To further improve clean sample identification, we expand the clean set by including samples that share the same observed label as low-loss samples within the same neighbor cluster. This approach not only introduces additional clean samples under high noise when semi-supervised learning methods struggle to converge with insufficient clean samples, but also helps balance the selected samples across classes by recovering more underrepresented examples from a broader neighborhood. Finally, during the semi-supervised learning phase, we leverage these feature neighbors to generate more accurate pseudo-labels for the unlabeled data, further improving overall performance.

In summary, our major contributions are as follows:

\begin{itemize}
\item \textbf{Class-wise noise rate estimation:} XMix estimates the class-wise noise rate within feature-space neighborhoods using maximum likelihood, enabling the model to automatically adjust hyperparameters without prior knowledge of the noise level.

\item \textbf{Balanced and expanded sample selection:} By leveraging feature-space neighbors, XMix expands the clean sample set beyond standard low-loss selection, recovering more clean samples under extreme noise and mitigating class imbalance in the selected data.

\item \textbf{Nearest neighbor pseudo-labeling:} XMix uses neighbor information to generate more reliable pseudo-labels during semi-supervised learning, improving robustness in highly noisy scenarios.

\item \textbf{Empirical validation:} Experiments show that XMix consistently boosts the performance of sample selection baselines across multiple LNL benchmarks, achieving particularly large improvements under severe label noise.
\end{itemize}

\section{Related Work}

Early research in LNL focused on achieving an unbiased estimation of clean sample losses using loss correction \cite{patrini_making_2017,xu_l_dmi_2019,ma_normalized_2020}. These methods include robust variants of cross-entropy (CE) loss, such as information-theoretic loss \cite{xu_l_dmi_2019}, and loss normalization \cite{ma_normalized_2020}. Another critical aspect of loss correction involves estimating the noise transition matrix \cite{patrini_making_2017,xia_are_2019,li_provably_2021}, which allows for explicit correction of the loss function using the inverted matrix. While these loss correction methods are effective under moderate noise levels, their reliability declines as the noise level increases, because estimations of clean sample losses become less accurate.

Recent sample selection methods \cite{han_co-teaching_2018,wei_combating_2020,li_dividemix_2020,xiao_promix_2023} focus on isolating clean samples. Early methods \cite{han_co-teaching_2018,wei_combating_2020} established a two-network framework, wherein one network is trained on clean samples selected by the other network according to the small loss criterion. Other studies \cite{li_dividemix_2020, ortego_multi-objective_2021, xiao_promix_2023} further incorporate semi-supervised learning techniques \cite{sohn_fixmatch_2020, chen_softmatch_2023}, treating the selected clean samples as labeled data and the remaining ones as unlabeled data. Hybrid sample selection methods, like ProMix \cite{xiao_promix_2023}, UNICON \cite{karim_unicon_2022}, and CrossSplit \cite{kim_crosssplit_2023}, progressively refine the process of sample selection by incorporating pseudo-labeling and robust loss functions, thereby boosting the overall performance of the models. Sample selection methods have emerged as a dominant paradigm for LNL due to their robustness and flexibility. However, they require prior knowledge of the noise level for tuning, often lead to class-imbalanced selection as some classes are easier to learn, and struggle to provide sufficient clean samples for semi-supervised learning under severe noise.

\begin{table*}[th!]
\centering
\resizebox{\linewidth}{!}{
\begin{tabular}{lllllllllll}
\toprule
Dataset& \multicolumn{5}{c}{CIFAR-10} & \multicolumn{5}{c}{CIFAR-100}     \\
\midrule
Method\textbackslash{}Noise rate & 20\%   & 50\%   & 80\%   & 90\%   & 40\% Asym. & 20\%   & 50\%   & 80\%   & 90\% & 40\% Asym.  \\
\midrule
DivideMix \cite{li_dividemix_2020} & 96.1\% & 94.6\% & 93.2\% & 76.0\% & 91.7\%     & 77.3\% & 74.6\% & 60.2\% & 31.5\% & 55.1\% \\
SELC \cite{lu_selc_2022} & 95.0\% & -      & 78.6\% & -      & 92.9\%     & 76.4\% & -      & 64.5\% & -  & 73.6\%    \\
UNICON \cite{karim_unicon_2022} & 96.0\% & 95.6\% & 93.9\% & 90.8\% & 94.1\%     & 78.9\% & 77.6\% & 63.9\% & 44.8\% & 74.8\% \\
CrossSplit \cite{kim_crosssplit_2023} & 96.9\% & 96.3\% & 95.4\% & 91.3\% & 96.0\%     & 79.9\% & 75.7\% & 64.6\% & 52.4\% & \textbf{76.8\%} \\
RankMatch \cite{zhang_rankmatch_2023} & 96.5\% & 95.6\% & 94.5\% & 92.6\% & 94.7\%     & 79.5\% & 77.9\% & 67.6\% & 50.6\% & -\\
ProMix \cite{xiao_promix_2023} & 97.7\% & \textbf{97.4\%} & 95.5\% & 93.4\% & \textbf{96.6\%}     & \textbf{82.6\%} & 80.1\% & 69.4\% & 42.9\% & 76.2\% \\
CLIPCleaner \cite{feng_clipcleaner_2024} & 95.9\% & 95.7\% & 95.0\% & 94.2\% & 94.9\%     & 78.2\% & 78.2\% & 69.7\% & 63.1\% & -\\
L2B \cite{zhou_l2b_2024} & 96.7\% & 95.6\% & 94.8\% & 94.4\% & 94.0\%     & 80.1\% & 78.1\% & 69.6\% & 60.7\% & -\\
PLReMix \cite{liu_rlremix_2025} & 96.6\% & 95.7\% & 95.1\% & 92.7\% & 95.1\%     & 78.0\% & 77.8\% & 68.8\% & 50.2\% & 64.9\%\\
\midrule
XMix (DivideMix+) & 96.4\% & 96.1\% & 94.4\% & 91.2\% & 94.6\%     & 79.6\% & 76.1\% & 60.9\% & 38.8\% & 64.2\% \\
XMix (ProMix+) & \textbf{97.8\%}  & \textbf{97.4\%}  & \textbf{96.3\%} & \textbf{95.9\%} & \textbf{96.6\%} & \textbf{82.6\%} & \textbf{80.3\%} & \textbf{75.2\%} & \textbf{68.0\%} & 76.5\%\\
\bottomrule
\end{tabular}
}
\caption{Classification accuracies on CIFAR-10 and CIFAR-100 with synthetic label noise benchmarks. The "+" sign following a method name means the method is enhanced with both balanced \& expanded selection and nearest neighbor pseudo-labeling. The best results are in \textbf{bold}.}
\label{tab:cifar_syn_noise}
\end{table*}

\section{Background}
\label{section: background}
Consider a supervised multi-class classification task with $C$ classes. Suppose a clean training dataset $\mathbb{D}$ contains $N$ samples, $\mathbb{D}=\{ (\vecbold{x}_i, \vecbold{y}_i) \}_{i=1}^N$, where $\vecbold{x}_i \in \mathbb{R}^d$ is an input image or feature, $\vecbold{y}_i \in [0,1]^C$ is the corresponding ground-truth label represented as a one-hot vector, and $N$ is the total number of training samples.
For supervised training with ground-truth labels, we minimize cross-entropy (CE) loss, $\mathcal{L}_{CE}$, over the entire training set $\mathbb{D}$,
\begin{equation}
\mathcal{L}_{CE} = - \frac{1}{N}\sum_{i=1}^{N}{{\vecbold{y}_i}^T \log(\hat{\vecbold{y}}_i)}, \quad \hat{\vecbold{y}}_i = f_{\theta}(\vecbold{x}_i).
\end{equation}
Here, $\hat{\vecbold{y}}_i$ is the model prediction for input $\vecbold{x}_i$ with model parameters $\theta$, and each logit of $\hat{\vecbold{y}}_i$ represents the predicted probability for each class.

However, annotation errors are inevitable, so in practice we often train on a noisy dataset $\tilde{\mathbb{D}}=\{(\vecbold{x}_i, \tilde{\vecbold{y}}_i)\}_{i=1}^N$, where the observed labels $\tilde{\vecbold{y}}_i$ may not match the true labels. Despite this, our goal is to develop a model that learns effectively from noisy samples and generalizes well to clean data. In our study, we consider both synthetic \cite{krizhevsky_learning_2009, wei_learning_2021} and real-world noisy datasets \cite{li_webvision_2017, song_selfie_2019, wei_learning_2021}. Synthetic noise includes symmetric label noise (randomly flipping a label to any other class) and asymmetric label noise (flipping labels only between easily confused classes). Real-world noise, in contrast, comes from actual annotation errors made by humans or automated systems.

\section{Proposed Method}
\label{section: method}
This study aims to address key limitations of existing sample selection methods for learning with noisy labels, such as their reliance on prior knowledge of the noise level, class imbalance in the selected clean subset, and poor performance under severe label noise. While current methods focus on extracting as much useful information as possible from the noisy dataset through sample selection and semi-supervised learning, they often overlook a valuable source of reliable information already present: the input images themselves. Shared visual patterns (such as textures, shapes, or color distributions) and similarities in feature space can provide strong cues that certain images are likely to belong to the same class, also known as the smoothness assumption \cite{van_engelen_survey_2020}.

Motivated by this, we propose using a pretrained self-supervised DINOv2 encoder $g_{\phi}(\cdot)$ \cite{oquab_dinov2_2023}, fine-tuned on the target dataset, to extract features from all images. For each image $\vecbold{x}_i$, we then include its $k$ nearest neighbors in the feature space to form a set of samples likely belonging to the same class.
\begin{equation}
\mathbb{G}_i^k = \big\{ (\vecbold{x}_j, \tilde{\vecbold{y}}_j) \;|\; j \in \text{Smallest-}k \big( \| \vecbold{v}_i - \vecbold{v}_j \|_2 \big) \big\}
\end{equation}
Here, $\vecbold{v}_i = g_{\phi}(\vecbold{x}_i)$ and $\vecbold{v}_j = g_{\phi}(\vecbold{x}_j)$ are encoded features. Finally, this neighborhood information is integrated into different phases of the sample selection baseline, as detailed in Algorithm \ref{alg:xmix}.

\begin{algorithm}[tb]
    \caption{Training process of XMix.}
    \label{alg:xmix}
    \textbf{Input}: Training set $\tilde{\mathbb{D}}=\{(\vecbold{x}_i, \tilde{\vecbold{y}}_i)\}_{i=1}^N$, Pretrained self-supervised encoder $g_{\phi}$, Classification model $f_{\theta}$\\
    \textbf{Parameter}: Number of samples $N$, Number of classes $C$, Number of neighbors $k$ \\
    \textbf{Output}: Trained model $f_{\theta^*}$
    \begin{algorithmic}[1] 
        \BeginBox[fill=pink]
        \LComment{Nearest Neighbors in Self-Supervised Feature Space}
        \State $\vecbold{v}_i = g_{\phi}(\vecbold{x}_i), i \in \{1, ..., N\}$
        \State Obtain $\mathbb{G}_i^k$ and $\mathbb{G}_i^{2k}$ according to Eq.(2), $i \in \{1, ..., N\}$
        \LComment{Class-Wise Noise Rate Estimation}
        \State Estimate $\hat\eta_c$ according to Eq.(4)-(5), $c \in \{1, ..., C\}$
        \EndBox
        \State $f_{\theta} = \text{WarmUp}(\tilde{\mathbb{D}}, f_{\theta})$
        \For {$epoch=1,2,\ldots,$}
         \State $\ell_i = \|f_{\theta}(\vecbold{x}_i)-\tilde{\vecbold{y}}_i\|, i \in \{1, ..., N\}$
         \For {$j=1,2,\ldots,C$}
             \State Fit GMM on losses of class $c$
             \State Given per-class threshold $\tau(\hat\eta_c)$, find per-class clean subset $\mathbb{D}_{clean}^c$ according to Eq.(6)
         \EndFor
         \State $\mathbb{D}_{clean} = \cup^C_{c=1} \mathbb{D}_{clean}^c $
         \BeginBox[fill=pink]
         \LComment{Balanced and Expanded Sample Selection}
         \For {$(\vecbold{x}_i, \tilde{\vecbold{y}}_i) \in \mathbb{D}_{clean}$}
             \State Find additional clean sample $\mathbb{D}_{clean}^+$ in $\mathbb{G}_i^{k}$   (or $\mathbb{G}_i^{2k}$ if $\tilde{\vecbold{y}}_i$ is least represented) according to Eq.(7)
            \State $\mathbb{D}_{clean} = \mathbb{D}_{clean} \cup \mathbb{D}_{clean}^+ $
         \EndFor
         \State $\mathbb{D}_{noisy} = \tilde{\mathbb{D}} \backslash \mathbb{D}_{clean}$
         \LComment{Nearest Neighbor Pseudo-Labeling}
         \State Generate pseudo-labeled dataset $\mathbb{D}_{noisy}^*$ from $\mathbb{D}_{clean}$ and $\mathbb{G}_i^{k}$ according to Eq.(9)-(11)
         \EndBox
         \State $f_{\theta} = \text{MixMatch}(\mathbb{D}_{clean}, \mathbb{D}_{noisy}^*, f_{\theta})$
        \EndFor
        \State \textbf{return} $f_{\theta^*}$
    \end{algorithmic}
\end{algorithm}

\subsection{Class-Wise Noise Rate Estimation}
For simplicity, we assume a classification dataset with $C$ classes has a symmetric label noise at level $\eta$. Given a sample $(\vecbold{x}_j, \tilde{\vecbold{y}}_j)$, its $k$ nearest neighbors in the feature space $(\vecbold{x}_j, \tilde{\vecbold{y}}_j) \in \mathbb{G}_i^k$ should share the same label according to the smoothness assumption \cite{van_engelen_survey_2020}. Assuming the observed label of the given sample is correct, the probability that the observed label matches within this local neighborhood is given by:
\begin{equation}
\Pr(\tilde{\vecbold{y}}_j \mid \tilde{\vecbold{y}}_i) =
\begin{cases}
1-\frac{C-1}{C}\eta, & \text{if } \tilde{\vecbold{y}}_j = \tilde{\vecbold{y}}_i,\\
\frac{C-1}{C}\eta, & \text{if } \tilde{\vecbold{y}}_j \neq \tilde{\vecbold{y}}_i.
\end{cases}
\end{equation}
This can be modeled as a Bernoulli trial where a ``success'' occurs if a neighbor's observed label matches that of the given sample, with success probability $q = 1 - \frac{C-1}{C}\eta$. Using the maximum likelihood principle, the empirical success probability is estimated by averaging over all neighbors:
\begin{equation}
\hat{q}_i = \frac{\sum  \textbf{I}(\tilde{\vecbold{y}}_j = \tilde{\vecbold{y}}_i) + 1
}{k + 1}, \quad \forall\, (\vecbold{x}_j, \tilde{\vecbold{y}}_j) \in \mathbb{G}_i^k
\end{equation}
\begin{equation}
\hat\eta = \frac{C(1 - \hat{q})}{C-1}, \quad \hat{q} = \frac{1}{N} \sum_{i=1}^N \hat{q}_i
\end{equation}
However, this estimate is inherently biased because the smoothness assumption holds for true labels, while the observed labels may already be noisy. The probability that the observed label matches the true label is itself $1 - \frac{C-1}{C}\eta$, so the correct success rate for the Bernoulli trial should be on the order of $\mathcal{O}(\eta^2)$, as both the anchor sample and its neighbors must have correct observed labels. In our class-wise noise rate estimation, we simulate multiple known noise levels on the CIFAR-10 dataset and compute the empirical label matching rates among each sample's nearest neighbors in feature space. These observed rates are then used to fit a quadratic correction model that adjusts for the bias. Once calibrated, this model serves as a coarse noise estimator for other datasets. Because it is calibrated on a specific synthetic setting and the underlying Bernoulli assumption is itself only approximate, the resulting estimate should not be interpreted as a precise recovery of the true noise rate. Instead, it is only meant to be accurate enough to place a dataset into one of a handful of noise regimes (e.g., low, moderate, high, or extreme), which is exactly the level of granularity the baseline sample selection methods need to pick their corresponding hyperparameter configuration (selection ratio and threshold schedule), without requiring prior ground-truth noise information.

\noindent \textbf{Remark.} For class-wise noise rate estimation, we apply the same procedure within each class according to the noisy labels. The class-wise noise rate estimation calibrated under synthetic noise also proves effective at identifying the appropriate hyperparameter regime on real-world datasets, despite not recovering their exact noise levels.

\subsection{Balanced and Expanded Sample Selection in Self-Supervised Feature Neighborhood}
Following DivideMix \cite{li_dividemix_2020}, we first perform supervised learning on noisy labels to activate the memorization effect of the classification network $f_{\theta}(\cdot)$. Subsequently, predictions are generated for each sample, and the discrepancies between predictions and given labels are measured. These discrepancies are then modeled using a two-component Gaussian Mixture Model (GMM) to partition the data into a clean subset and a noisy subset.
\begin{equation}
\mathbb{D}_{clean}
= \big\{ (\vecbold{x}_i, \tilde{\vecbold{y}}_i)
\;\big|\; \text{GMM}(\text{clean} \mid \ell_i) > \tau \big\}
\end{equation}
Here, $\ell_i=\|f_{\theta}(\vecbold{x}_i)-\tilde{\vecbold{y}}_i\|$ represents the loss for sample $i$, $\text{GMM}(\text{clean} \mid \ell_i)$ denotes the posterior probability that sample $(\vecbold{x}_i, \tilde{\vecbold{y}}_i)$ is clean, given its loss $\ell_i$, as estimated by the GMM. A sample is classified as clean if this probability exceeds a predefined threshold $\tau(\hat\eta)$. This partition is class-wise imbalanced and will become less effective in selecting the correct clean samples when label noise increases, as shown in Figure \ref{fig:selected_samples_noise}.

Therefore, we propose to leverage information from self-supervised feature neighbors, following the smoothness assumption that feature-similar samples are likely to belong to the same class. This allows us to identify additional clean samples within the neighborhood of an already selected clean sample. Given a clean sample $(\vecbold{x}_i, \tilde{\vecbold{y}}_i)$ and its feature neighbors $\mathbb{G}_i^k$, the expansion of the clean sample subset can be formulated as follows.
\begin{equation}
\begin{split}
\mathbb{D}_{clean}^+ = \big\{ (\vecbold{x}_j, \tilde{\vecbold{y}}_j) \;|\; &(\vecbold{x}_i, \tilde{\vecbold{y}}_i) \in \mathbb{D}_{clean} \\ \wedge &(\vecbold{x}_j, \tilde{\vecbold{y}}_j) \in \mathbb{G}_i^k \wedge \tilde{\vecbold{y}}_j = \tilde{\vecbold{y}}_i \big\}
\end{split}
\end{equation}
\begin{equation}
\mathbb{D}_{clean} = \mathbb{D}_{clean} \cup \mathbb{D}_{clean}^+
\end{equation}
This expansion helps address the challenge of selecting a sufficient number of clean samples in the presence of heavy label noise. To further balance the number of selected samples across classes, we simply apply a different rule for the least represented class by considering $2k$ neighbors instead of $k$ neighbors used for other classes. As training progresses, this strategy gradually helps the selected samples become more class-balanced. It is worth noting that any sample not included in the clean subset is treated as a noisy sample, defined as $\mathbb{D}_{noisy} = \tilde{\mathbb{D}} \backslash \mathbb{D}_{clean}$. Additionally, we employ two networks in a co-training setup, where each network selects clean samples for the other. The expansion and balancing of clean samples are applied symmetrically to both branches.

\subsection{Semi-Supervised Learning with Nearest Neighbor Pseudo-Labeling}
The selected clean subset is treated as labeled data, while the remaining noisy samples are treated as unlabeled. We first train the model on the clean subset and then generate pseudo-labels for the noisy subset by aggregating predictions from their feature-space neighbors. Formally, the pseudo-labeled subset is defined as:
\begin{equation}
    \mathbb{D}_{unlabeled}^*=\mathbb{D}_{noisy}^*=\{(\vecbold{x}_j, \overline{\vecbold{y}}_j)\}_{j=1}^M
\end{equation}
\begin{equation}
    \overline{\vecbold{y}}_j = \text{SoftMax}\bigg( \sum_{n=1}^{k} w_n \cdot f_{\theta}(\vecbold{x}_n) \bigg), \; \forall \vecbold{x}_n \in \mathbb{G}_j^k
\end{equation}
where the weight $w_n$ is defined as:
\begin{equation}
    w_n =
\begin{cases}
2 & \text{if } \vecbold{x}_n \in \mathbb{D}_{\text{clean}}, \\
1 & \text{if } \vecbold{x}_n \in \mathbb{D}_{\text{noisy}}.
\end{cases}
\end{equation}
Here, the model $f_{\theta}$, trained on the labeled data, generates predictions for each unlabeled input's feature-space neighbors. XMix then computes the pseudo-label by taking the softmax of the weighted sum of these predictions, assigning higher weight to predictions from identified clean samples.

The model is then retrained on both the labeled and pseudo-labeled data, following MixMatch \cite{berthelot_mixmatch_2019} to blend labeled examples with unlabeled samples and their predicted pseudo-labels.
\begin{align}
\lambda &\sim \text{Beta}(\alpha, \alpha) \\
\lambda' &= \max(\lambda, 1 - \lambda) \\
\vecbold{x'} &= \lambda' \vecbold{x_1} + (1 - \lambda') \vecbold{x_2} \\
\vecbold{y'} &= \lambda' \vecbold{y_1} + (1 - \lambda') \vecbold{y_2}
\end{align}
where image-label pairs $(\vecbold{x_1},\vecbold{y_1})$ and $(\vecbold{x_2},\vecbold{y_2})$ are randomly sampled from the combined set $\mathbb{D}_{labeled} \cup \mathbb{D}_{unlabeled}^*$.

\subsection{Discussion}
\begin{figure*}[th!]
\centering
\begin{minipage}[htb]{0.23\linewidth}
    \centering
    {\includegraphics[width=\linewidth]{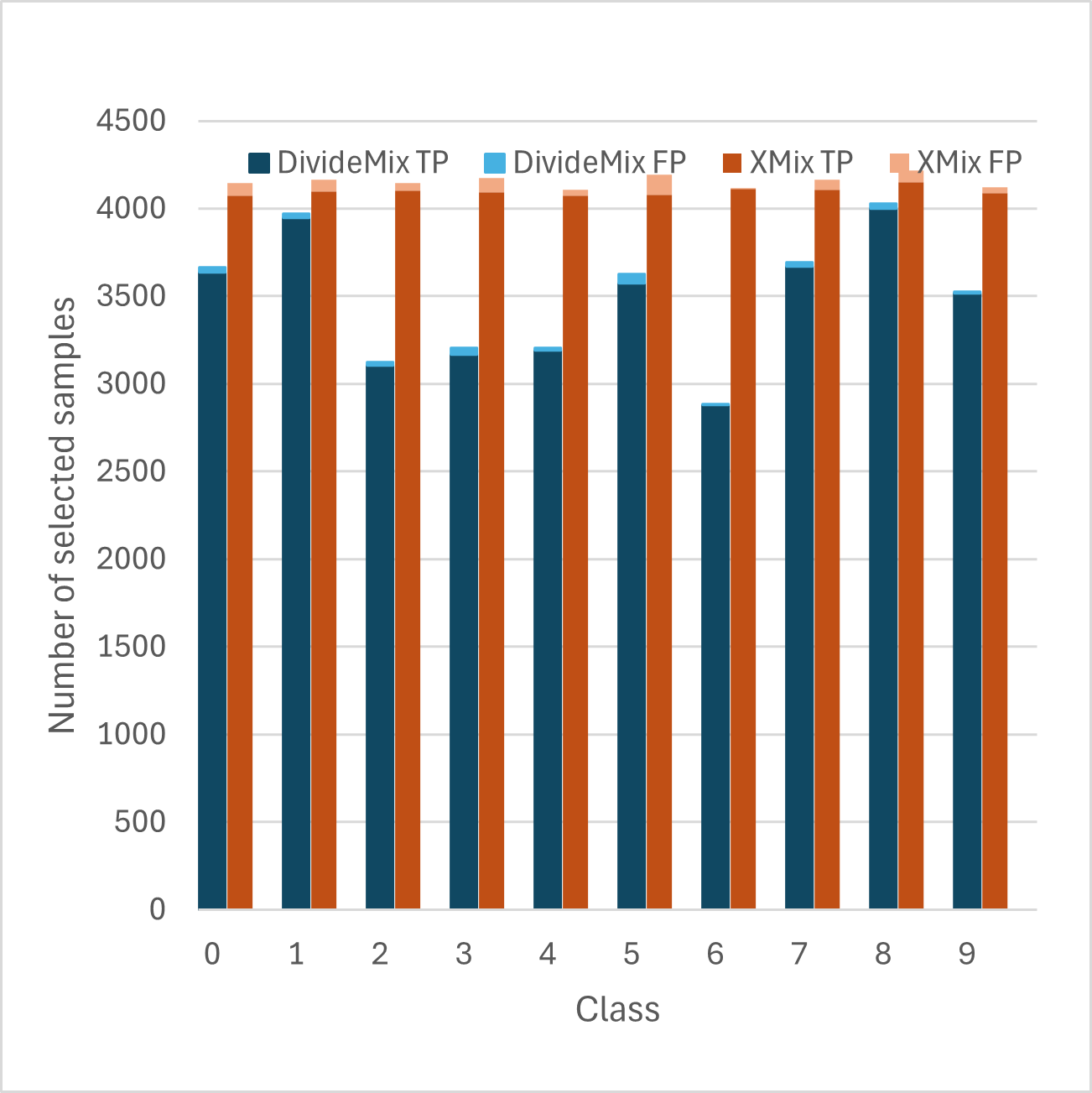}}
    {\includegraphics[width=\linewidth]{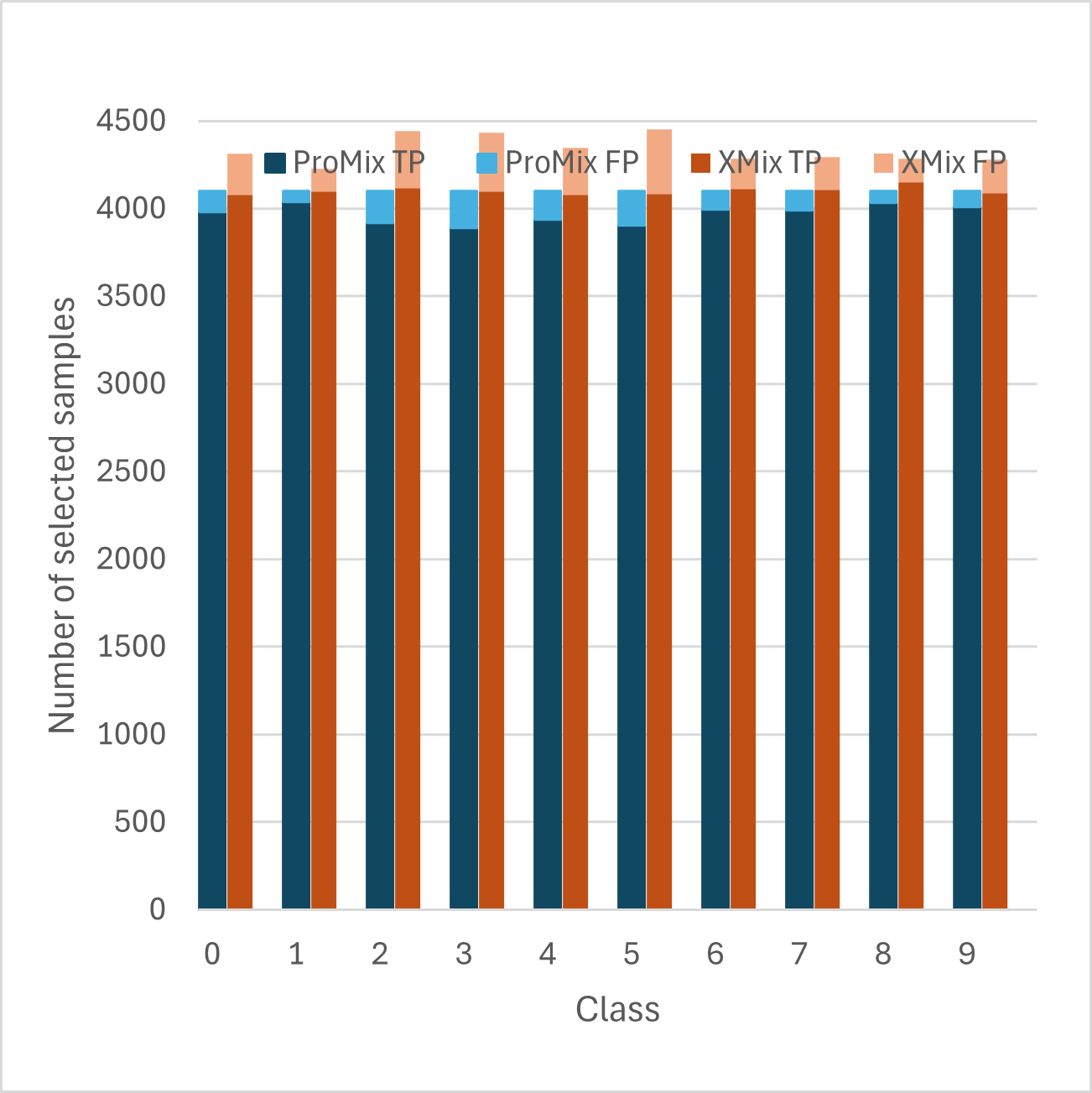}}
    \centerline{(a) 20\% Sym}
\end{minipage}
\begin{minipage}[htb]{0.23\linewidth}
    \centering
    {\includegraphics[width=\linewidth]{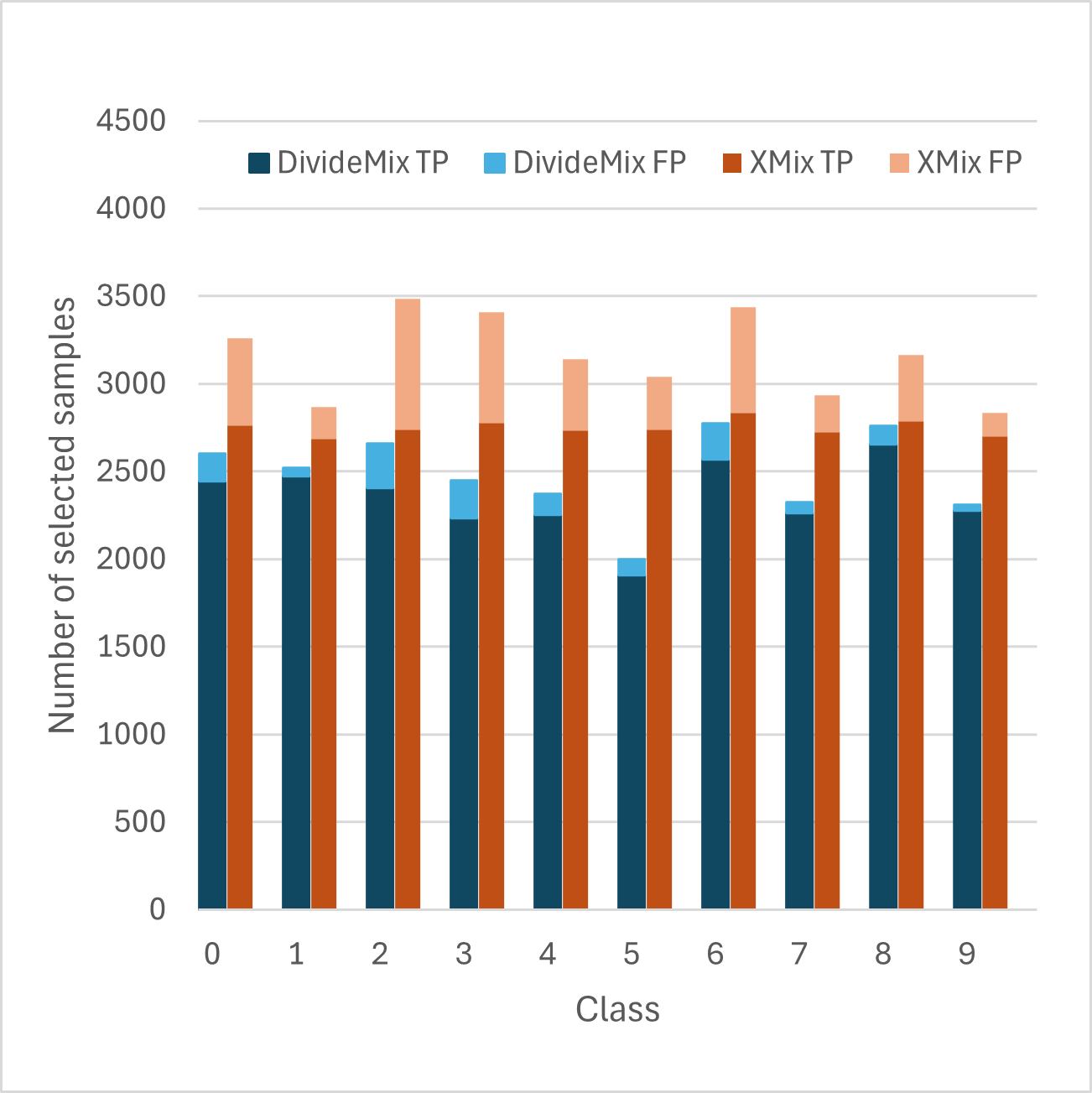}}
    {\includegraphics[width=\linewidth]{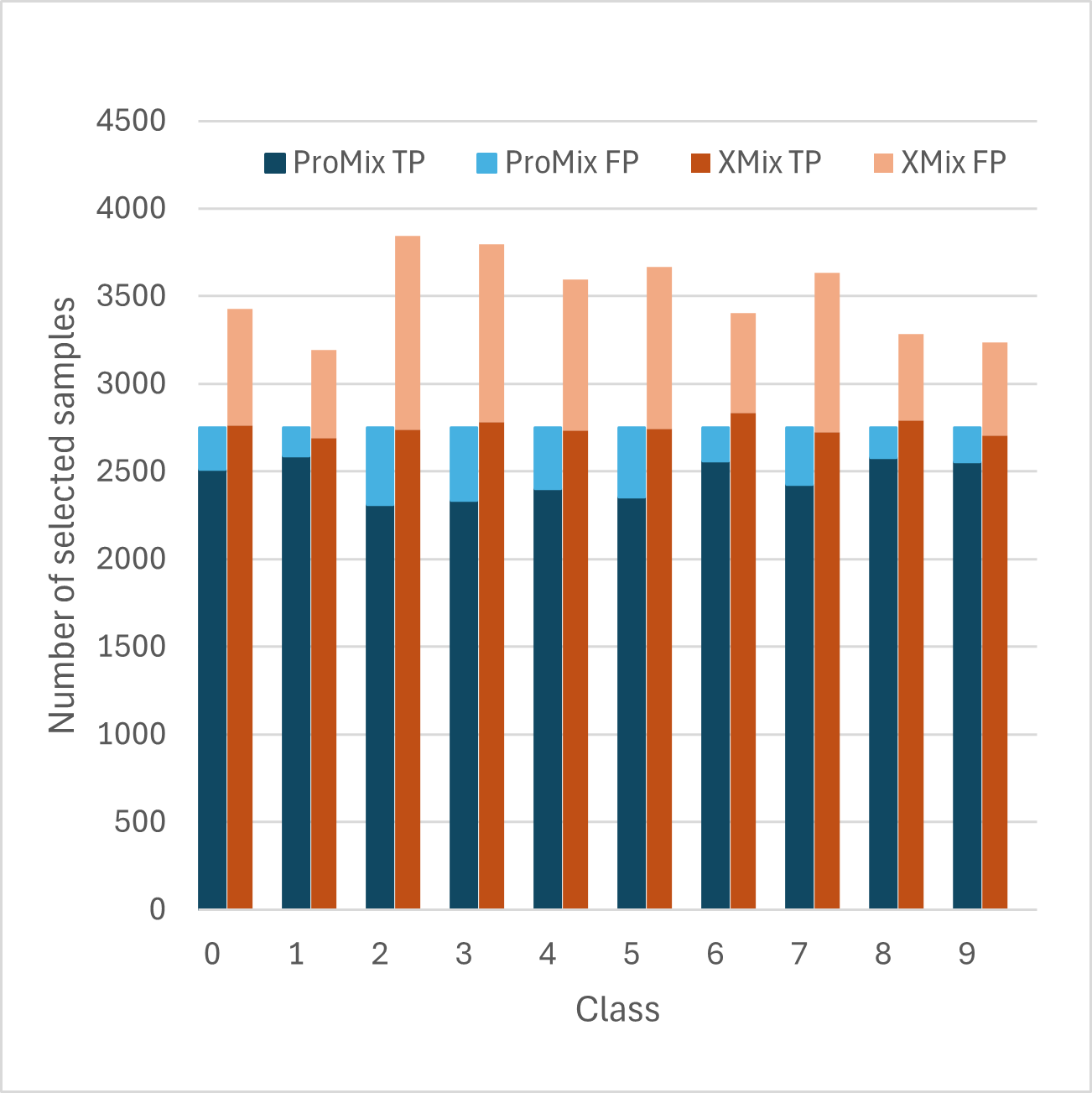}}
    \centerline{(b) 50\% Sym.}
\end{minipage}
\begin{minipage}[htb]{0.23\linewidth}
    \centering
    {\includegraphics[width=\linewidth]{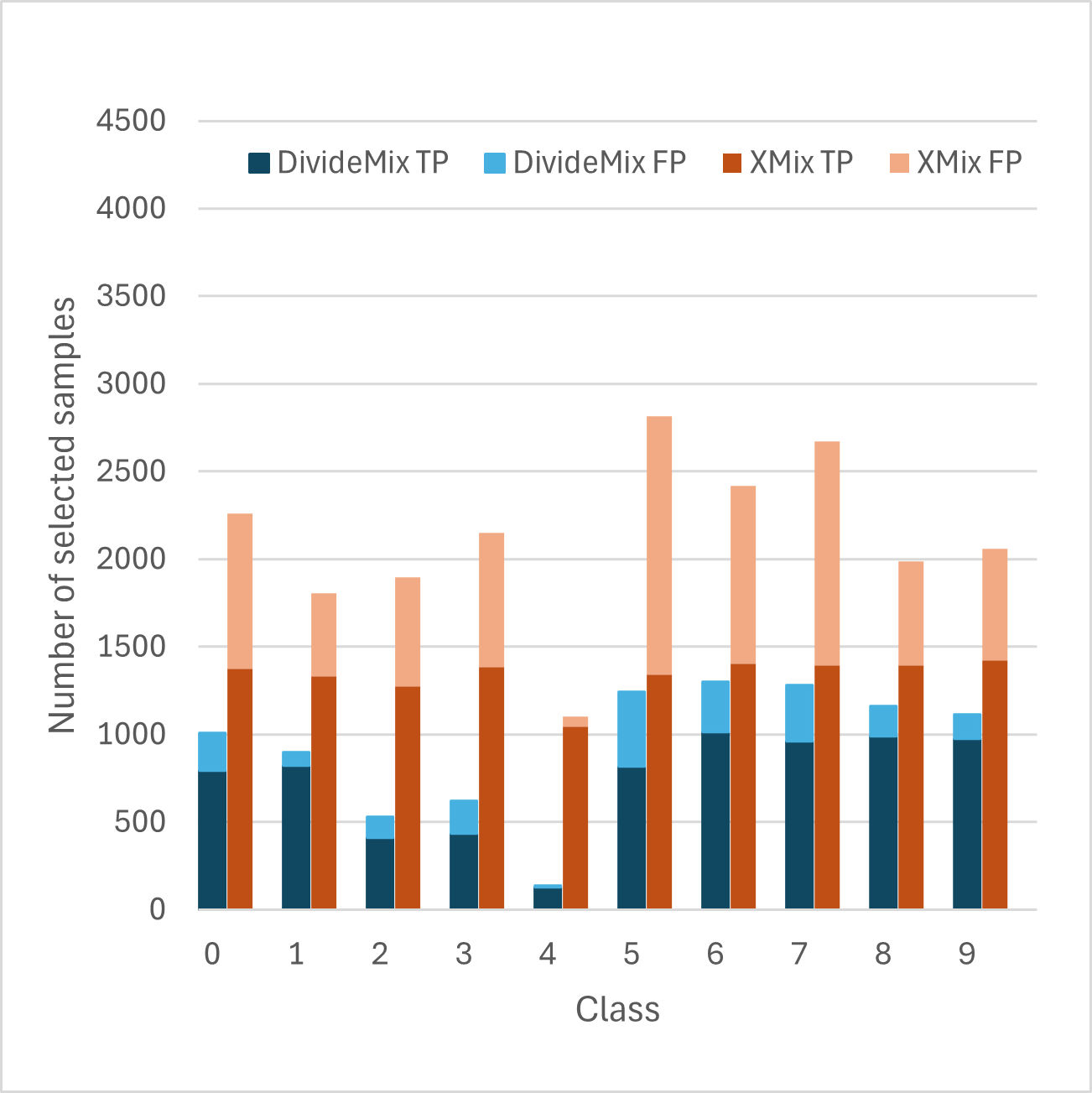}}
    {\includegraphics[width=\linewidth]{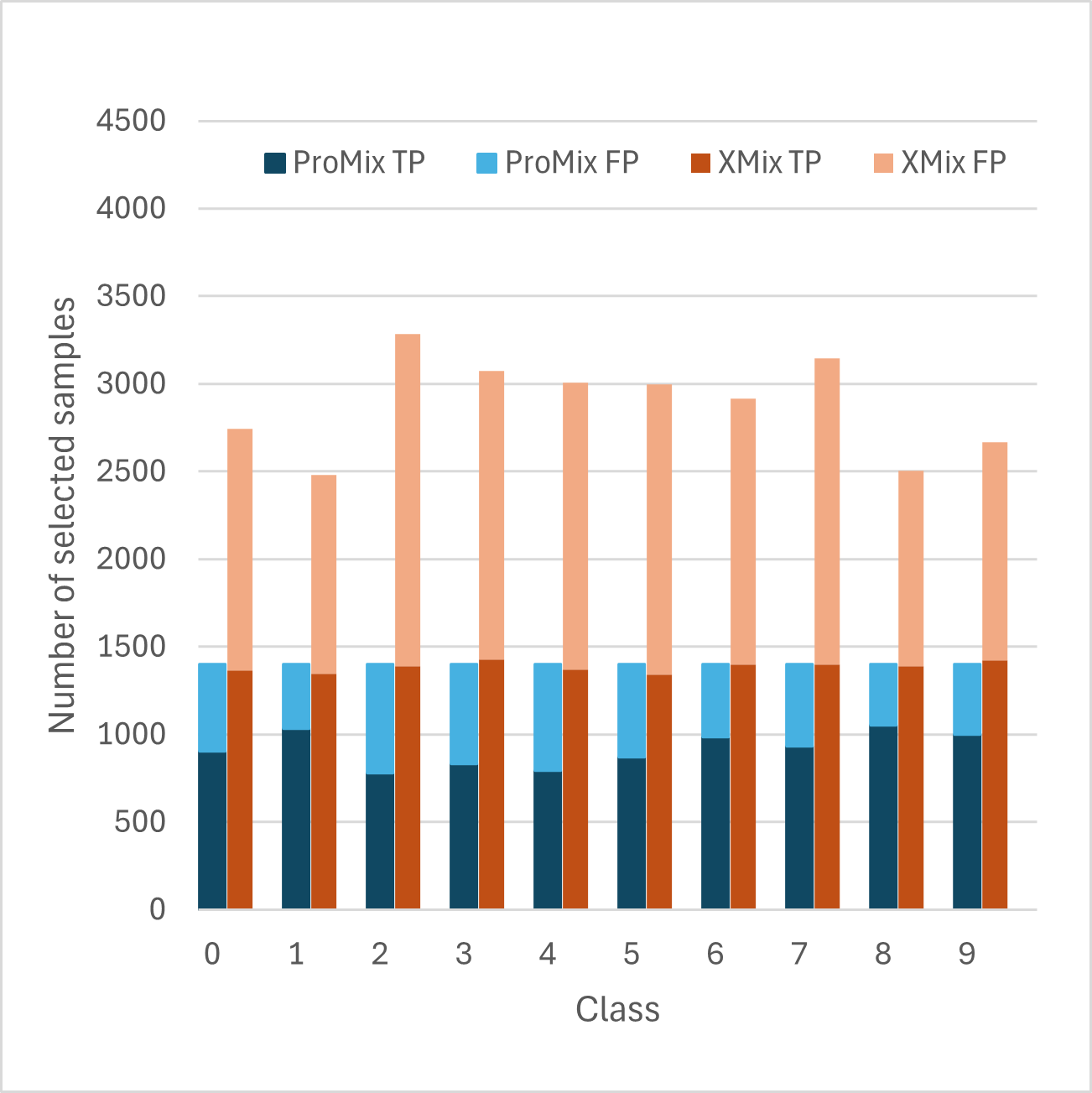}}
    \centerline{(c) 80\% Sym.}
\end{minipage}
\begin{minipage}[htb]{0.23\linewidth}
    \centering
    {\includegraphics[width=\linewidth]{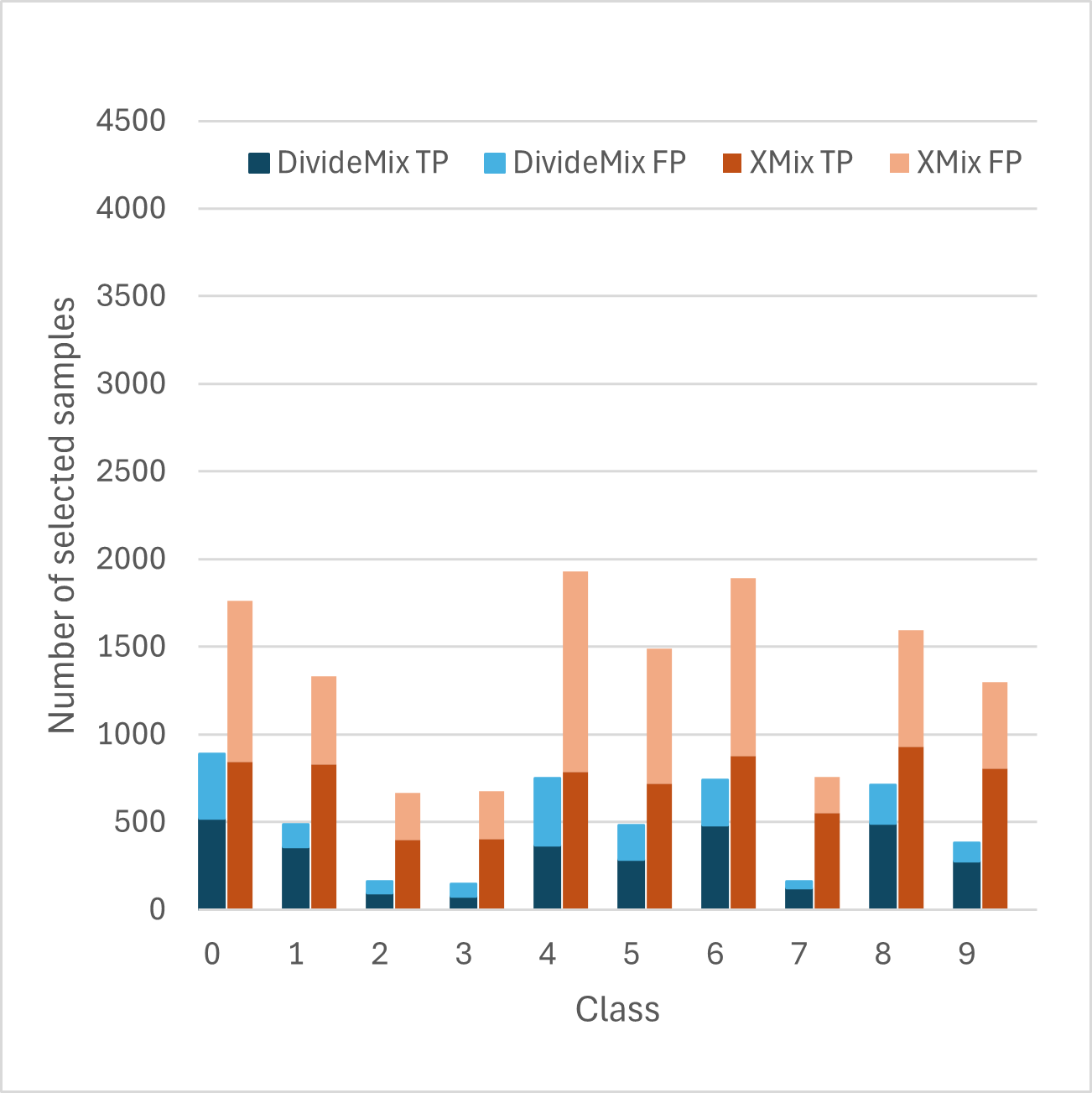}}
    {\includegraphics[width=\linewidth]{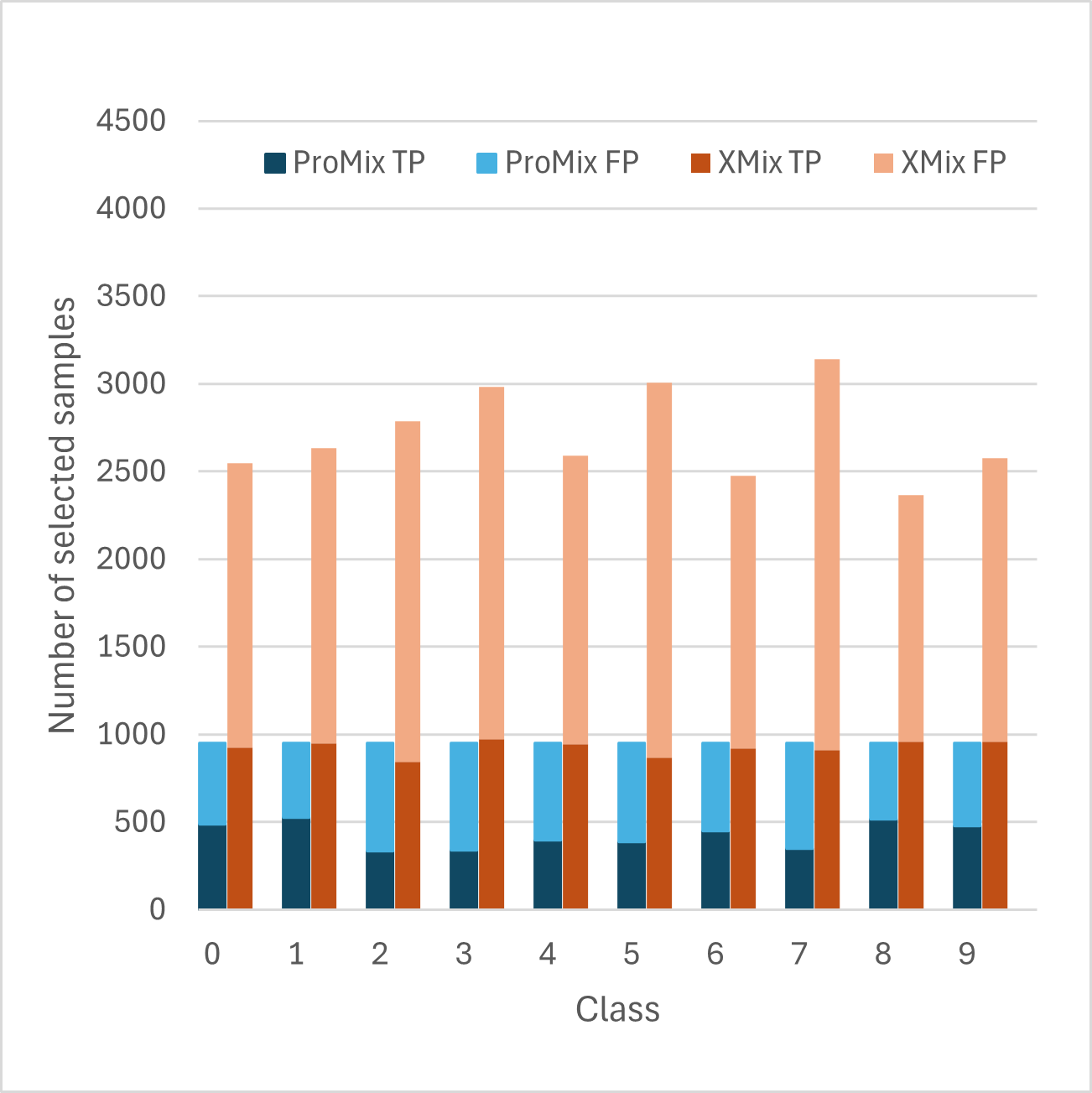}}
    \centerline{(d) 90\% Sym.}
\end{minipage}
\caption{Selected true clean samples at an early epoch of training on the CIFAR-10 dataset with different noise settings. \textbf{Upper Row:} DivideMix vs.\ XMix (DivideMix+); \textbf{Bottom Row:} ProMix vs.\ XMix (ProMix+). TP (true positives) and FP (false positives) represent true and false clean samples, displayed in light and dark shades, respectively. Zoom in for details.}
\label{fig:selected_samples_noise}
\end{figure*}

As highlighted by the \colorbox{pink}{red} blocks in Algorithm~\ref{alg:xmix}, XMix introduces several improvements over traditional sample selection methods such as DivideMix.

\noindent \textbf{Estimating noise rate.} Unlike DivideMix, which relies on prior knowledge of the noise level or manual tuning of thresholds, XMix adaptively estimates the class-wise noise rate using local neighbors in self-supervised feature space. This makes our approach more practical and robust, especially in real-world scenarios where the true noise level is unknown.

\noindent \textbf{Balanced and expanded sample selection.} By leveraging local smoothness in feature space, XMix can identify additional true clean samples. This improvement is particularly significant when the label noise is heavy, where standard methods often fail to collect enough clean samples for effective semi-supervised learning. Our method also addresses the common issue of class imbalance by adjusting the neighbor expansion strategy for underrepresented classes, resulting in a more balanced clean subset.

\noindent \textbf{More accurate pseudo labels.} Instead of simply averaging potentially unreliable predictions from augmented samples, our nearest neighbor pseudo-labeling mechanism leverages trustworthy local information in feature space. This enables XMix to generate more accurate pseudo-labels for noisy data, providing stronger training signals during the semi-supervised phase.

Together, these contributions make XMix simple yet highly effective, and easily compatible with existing sample selection frameworks.

\section{Experiments}
\label{section: experiments}

\subsection{Settings}
\noindent \textbf{Datasets.}
We evaluate the effectiveness of XMix on several datasets, including CIFAR-10, CIFAR-100 \cite{krizhevsky_learning_2009}, CIFARN \cite{wei_learning_2021}, ANIMAL-10N \cite{song_selfie_2019}, and Mini-WebVision \cite{li_webvision_2017}. For CIFAR-10 and CIFAR-100, we examine performance under synthetic symmetric and asymmetric label noise conditions, as outlined in Section \ref{section: background}. Additionally, CIFAR-10N and CIFAR-100N \cite{wei_learning_2021} variants, which use the same input images but real-world noisy labels annotated via Amazon Mechanical Turk, are included. The noise label portion in CIFARN ranges from $9\%$ to $40\%$.
The ANIMAL-10N dataset, another real-world collection of animal images, contains five pairs of easily confusable species and has a noise rate of approximately $8\%$. The last real-world noisy dataset is Mini-WebVision which contains about 61K Google images on the first 50 classes from the WebVision dataset, with an estimated label noise rate of $20\%$.

\noindent \textbf{Implementation details.}
We follow the same experimental settings as DivideMix and ProMix, employing a dual ResNet-18 architecture as the classification backbone for both CIFAR-10 and CIFAR-100. For real-world datasets, we use different architectures to better suit their characteristics---for example, VGG-19 with batch normalization for ANIMAL-10N and InceptionResNetV2 for WebVision. The model is trained for 300 epochs (600 epochs for the ProMix baseline), including a warm-up phase of 10 epochs for CIFAR-10 and 30 epochs for CIFAR-100 to activate the memorization effect. Training is optimized using SGD with a momentum of 0.9 and a weight decay of 0.0005. We use a batch size of 256 and an initial learning rate of 0.05, which decays according to a cosine schedule.

To adapt threshold selection, we estimate the noise rate class-wise and apply this estimate to determine the selection thresholds for DivideMix and ProMix according to their respective procedures. We utilize a pretrained DINOv2 self-supervised model to extract image features. Nearest neighbors are then computed for each sample in feature space using Euclidean distance; specifically, both the $k$ and $2k$ nearest neighbors are identified, with $k = 10$.

\subsection{Results on Synthetic Benchmarks}
Classification results for CIFAR-10 and CIFAR-100 under different synthetic noise levels (symmetric or asymmetric) are presented in Table~\ref{tab:cifar_extreme_noise} and Table~\ref{tab:cifar_syn_noise}. Compared to state-of-the-art methods such as DivideMix, ProMix, and other recent approaches, our XMix framework consistently achieves competitive or superior results across different noise ratios.

Notably, XMix outperforms the baselines more significantly as the noise level increases. For example, at extreme symmetric noise levels of 80\% and 90\% on CIFAR-10 and CIFAR-100, XMix achieves larger gains over its base models (DivideMix and ProMix), highlighting the benefit of our balanced and expanded sample selection strategy. Additionally, the results under 40\% asymmetric noise indicate that XMix remains robust to structured noise, matching or exceeding other methods.

These results confirm that XMix can effectively enhance standard sample selection methods, removing the need for prior noise knowledge and providing more balanced and reliable training signals, especially in challenging high-noise scenarios.

A finer breakdown across $10\%$, $30\%$, and $40\%$ asymmetric noise (Table~\ref{tab:cifar_asym_noise_full} in the supplementary material) confirms that XMix (ProMix+) achieves the best or near-best accuracy at every level, e.g., $97.6\%$ at $10\%$ noise on CIFAR-10 and $81.1\%$ at $10\%$ noise on CIFAR-100.

\subsection{Results on Real-World Benchmarks}
We adopt ProMix as the base model for real-world benchmarks. On CIFAR-10N and CIFAR-100N (Table~\ref{tab:cifarn}), XMix outperforms all other methods, with XMix showing clear advantages under higher noise. On ANIMAL-10N (Table~\ref{tab:animal_and_web}), XMix achieves the best accuracy (90.3\%), surpassing all baselines. Results on Mini-WebVision and ILSVRC12 (Table~\ref{tab:animal_and_web}) further demonstrate that XMix remains strong and competitive across diverse real-world noise settings.

\begin{table}[bth!]
\centering
\resizebox{0.33\linewidth}{!}{
\begin{tabular}{ll}
\toprule
Method & Acc. \\
\midrule
PLC \cite{zhang_plc_2021} & 83.4\% \\
SELC \cite{lu_selc_2022} & 83.7\% \\
ProMix \cite{xiao_promix_2023} & 90.0\% \\
CLIPCleaner \cite{feng_clipcleaner_2024} & 88.9\% \\
LSL \cite{Kim_LSL_2024} & 89.1\% \\
\midrule
XMix & \textbf{90.3\%} \\
\bottomrule
\end{tabular}
}
\quad
\resizebox{0.6\linewidth}{!}{
\begin{tabular}{lllll}
\toprule
\multirow{2}{*}{Method} & \multicolumn{2}{c}{WebVision} & \multicolumn{2}{c}{ILSVRC12} \\
& top1 & top5 & top1 & top5 \\
\midrule
DivideMix \cite{li_dividemix_2020} & 77.3\% & 91.6\% & 75.2\% & 90.8\% \\
SELC \cite{lu_selc_2022} & 74.4\% & 90.7\% & 70.9\% & 90.7\% \\
UNICON \cite{karim_unicon_2022} & 77.6\% & 93.4\% & 75.3\% & 93.7\% \\
RankMatch \cite{zhang_rankmatch_2023} & 79.9\% & 93.6\% & 77.4\% & 94.3\% \\
CLIPCleaner \cite{feng_clipcleaner_2024} & \textbf{81.6\%} & 93.3\% & \textbf{77.8\%} & 92.1\% \\
LSL \cite{Kim_LSL_2024} & 81.4\% & 93.0\% & 77.0\% & 91.8\% \\
PLReMix \cite{liu_rlremix_2025} & 81.4\% & 93.7\% & 77.7\% & 93.0\% \\
\midrule
XMix & 80.2\% & \textbf{94.7\%} & 76.9\% & \textbf{95.0\%} \\
\bottomrule
\end{tabular}
}
\caption{\textbf{Left}: Classification accuracies on ANIMAL-10N. The best results are in \textbf{bold}. \textbf{Right}: Top-1 (Top-5) classification accuracies on Mini-WebVision and ILSVRC12 validation set. The best results are in \textbf{bold}.}
\label{tab:animal_and_web}
\end{table}

\begin{table}[bth!]
\centering
\resizebox{\linewidth}{!}{
\begin{tabular}{lllll}
\toprule
Dataset                           & \multicolumn{3}{c}{CIFAR-10N}               & CIFAR-100N  \\
\midrule
Methods\textbackslash{}Noise type & Aggre    & Rand1  & Worst  & Noisy Fine \\
\midrule
DivideMix \cite{li_dividemix_2020}                       & 95.0\%   & 95.2\% & 92.6\% & 71.1\%     \\
ELR+ \cite{liu_early-learning_2020}                             & 94.8\%   & 94.4\% & 91.1\% & 66.7\%     \\
CORES \cite{cheng_learning_2021}                           & 95.3\%   & 94.5\% & 91.7\% & 55.7\%     \\
PES(Semi) \cite{bai_understanding_2021}                       & 94.7\%   & 95.1\% & 92.7\% & 70.4\%     \\
ProMix \cite{xiao_promix_2023}                           & \textbf{97.7\%}  & 97.4\% & 96.3\% & 73.8\%     \\
PADDLES \cite{Huang_PADDLES_2023} & 95.5\% & 95.9\% & 93.9\% & 71.3\%  \\
CS-Isolate \cite{Lin_CSIsolate_2023}                           & 95.3\%  & 95.4\% & 94.3\% & -     \\
LSL \cite{Kim_LSL_2024}                           & -  & - & 94.6\% & \textbf{74.5\%}     \\
\midrule
XMix                            & \textbf{97.7\%} & \textbf{97.5\%} & \textbf{96.4\%} & 74.2\%    \\
\bottomrule
\end{tabular}
}
\caption{Classification accuracies on CIFAR-10N and CIFAR-100N under real-world label noise (full breakdown including Rand2/Rand3 in the supplementary material, Table~\ref{tab:cifarn_full}). The best results are in \textbf{bold}.}
\label{tab:cifarn}
\end{table}

\subsection{Ablation Study}
\label{ablation}

\noindent \textbf{Compatibility of Proposed Modules.}
As shown in Table~\ref{tab:cifar_syn_noise}, XMix is compatible with existing sample selection methods such as DivideMix and ProMix; incorporating self-supervised features consistently improves the performance of these base methods across different noise settings.

\noindent \textbf{Efficacy of class-wise noise rate estimation.}
For class-wise noise rate estimation, we first simulate various known noise levels on the CIFAR-10 dataset and measure the label matching rates within each sample's nearest neighbors in feature space. We then fit a quadratic correction model to adjust for estimation bias. This calibrated model is subsequently applied to other datasets to provide a coarse class-wise noise estimate without requiring prior knowledge of the true noise level. This estimate is not intended to recover the precise noise rate; rather, it only needs to be accurate enough to place a dataset into one of a few discrete noise regimes, which is what determines the hyperparameter configuration (e.g., selection ratio and threshold schedule) used by the baseline sample selection method. In practice, we find this coarse estimate sufficient: baseline methods are far more sensitive to which regime they are assigned to than to the exact noise rate within that regime.

\begin{table}[th!]
\centering
\resizebox{\linewidth}{!}{
\begin{tabular}{lllll}
\toprule
Method\textbackslash{}Noise rate    & 20\%   & 50\%   & 80\%   & 90\%   \\
\midrule
XMix & 96.4\% & 96.1\% & 94.4\% & 91.2\% \\
XMix (SimCLR \cite{chen_simple_2020}) & 96.2\% & 95.1\% & 93.9\% & 87.2\% \\
XMix w/o B\&E & 96.1\% & 94.7\% & 93.3\% & 82.3\% \\
XMix w/o NN & 96.2\% & 95.5\% & 93.9\% & 90.1\% \\
DivideMix & 96.1\% & 94.6\% & 93.2\% & 76.0\% \\
\bottomrule
\end{tabular}
}
\caption{Ablation study of XMix (DivideMix+) on CIFAR-10 with symmetric label noise. The method name in parentheses indicates the feature encoder; DINOv2 \cite{oquab_dinov2_2023} is used by default unless otherwise specified. "B\&E" denotes the balanced \& expanded sample selection module, and "NN" refers to the nearest neighbor pseudo-labeling component.}
\label{tab:ablation}
\end{table}

Table \ref{tab:ablation} presents an ablation study of XMix on CIFAR-10 with symmetric label noise. Using a stronger encoder (DINOv2) yields a more consistent feature space, but replacing it with SimCLR results in comparable performance, indicating that the gains of XMix are not primarily driven by encoder strength. In contrast, removing either the balanced-and-expanded (B\&E) sample selection or nearest-neighbor (NN) pseudo-labeling leads to noticeable degradation, particularly under extreme noise. These results show that XMix's performance improvements mainly stem from its feature-space clustering and filtering mechanisms, which consistently increase the number of correctly identified clean samples across different encoders.

\noindent \textbf{Balanced and expanded sample selection.}
We evaluate the effectiveness of the XMix sample selection module by comparing the number of true clean samples it selects with those selected by base models. As shown in Figure~\ref{fig:selected_samples_noise}, XMix consistently identifies significantly more true clean samples, sometimes doubling or even tripling the count. This advantage is particularly clear under severe noise conditions, as illustrated in Figure~\ref{fig:selected_samples_noise}(b) and (c), where XMix substantially expands the clean sample set for classes that would otherwise have very few selected examples, albeit with some errors. The results also reveal that class imbalance in the selected clean subset becomes more pronounced as noise levels increase. By leveraging class-aware expansion, XMix helps mitigate this imbalance, leading to a more balanced distribution of selected samples across classes.

Conversely, at lower noise levels, the standard low-loss criterion alone is effective at identifying most clean samples. Therefore, the benefit of expanding the clean set using neighborhood information becomes marginal. Moreover, XMix may sometimes misclassify noisy samples as clean samples. As shown in Figure~\ref{fig:selected_samples_noise}(b), although XMix does slightly increase the number of true clean samples under lower noise, it also introduces a noticeable number of false positives. This trade-off means that the disadvantages of inaccurate expansion can outweigh its benefits in low-noise settings, which aligns with the observation that XMix's performance gain diminishes at lower noise levels, as reflected in Table~\ref{tab:cifar_syn_noise} and \ref{tab:ablation}.

\noindent \textbf{Nearest neighbor pseudo-labeling.}
As shown in Figure~\ref{fig:pseudo_label_recall} in the supplementary material, nearest neighbor pseudo-labeling produces more accurate pseudo-labels for the unlabeled data than the base models. This improved pseudo-label quality strengthens the semi-supervised learning phase. While beneficial, the gains from nearest neighbor pseudo-labeling are incremental relative to the improvements achieved through balanced and expanded sample selection, as evidenced in Table~\ref{tab:ablation}.

\FloatBarrier

\section{Conclusion}
\label{section: conclusion}
We introduced XMix, a simple yet robust LNL framework that leverages the local smoothness of self-supervised feature space to adaptively select balanced clean samples and generate more accurate pseudo-labels, outperforming existing methods without requiring prior noise information.

{\small
\bibliographystyle{ieeenat_fullname}
\bibliography{main}
}

\maketitlesupplementary

\appendix

\section{Additional Quantitative Results}
\label{app:quant}
Table~\ref{tab:selection_precision_cifar10_sym} and Table~\ref{tab:selection_recall_cifar10_sym} report the precision and recall of sample selection on CIFAR-10 across the full range of symmetric label noise levels studied in the main paper. As the noise level increases, both precision and recall decrease across all methods, reflecting the growing difficulty of accurately identifying clean samples.

XMix exhibits a distinct trade-off between precision and recall, particularly under extreme noise. While it achieves slightly lower precision than its base methods (DivideMix and ProMix), XMix substantially improves recall at higher noise levels. For example, at $98\%$ noise, XMix (DivideMix+) achieves a recall of $75.08\%$, compared to DivideMix's $31.43\%$, while maintaining competitive precision. This trade-off lets XMix recover far more clean samples than its base methods, which is essential for enabling effective semi-supervised learning once label noise becomes severe; the results support prioritizing recall over precision in this regime.

\FloatBarrier

\begin{table}[h]
\centering
\resizebox{\linewidth}{!}{
\begin{tabular}{lllll}
\toprule
Noise Level & DivideMix & XMix (DivideMix+) & ProMix  & XMix (ProMix+) \\
\midrule
20\%        & 99.31\%   & 98.60\%           & 96.82\% & 94.61\%        \\
50\%        & 94.88\%   & 87.41\%           & 89.56\% & 78.70\%        \\
80\%        & 80.32\%   & 65.77\%           & 65.54\% & 48.46\%        \\
90\%        & 38.59\%   & 25.59\%           & 44.83\% & 34.44\%        \\
92\%        & 31.75\%   & 22.28\%           & 37.34\% & 28.68\%        \\
95\%        & 21.49\%   & 16.42\%           & 32.01\% & 26.14\%        \\
98\%        & 17.33\%   & 15.27\%           & 14.78\% & 14.59\%       \\
\bottomrule
\end{tabular}
}
\caption{Precision of sample selection on CIFAR-10 under different levels of symmetric label noise.}
\label{tab:selection_precision_cifar10_sym}
\end{table}

\begin{table}[h]
\centering
\resizebox{\linewidth}{!}{
\begin{tabular}{lllll}
\toprule
Noise Level & DivideMix & XMix (DivideMix+) & ProMix  & XMix (ProMix+) \\
\midrule
20\%        & 84.53\%   & 99.86\%           & 96.76\% & 99.97\%        \\
50\%        & 85.33\%   & 99.88\%           & 89.53\% & 99.97\%        \\
80\%        & 52.71\%   & 95.73\%           & 65.80\% & 99.24\%        \\
90\%        & 77.02\%   & 99.22\%           & 44.37\% & 96.30\%        \\
92\%        & 77.55\%   & 99.63\%           & 39.27\% & 94.08\%        \\
95\%        & 72.46\%   & 98.48\%           & 32.98\% & 87.64\%        \\
98\%        & 31.43\%   & 75.08\%           & 15.07\% & 63.06\%       \\
\bottomrule
\end{tabular}
}
\caption{Recall of sample selection on CIFAR-10 under different levels of symmetric label noise.}
\label{tab:selection_recall_cifar10_sym}
\end{table}

A finer breakdown of asymmetric-noise results (Table~\ref{tab:cifar_asym_noise_full}) and the full CIFAR-N breakdown including the Rand2/Rand3 splits (Table~\ref{tab:cifarn_full}) are given below for completeness.

\begin{table}[h]
\centering
\resizebox{\linewidth}{!}{
\begin{tabular}{lllllll}
\toprule
Dataset                           & \multicolumn{3}{c}{CIFAR-10}  & \multicolumn{3}{c}{CIFAR-100} \\
\midrule
Method\textbackslash{}Noise rate & 10\%     & 30\%   & 40\%   & 10\%          & 30\%          & 40\%   \\
\midrule
DivideMix \cite{li_dividemix_2020}                        & 93.8\%   & 92.5\% & 91.7\% & 71.6\%        & 69.5\%        & 55.1\% \\
SELC \cite{lu_selc_2022}                             & -    & -  & 92.9\% & -         & -         & 73.6\% \\
UNICON \cite{karim_unicon_2022}                           & 95.3\%   & 94.8\% & 94.1\% & 78.2\%        & 75.6\%        & 74.8\% \\
CrossSplit \cite{kim_crosssplit_2023}              & 96.9\%   & 96.4\% & 96.0\% & 80.7\%        & 78.5\%        & \textbf{76.8\%} \\
ProMix \cite{xiao_promix_2023}                           & 97.4\%    & 97.0\%  & 96.6\% & 80.3\%        & 80.1\%         & 76.2\%  \\
CLIPCleaner \cite{feng_clipcleaner_2024} & - & - & 94.9\%     & - & - & - \\
L2B \cite{zhou_l2b_2024} & - & - & 94.0\%     & - & - & - \\
\midrule
XMix (ProMix+) & \textbf{97.6\%}    & \textbf{97.2\%}  & \textbf{96.6\%}  & \textbf{81.1\%}         & \textbf{80.8\%}         & 76.5\% \\
\bottomrule
\end{tabular}
}
\caption{Classification accuracies on CIFAR-10 and CIFAR-100 under asymmetric label noise, with noise levels ranging from $10\%$ to $40\%$. The best results are in \textbf{bold}.}
\label{tab:cifar_asym_noise_full}
\end{table}

\begin{table}[h]
\centering
\resizebox{\linewidth}{!}{
\begin{tabular}{lllllll}
\toprule
Dataset                           & \multicolumn{5}{c}{CIFAR-10N}               & CIFAR-100N  \\
\midrule
Methods\textbackslash{}Noise type & Aggre    & Rand1  & Rand2  & Rand3  & Worst  & Noisy Fine \\
\midrule
DivideMix \cite{li_dividemix_2020}                       & 95.0\%   & 95.2\% & 95.2\% & 95.2\% & 92.6\% & 71.1\%     \\
ELR+ \cite{liu_early-learning_2020}                             & 94.8\%   & 94.4\% & 94.2\% & 94.3\% & 91.1\% & 66.7\%     \\
CORES \cite{cheng_learning_2021}                           & 95.3\%   & 94.5\% & 94.9\% & 94.7\% & 91.7\% & 55.7\%     \\
PES(Semi) \cite{bai_understanding_2021}                       & 94.7\%   & 95.1\% & 95.2\% & 95.2\% & 92.7\% & 70.4\%     \\
ProMix \cite{xiao_promix_2023}                           & \textbf{97.7\%}  & 97.4\% & \textbf{97.6\%} & \textbf{97.5\%} & 96.3\% & 73.8\%     \\
PADDLES \cite{Huang_PADDLES_2023} & 95.5\% & 95.9\% & 96.0\% & 96.0\% & 93.9\% & 71.3\%  \\
CS-Isolate \cite{Lin_CSIsolate_2023}                           & 95.3\%  & 95.4\% & 95.3\% & 95.5\% & 94.3\% & -     \\
LSL \cite{Kim_LSL_2024}                           & -  & - & - & - & 94.6\% & \textbf{74.5\%}     \\
\midrule
XMix                            & \textbf{97.7\%} & \textbf{97.5\%} & 97.4\% & \textbf{97.5\%} & \textbf{96.4\%} & 74.2\%    \\
\bottomrule
\end{tabular}
}
\caption{Classification accuracies on CIFAR-10N and CIFAR-100N under real-world label noise, with the full set of noise types (Aggre, Rand1, Rand2, Rand3, Worst, Noisy Fine). The best results are in \textbf{bold}.}
\label{tab:cifarn_full}
\end{table}

\section{Additional Qualitative Results}
\label{app:qual}
\noindent \textbf{Symmetric label noise.} Figure~\ref{fig:selected_samples_noise_full} and Figure~\ref{fig:selected_samples_noise_full_2} extend the qualitative comparison in Figure~\ref{fig:selected_samples_noise} to the full range of symmetric noise levels studied in this paper. XMix consistently selects more true clean samples, nearly doubling the count in some cases; this is most evident in the extreme settings ($92\%$--$98\%$), where only a handful of samples would otherwise be selected for certain classes. At lower noise levels, the low-loss criterion alone already recovers most clean samples, so the benefit of neighborhood expansion is smaller and occasionally introduces additional false positives, consistent with the diminishing gains observed in Table~\ref{tab:cifar_syn_noise}.

\begin{figure*}[!htb]
\centering
\begin{minipage}[htb]{0.24\linewidth}
    \centering
    {\includegraphics[width=\linewidth]{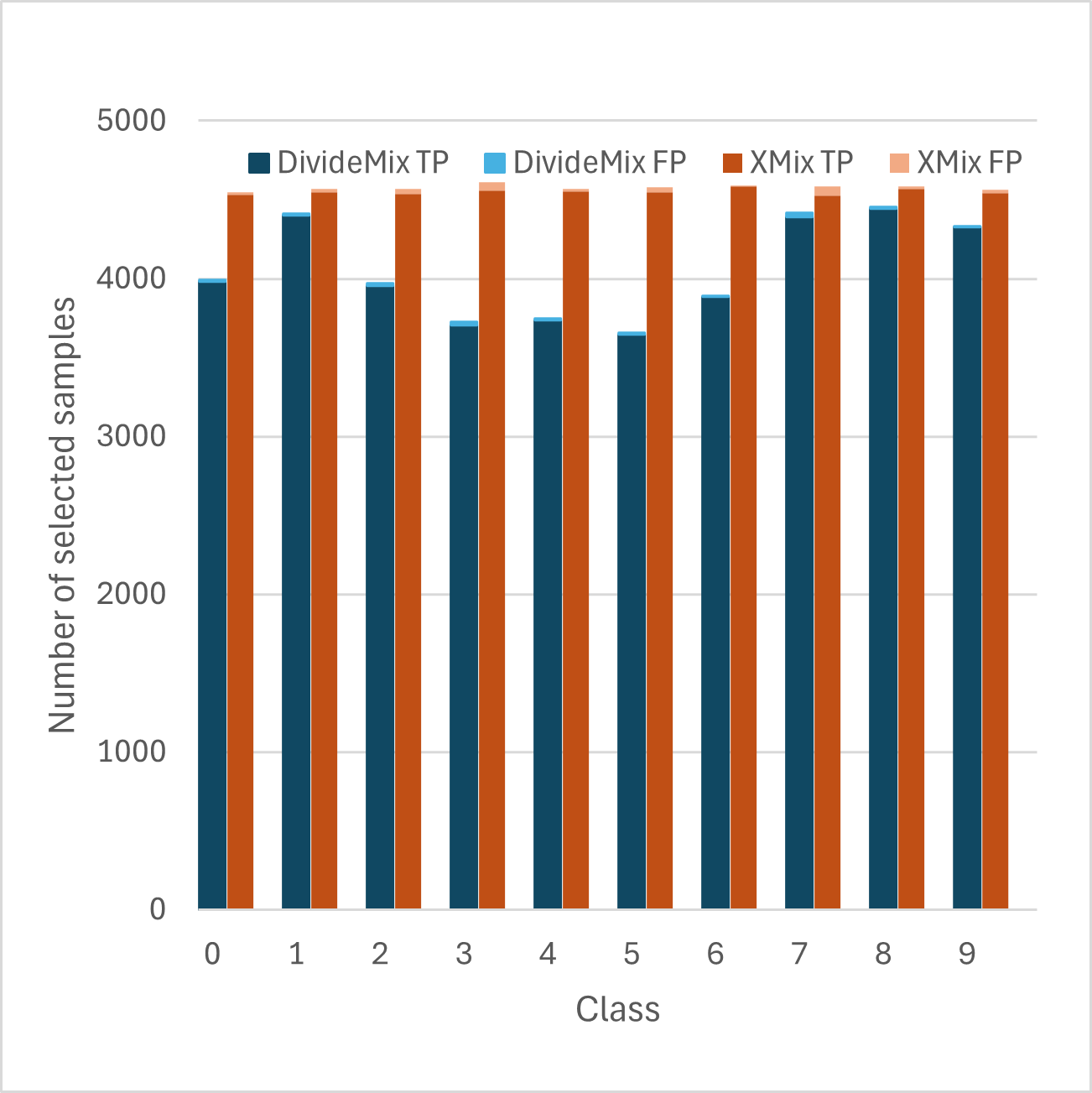}}
    {\includegraphics[width=\linewidth]{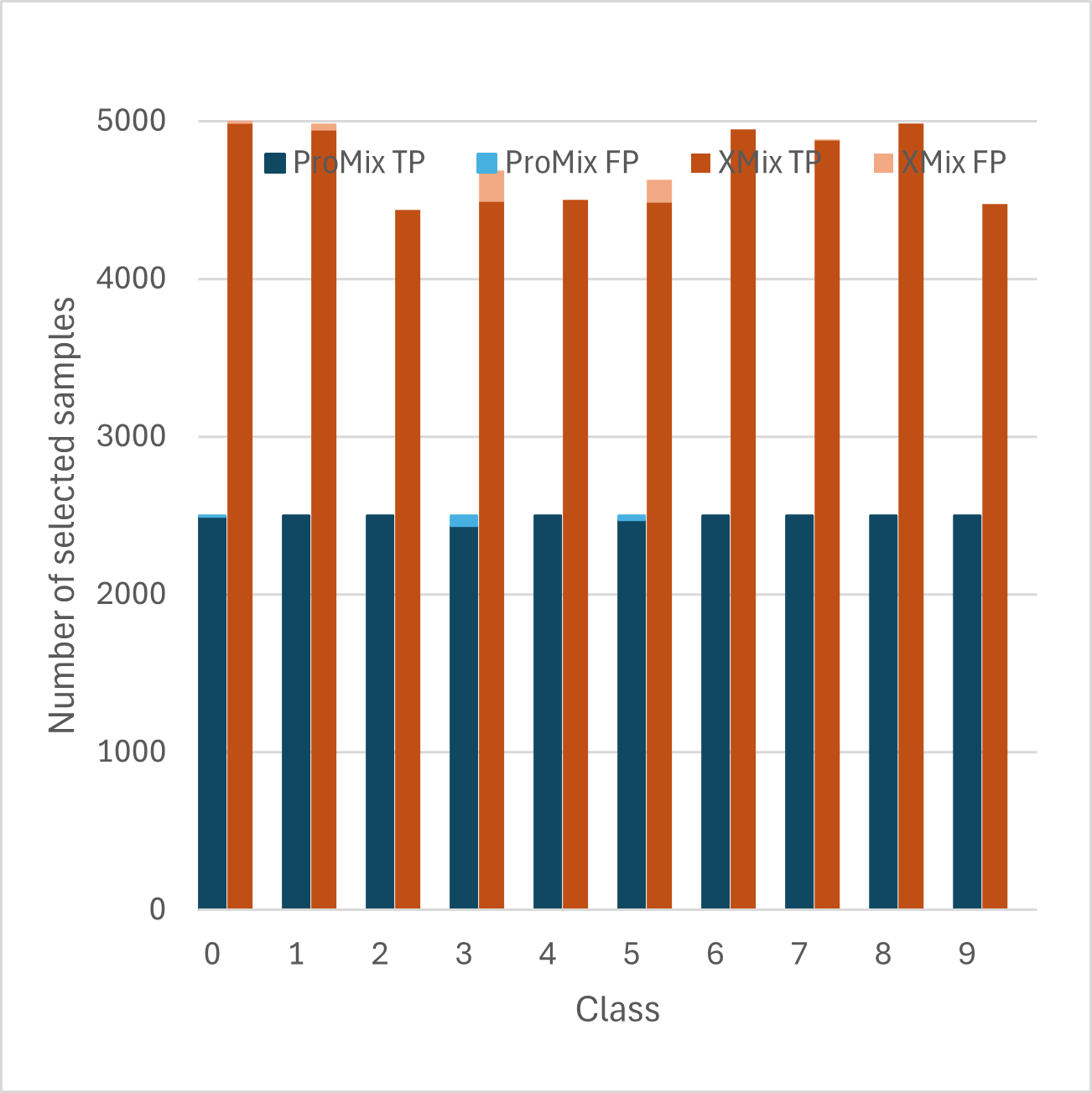}}
    \centerline{(a) 10\% Asym.}
\end{minipage}
\begin{minipage}[htb]{0.24\linewidth}
    \centering
    {\includegraphics[width=\linewidth]{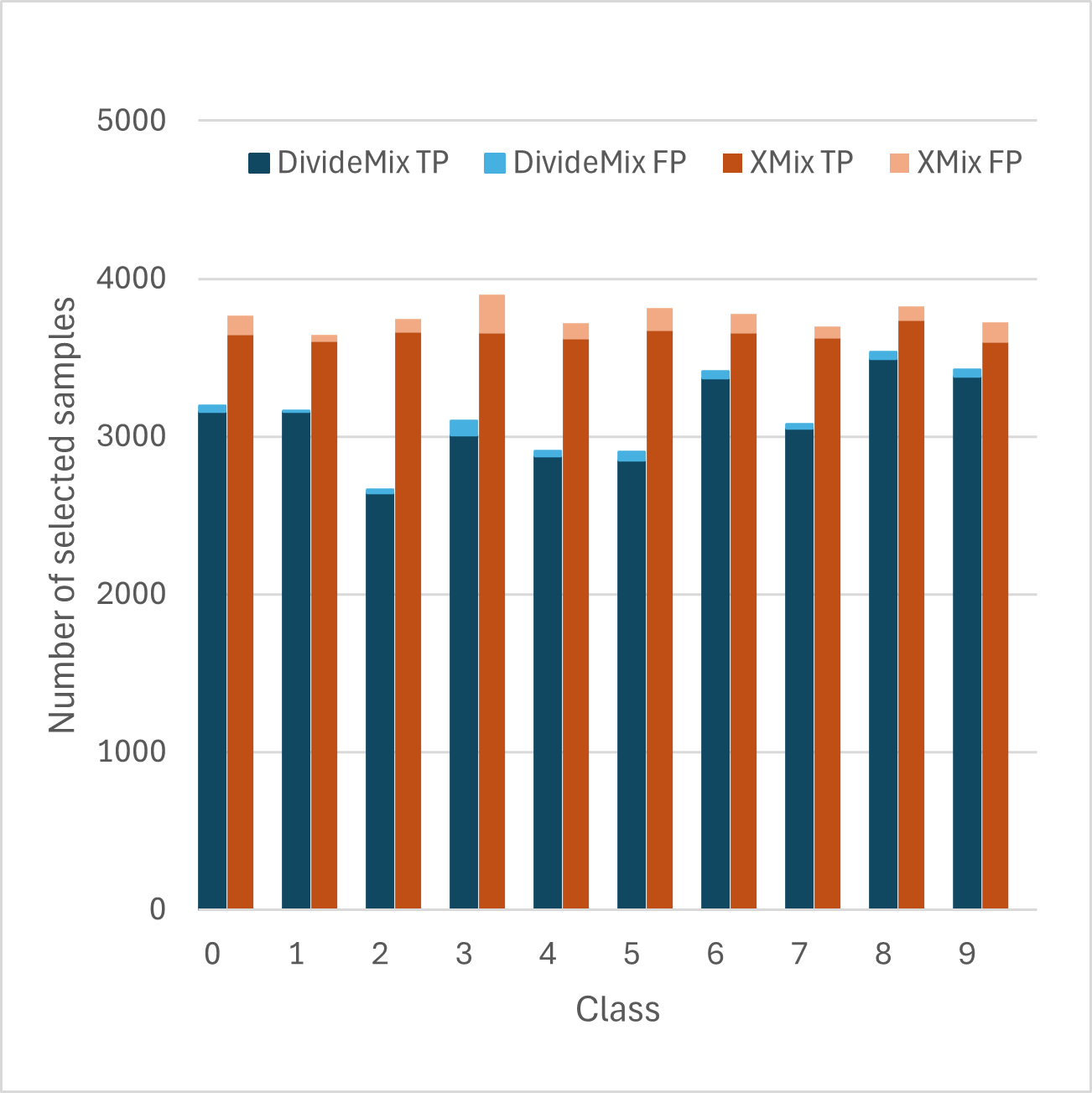}}
    {\includegraphics[width=\linewidth]{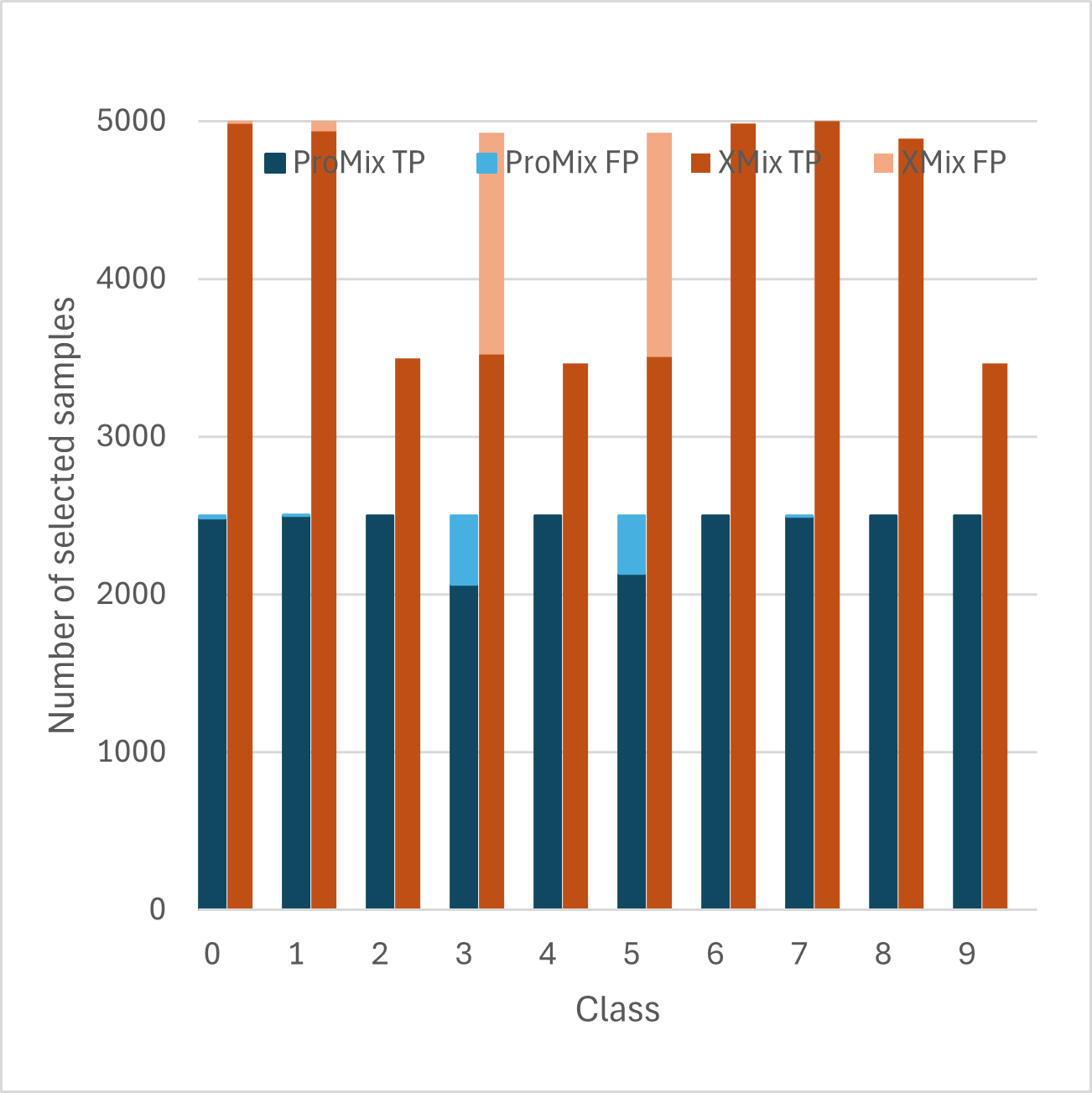}}
    \centerline{(b) 30\% Asym.}
\end{minipage}
\begin{minipage}[htb]{0.24\linewidth}
    \centering
    {\includegraphics[width=\linewidth]{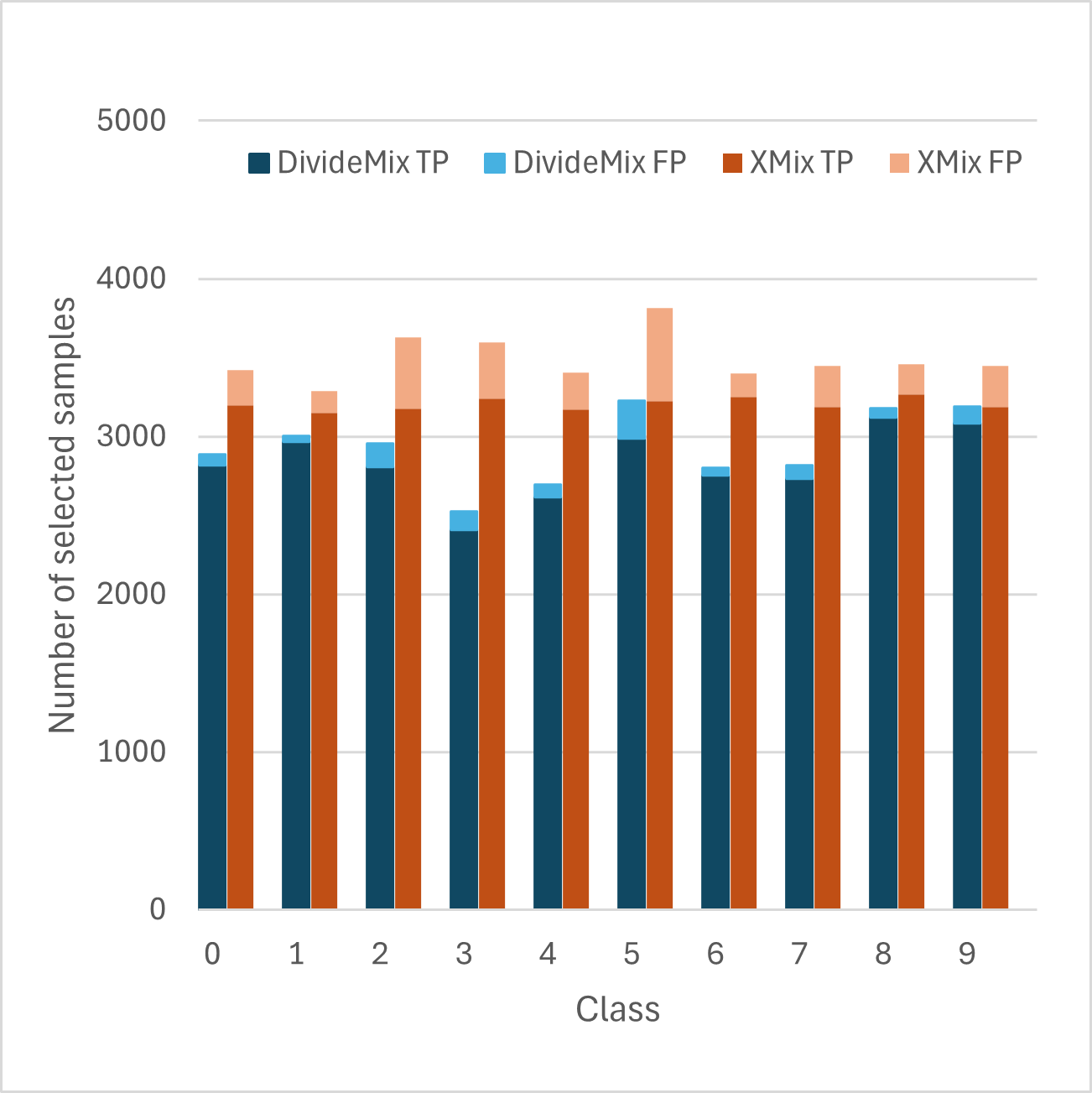}}
    {\includegraphics[width=\linewidth]{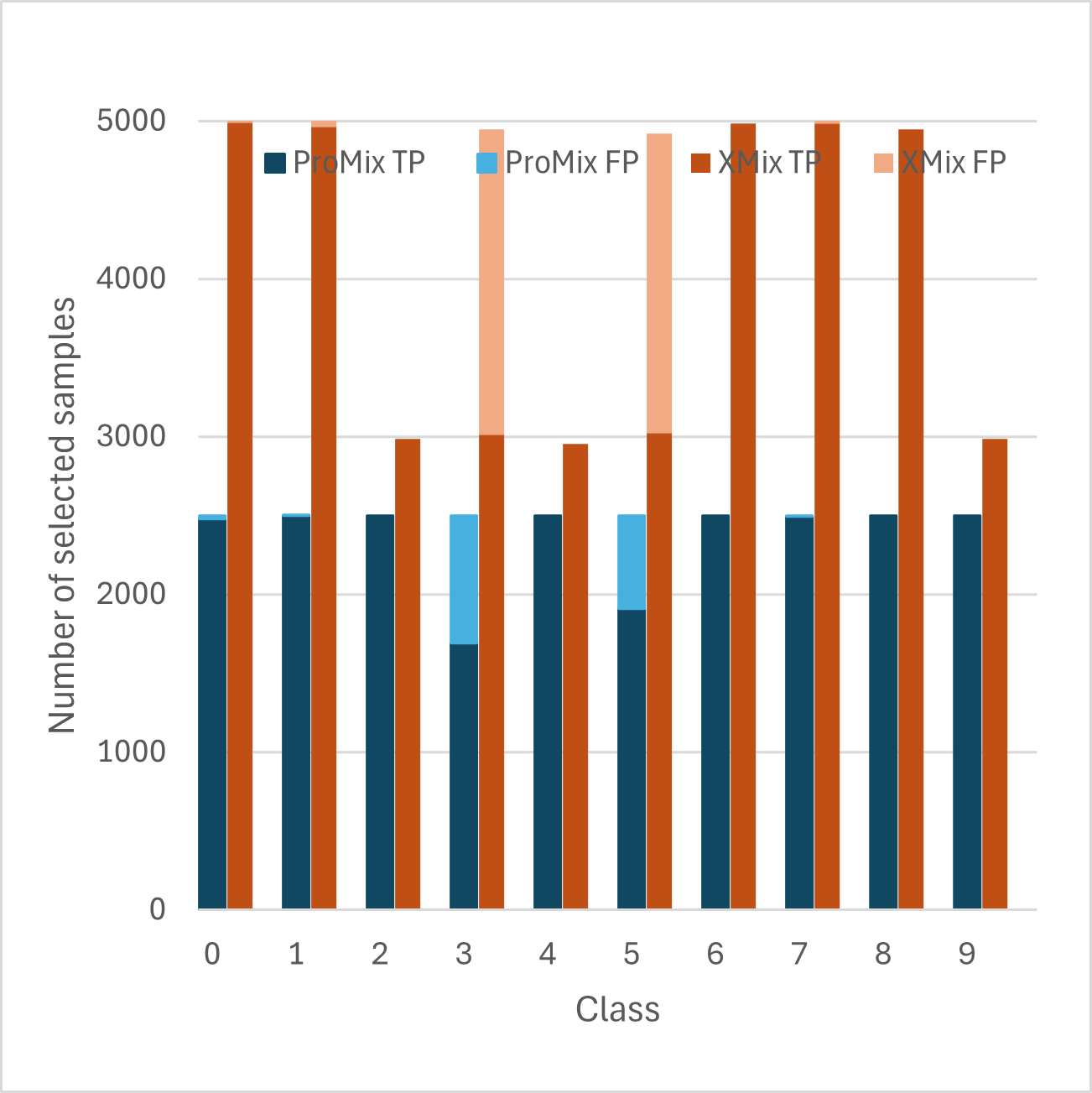}}
    \centerline{(c) 40\% Asym.}
\end{minipage}
\caption{Selected samples for each class at an early epoch of training on the CIFAR-10 dataset with different asymmetric label noise levels. \textbf{Upper Row:} DivideMix vs.\ XMix; \textbf{Bottom Row:} ProMix vs.\ XMix. TP/FP denotes true and false clean samples, shown in light and dark shades, respectively. Zoom in for details.}
\label{fig:selected_samples_asym_full}
\end{figure*}

\begin{figure*}[!htb]
\centering
\begin{minipage}[htb]{0.19\linewidth}
    \centering
    {\includegraphics[width=\linewidth]{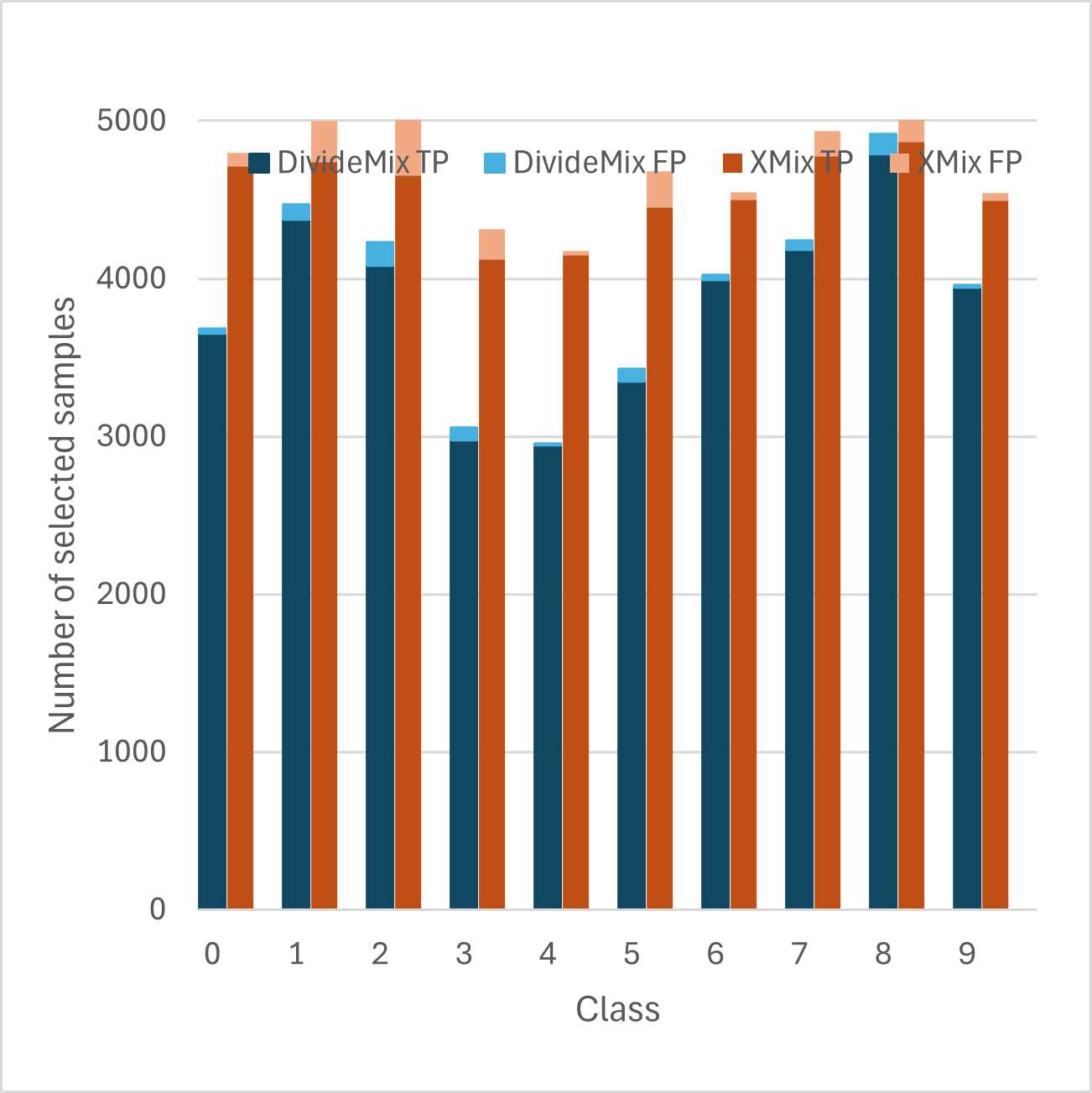}}
    {\includegraphics[width=\linewidth]{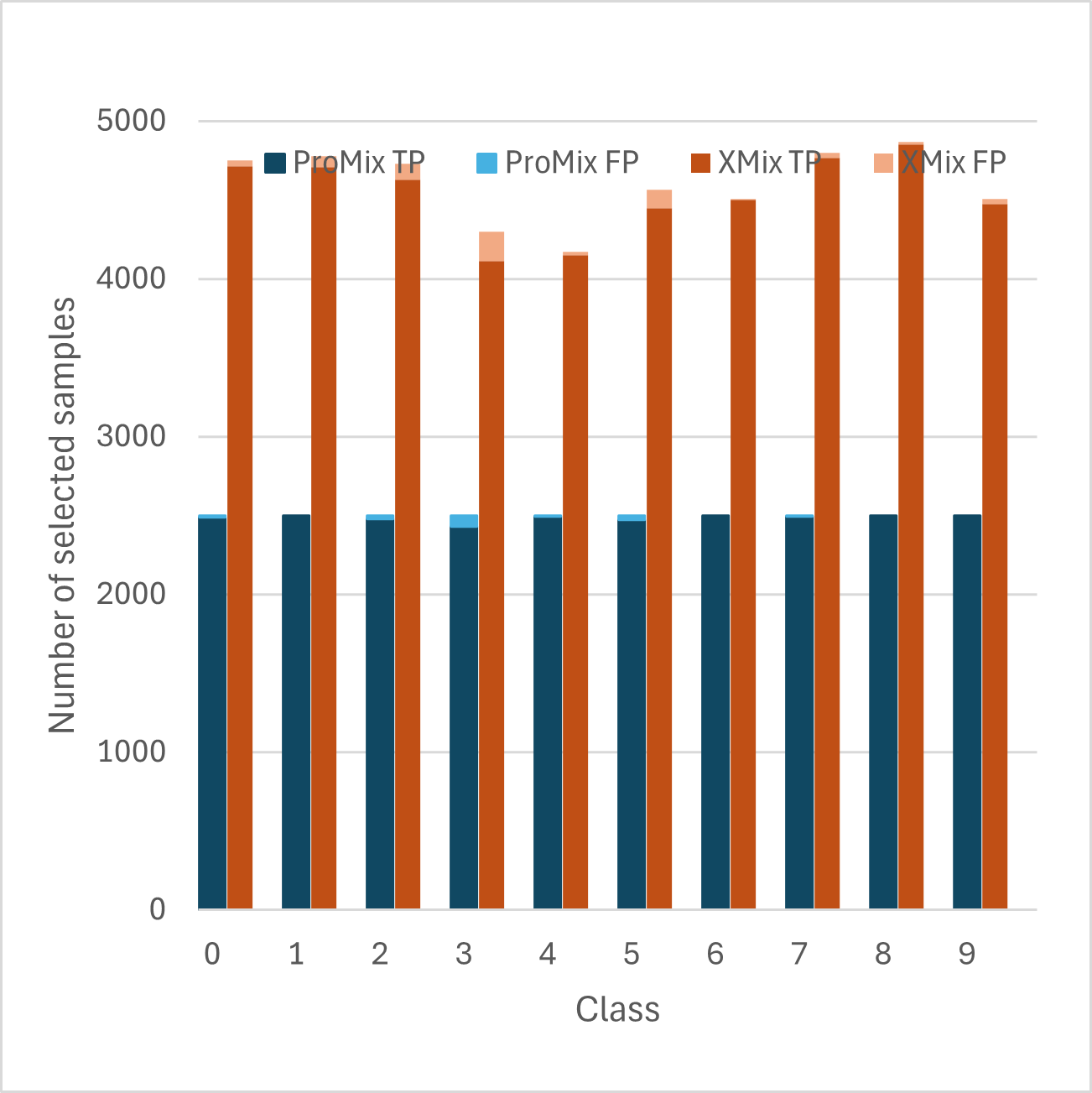}}
    \centerline{(a) Aggregate}
\end{minipage}
\begin{minipage}[htb]{0.19\linewidth}
    \centering
    {\includegraphics[width=\linewidth]{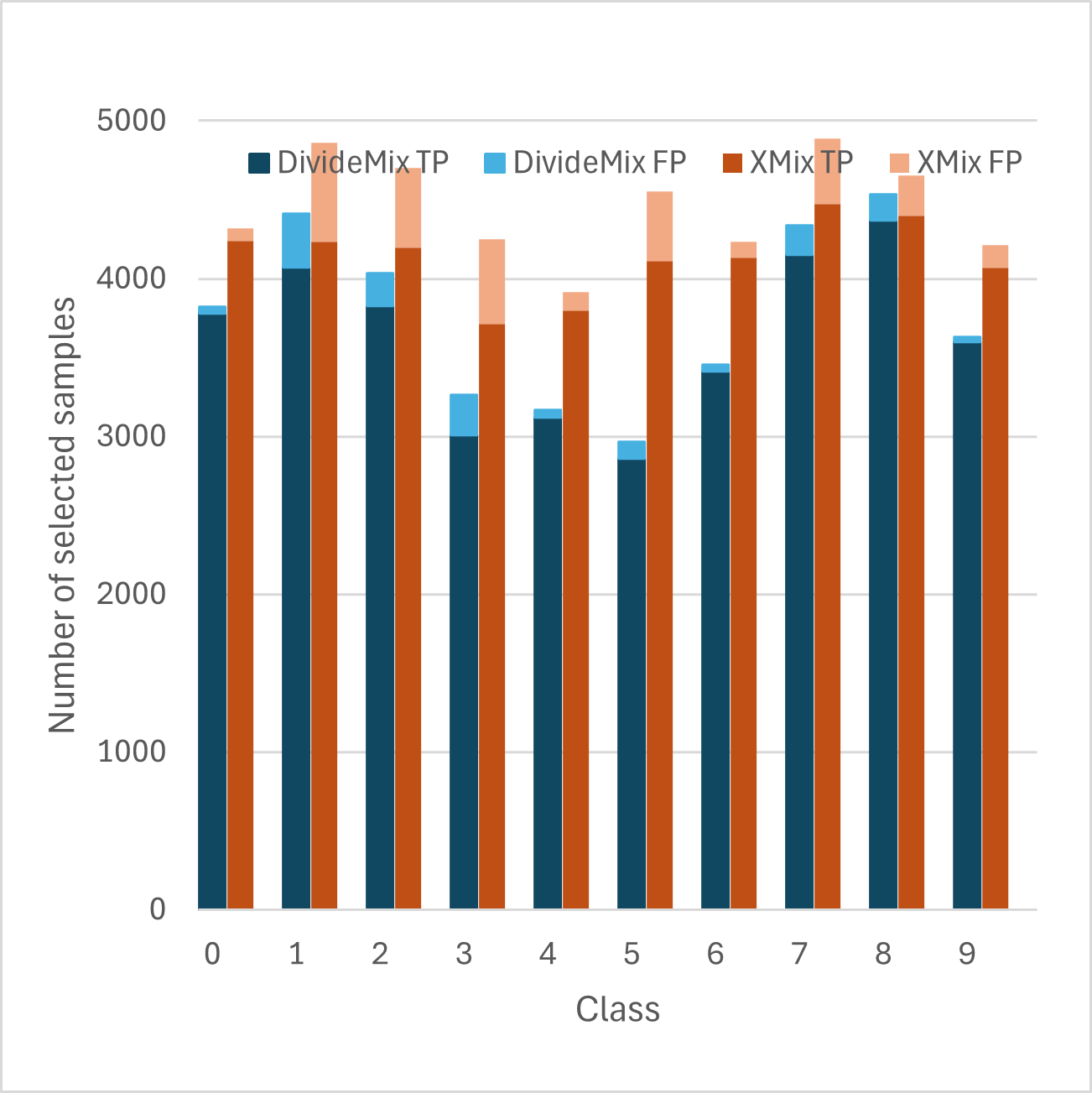}}
    {\includegraphics[width=\linewidth]{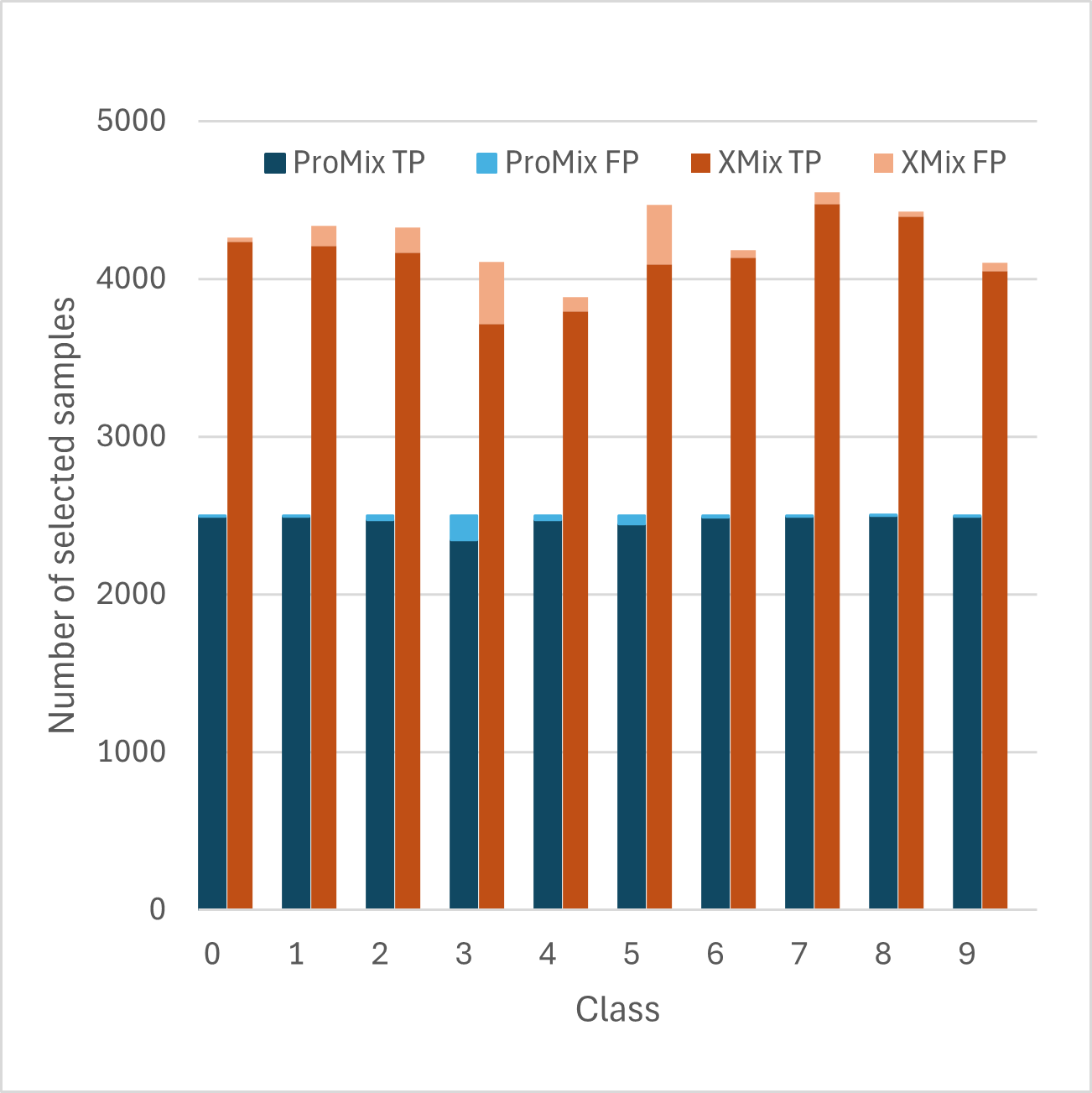}}
    \centerline{(b) Random 1}
\end{minipage}
\begin{minipage}[htb]{0.19\linewidth}
    \centering
    {\includegraphics[width=\linewidth]{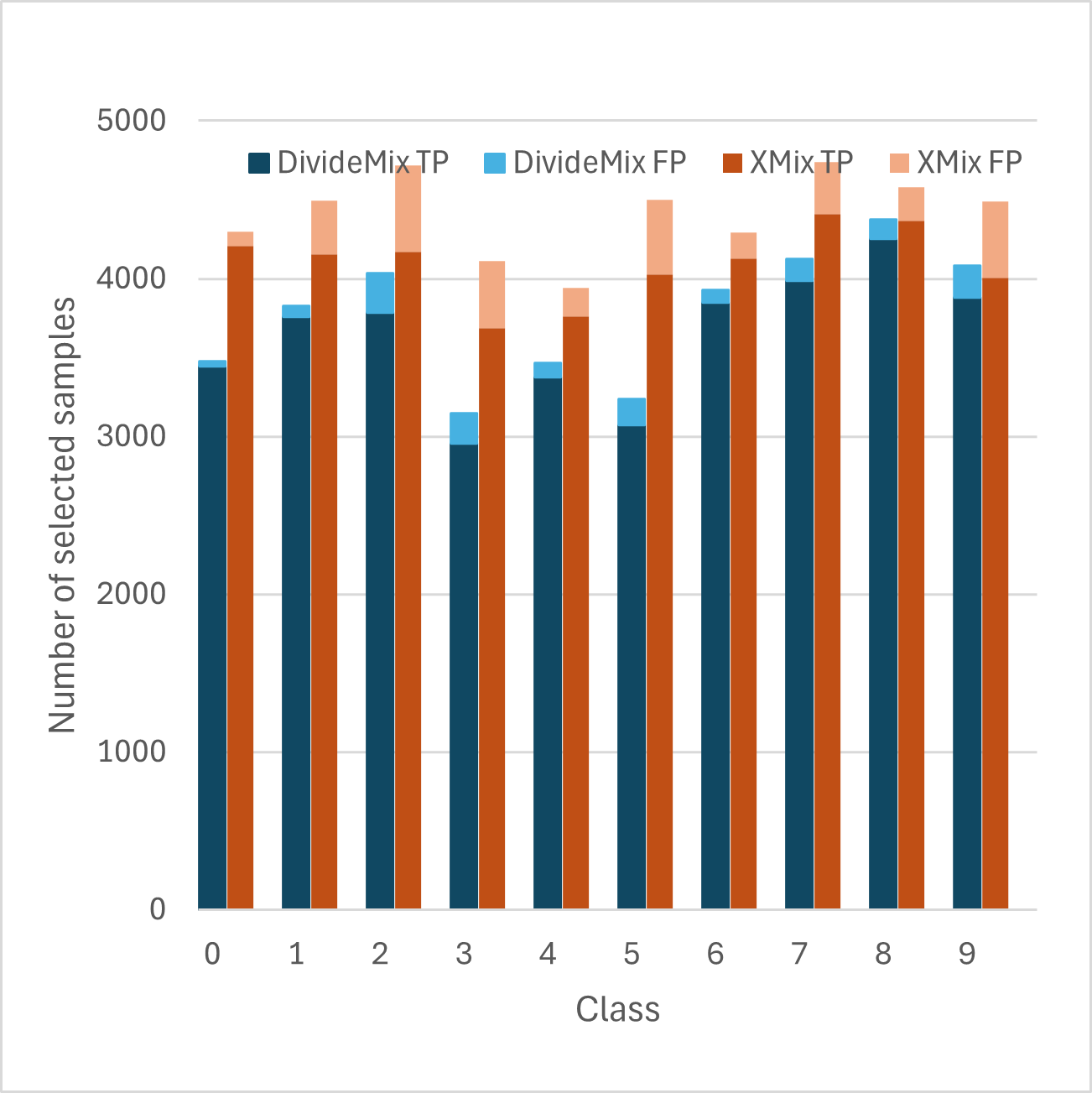}}
    {\includegraphics[width=\linewidth]{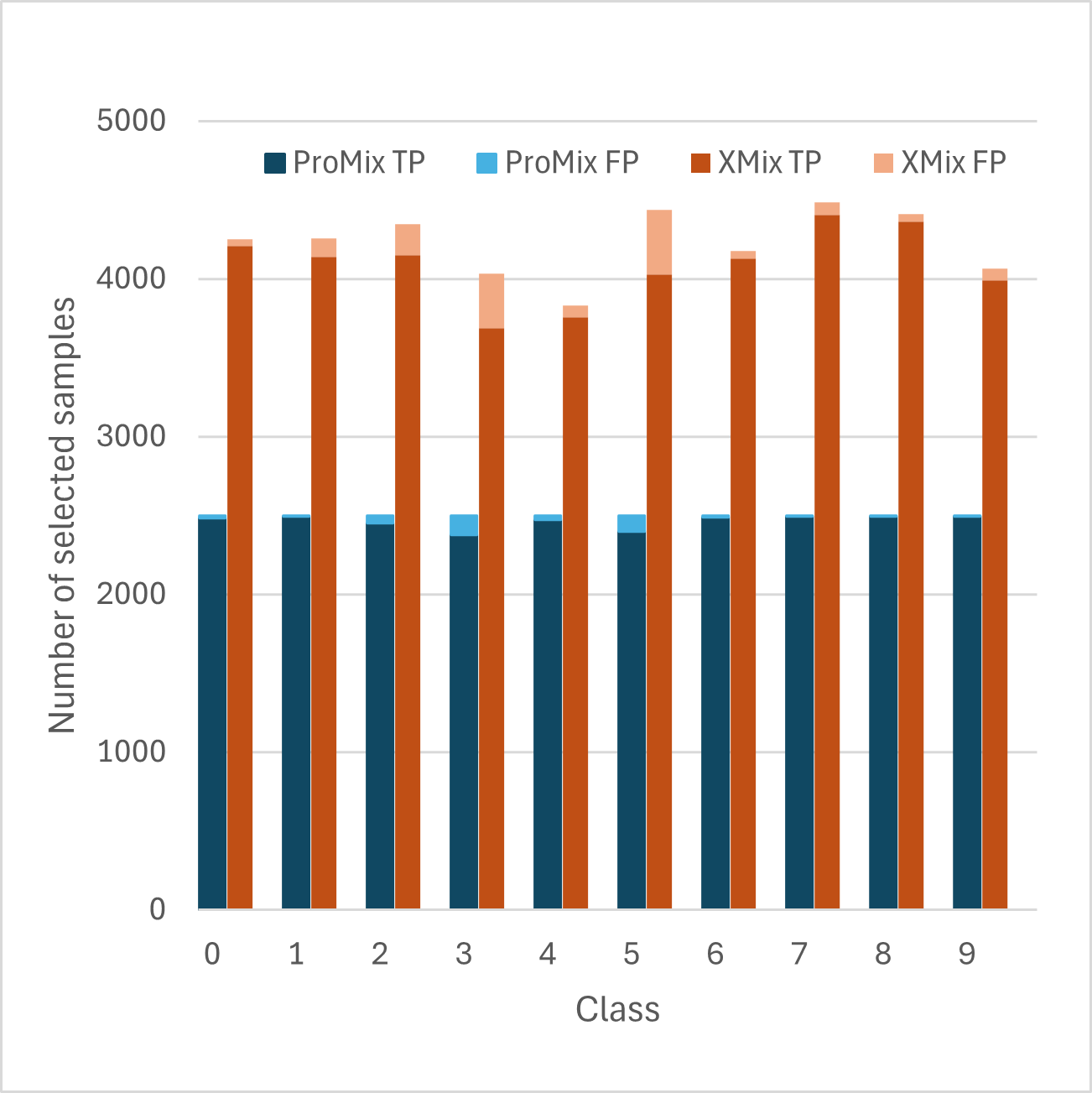}}
    \centerline{(c) Random 2}
\end{minipage}
\begin{minipage}[htb]{0.19\linewidth}
    \centering
    {\includegraphics[width=\linewidth]{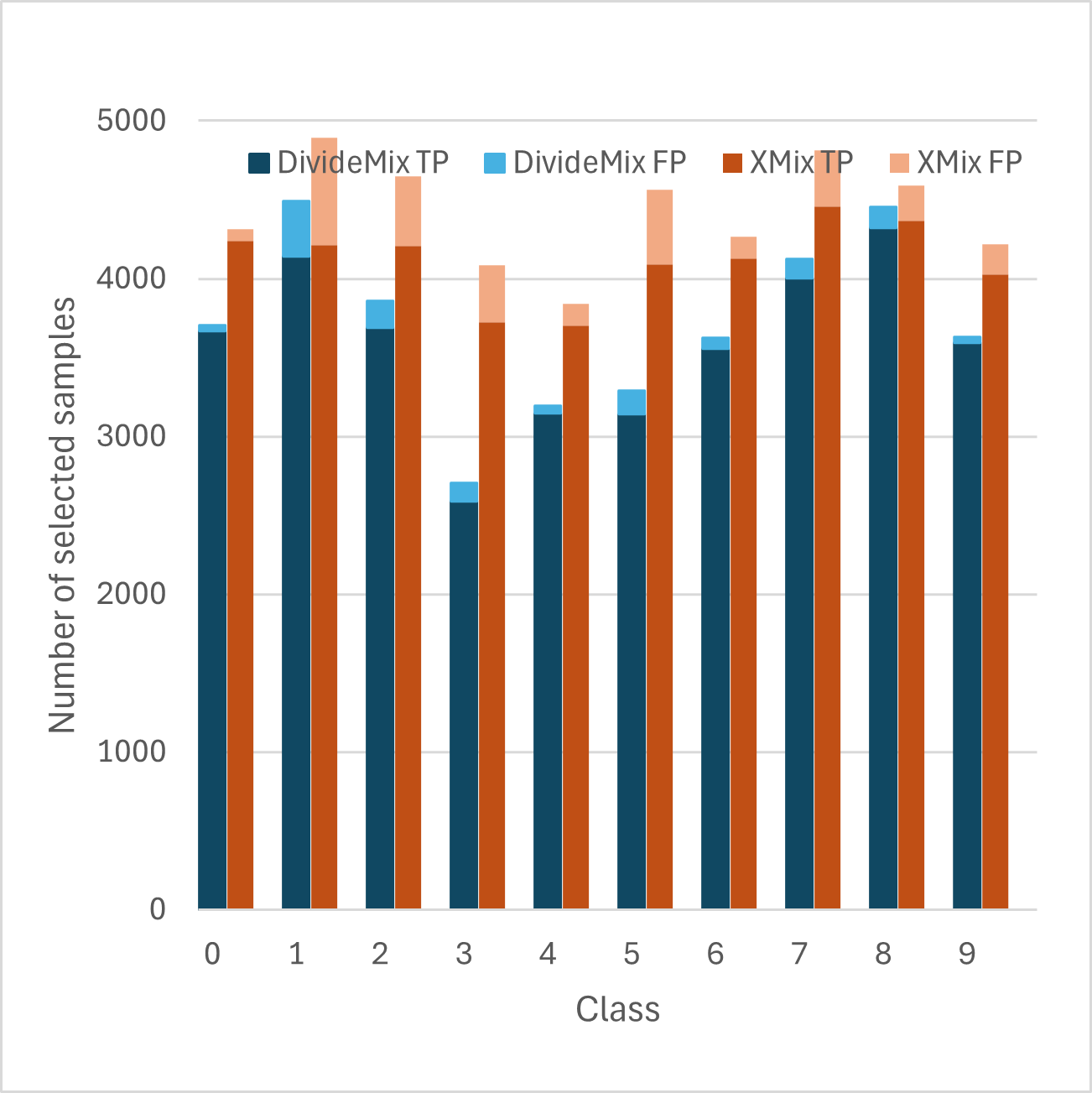}}
    {\includegraphics[width=\linewidth]{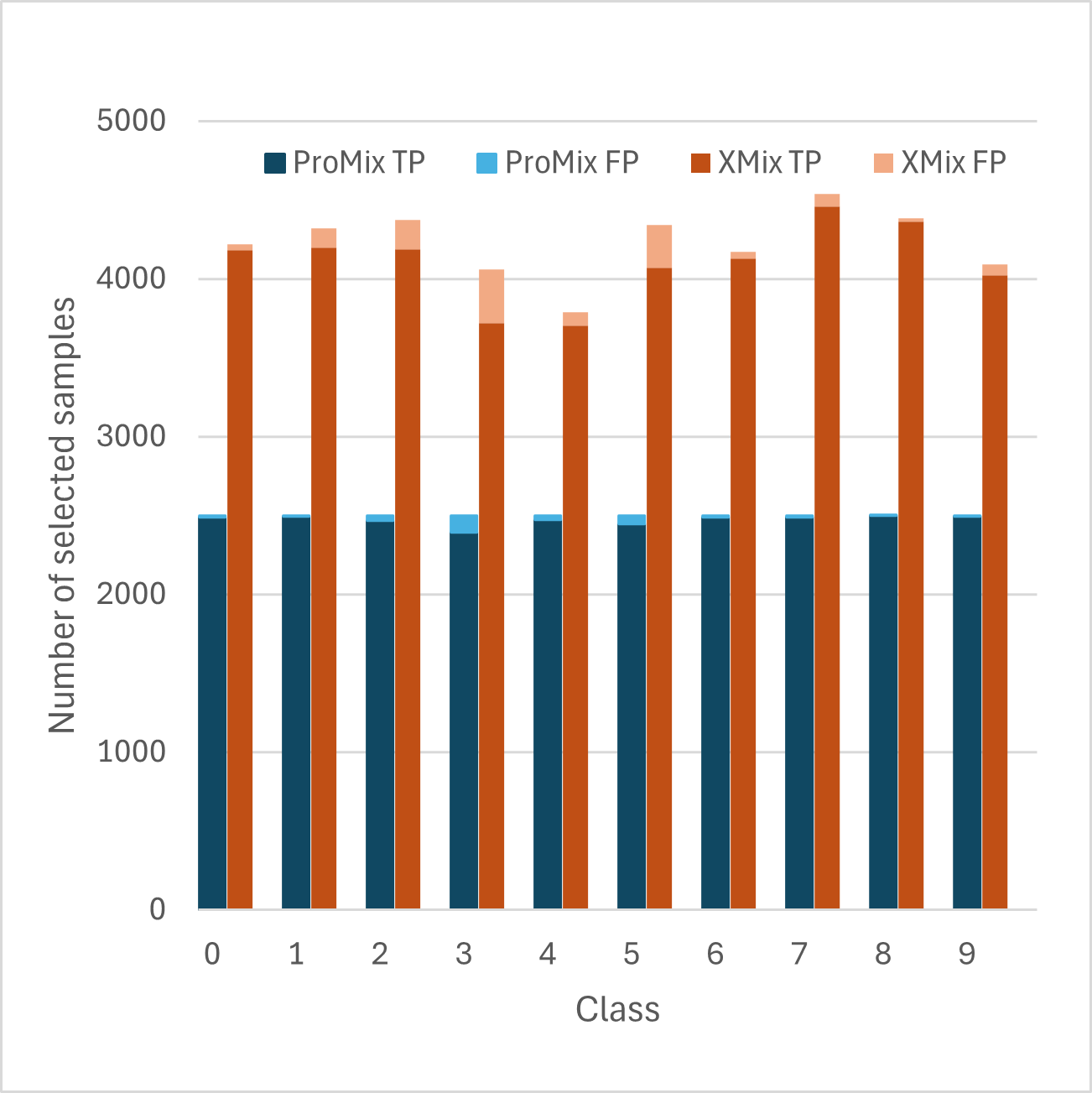}}
    \centerline{(d) Random 3}
\end{minipage}
\begin{minipage}[htb]{0.19\linewidth}
    \centering
    {\includegraphics[width=\linewidth]{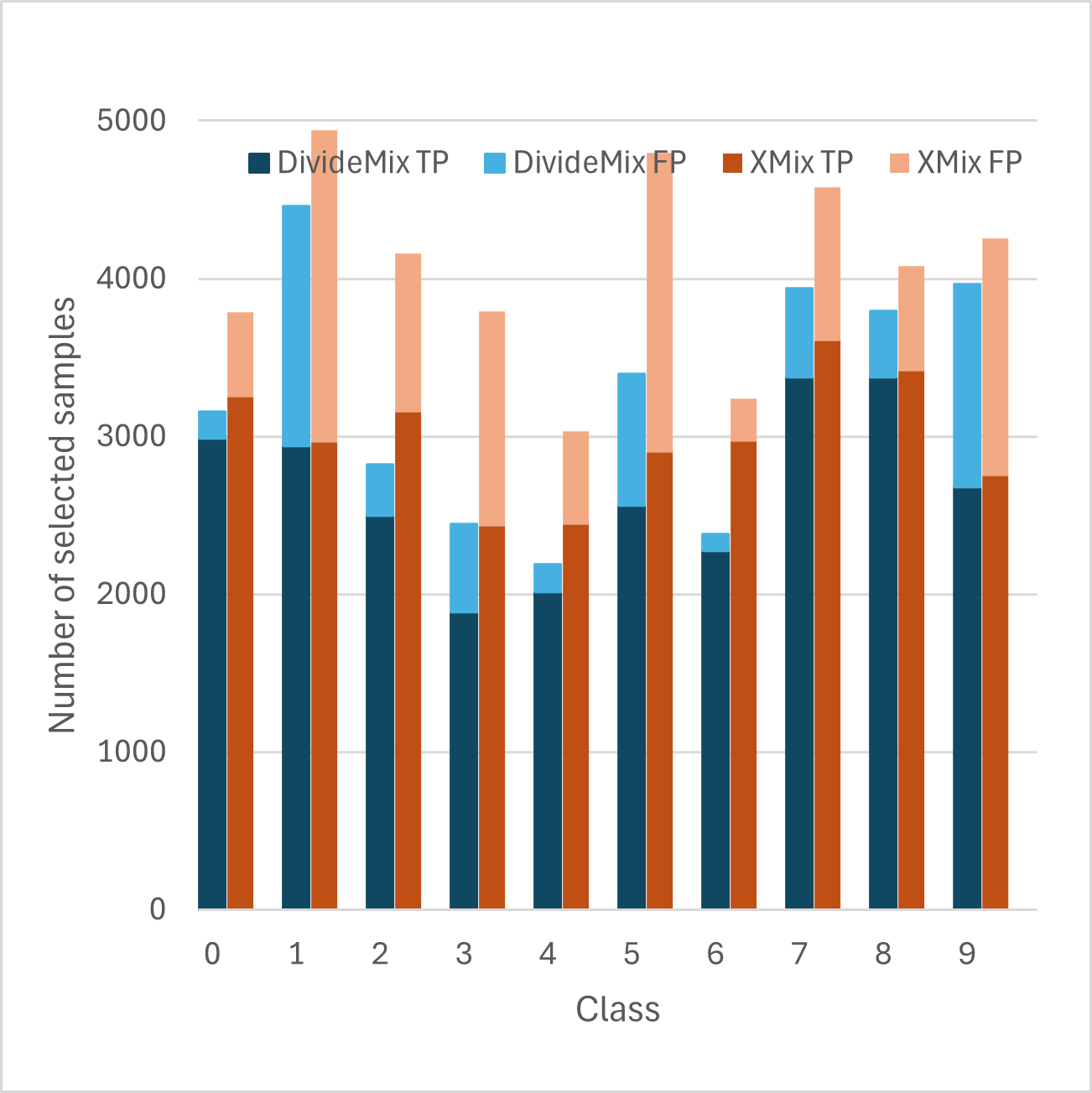}}
    {\includegraphics[width=\linewidth]{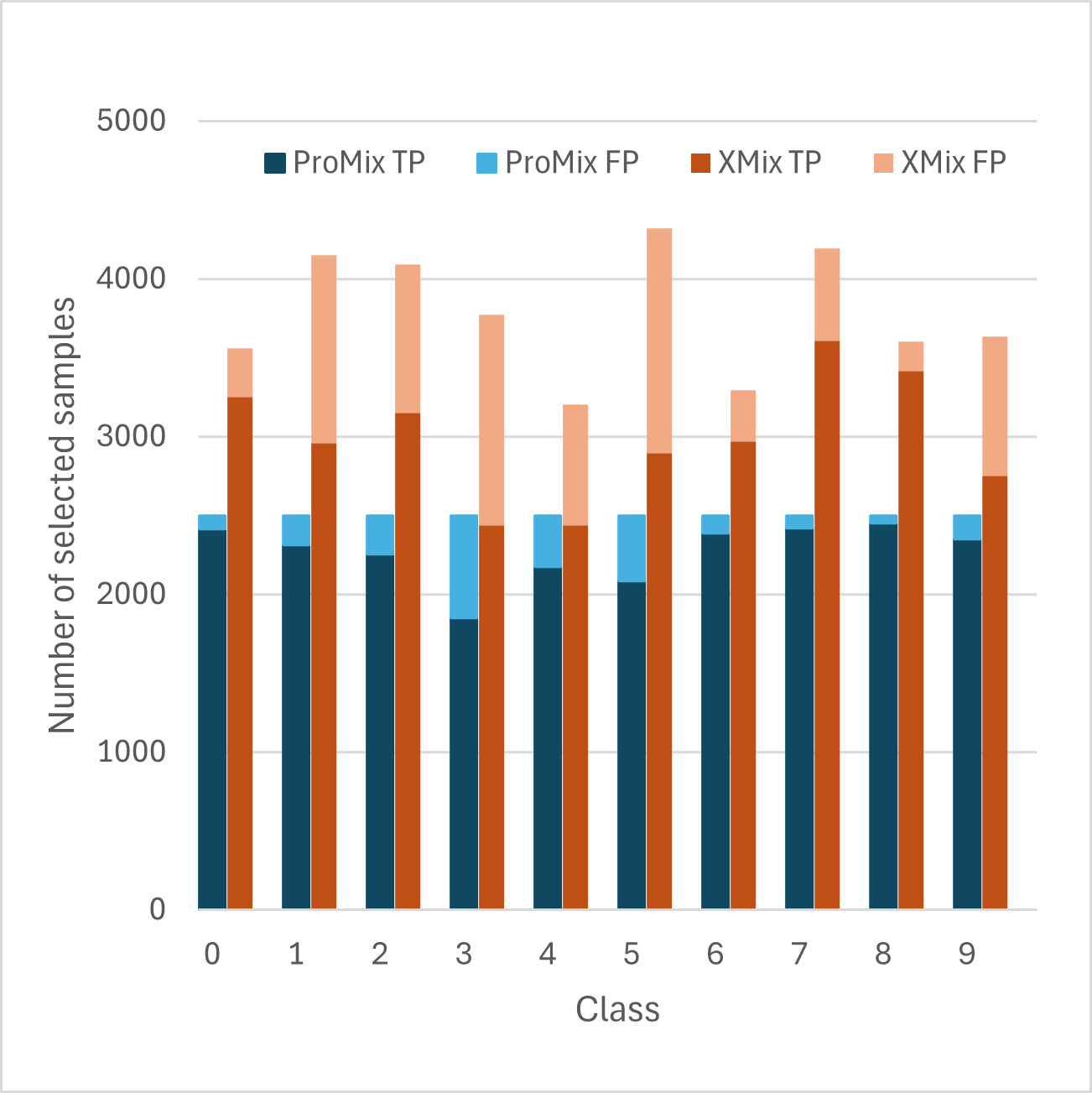}}
    \centerline{(e) Worst}
\end{minipage}
\caption{Selected samples for each class at an early epoch of training on CIFAR-N with real-world label noise. \textbf{Upper Row:} DivideMix vs.\ XMix; \textbf{Bottom Row:} ProMix vs.\ XMix. TP/FP denotes true and false clean samples, shown in light and dark shades, respectively. Zoom in for details.}
\label{fig:selected_samples_cifarn_full}
\end{figure*}

\noindent \textbf{Asymmetric and real-world label noise.} Figure~\ref{fig:selected_samples_asym_full} and Figure~\ref{fig:selected_samples_cifarn_full} show the corresponding sample selection results under asymmetric and CIFAR-N real-world label noise, respectively. In both settings, XMix selects noticeably more true clean samples than the base methods (DivideMix and ProMix), mirroring the trend observed under symmetric noise.

\begin{figure*}[!htb]
\centering
\begin{minipage}[htb]{0.23\linewidth}
    \centering
    {\includegraphics[width=\linewidth]{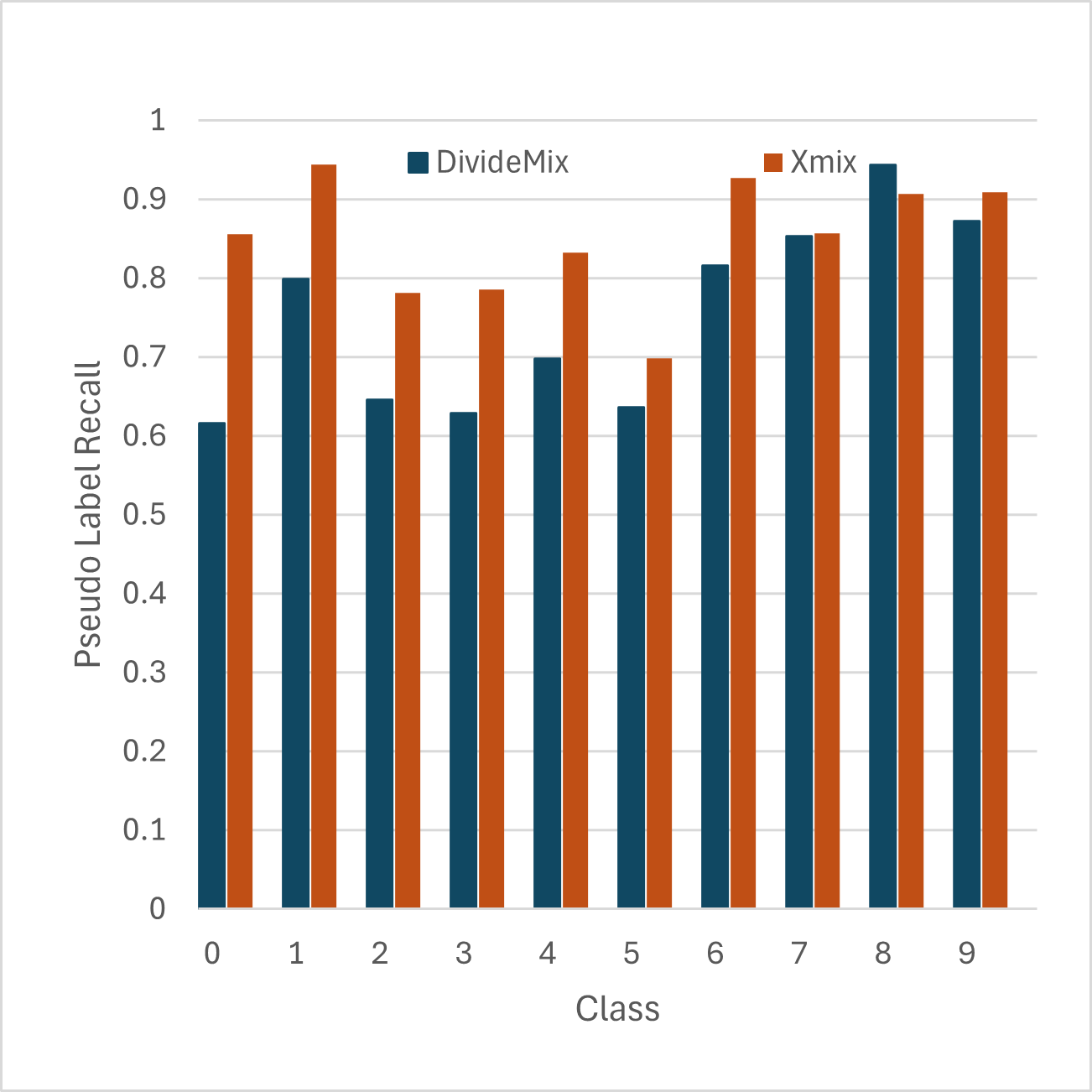}}
    {\includegraphics[width=\linewidth]{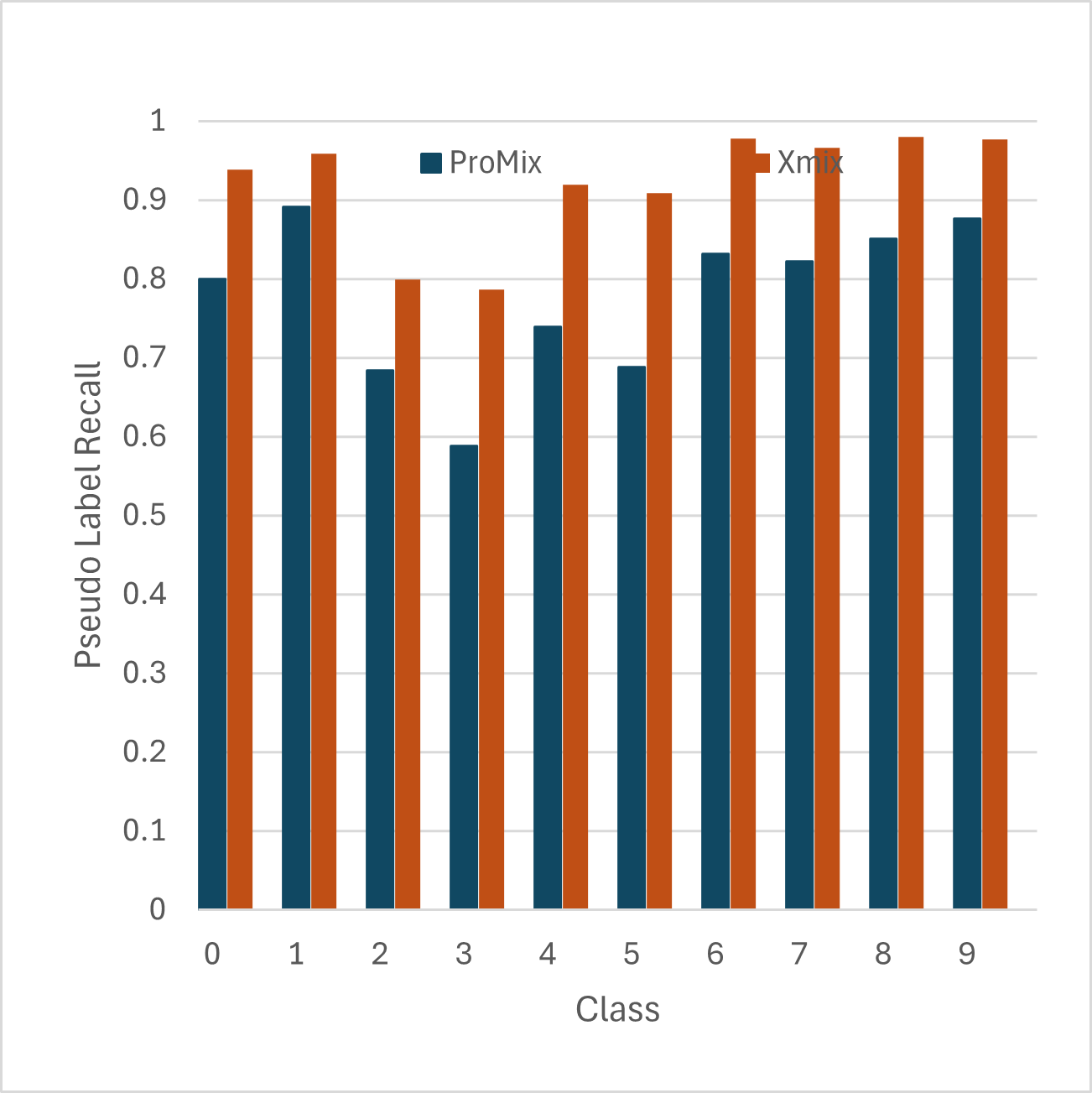}}
    \centerline{(a) 20\% Sym}
\end{minipage}
\begin{minipage}[htb]{0.23\linewidth}
    \centering
    {\includegraphics[width=\linewidth]{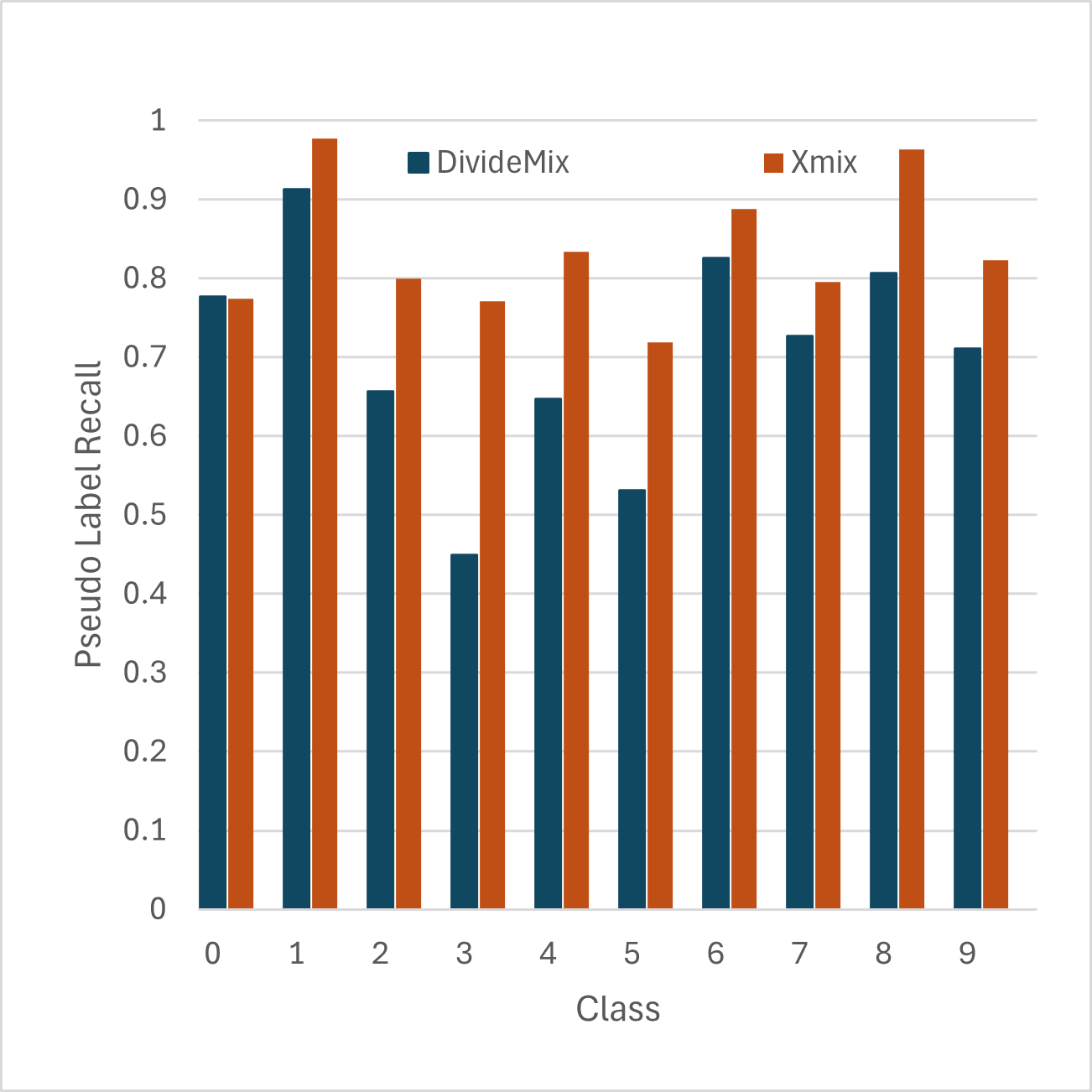}}
    {\includegraphics[width=\linewidth]{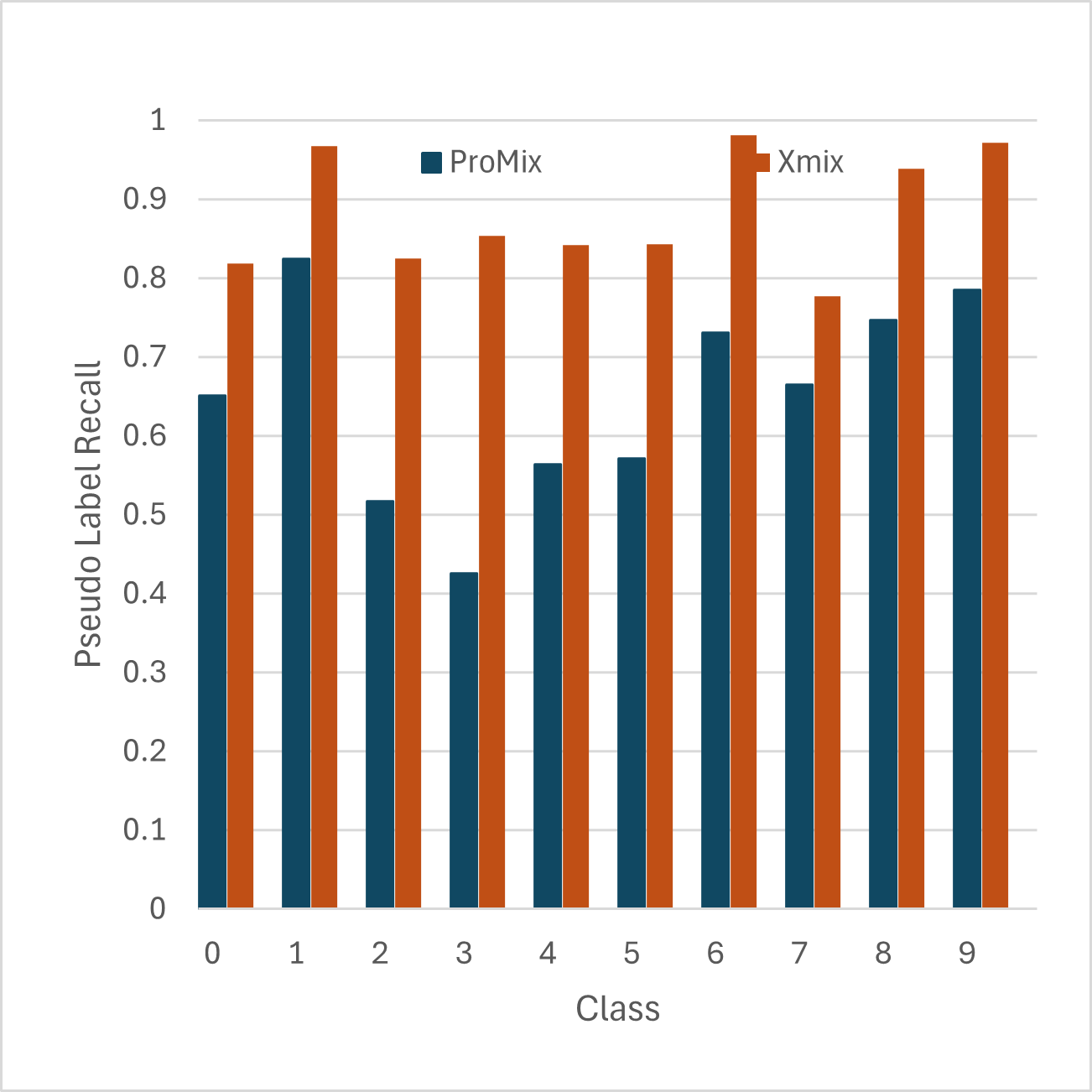}}
    \centerline{(b) 50\% Sym.}
\end{minipage}
\begin{minipage}[htb]{0.23\linewidth}
    \centering
    {\includegraphics[width=\linewidth]{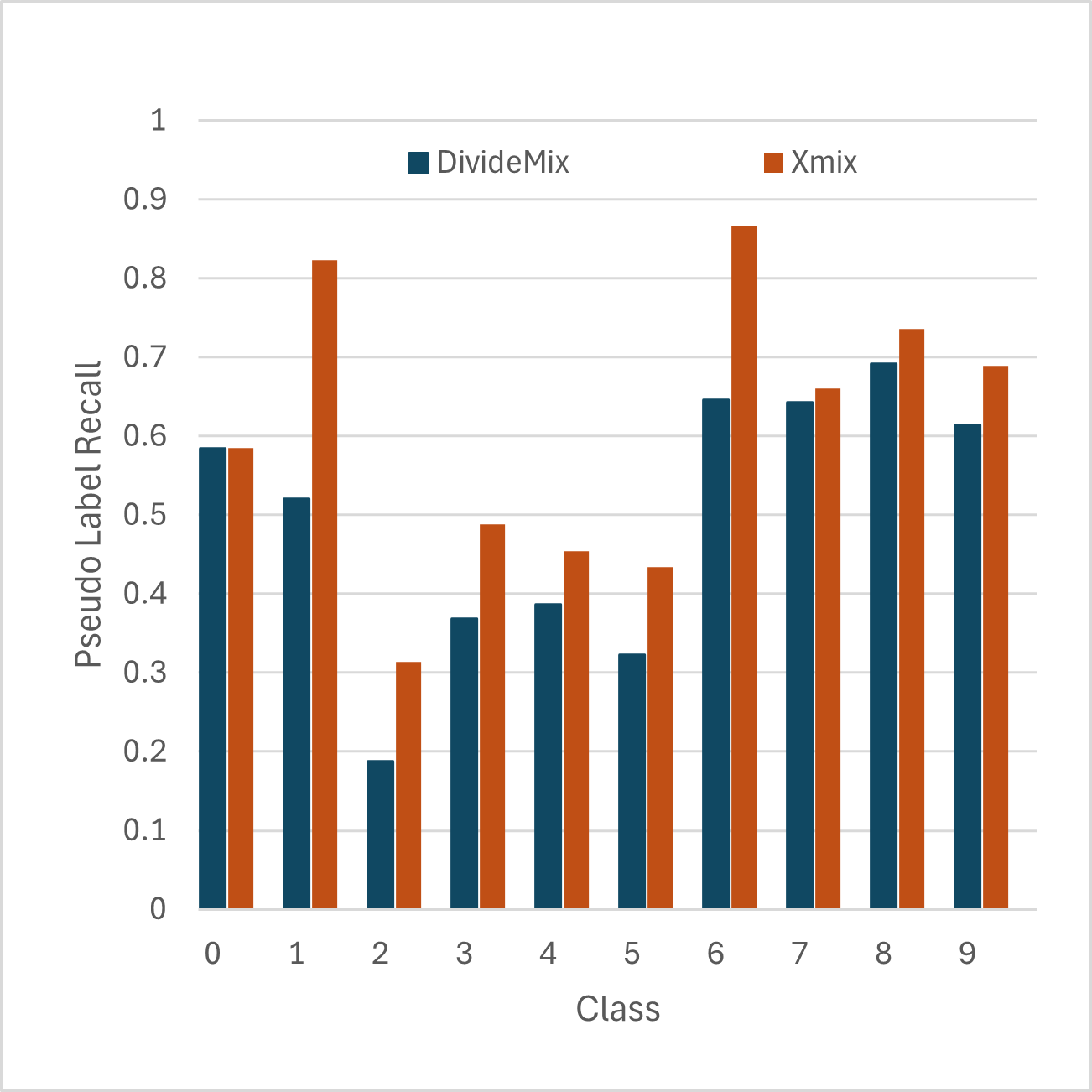}}
    {\includegraphics[width=\linewidth]{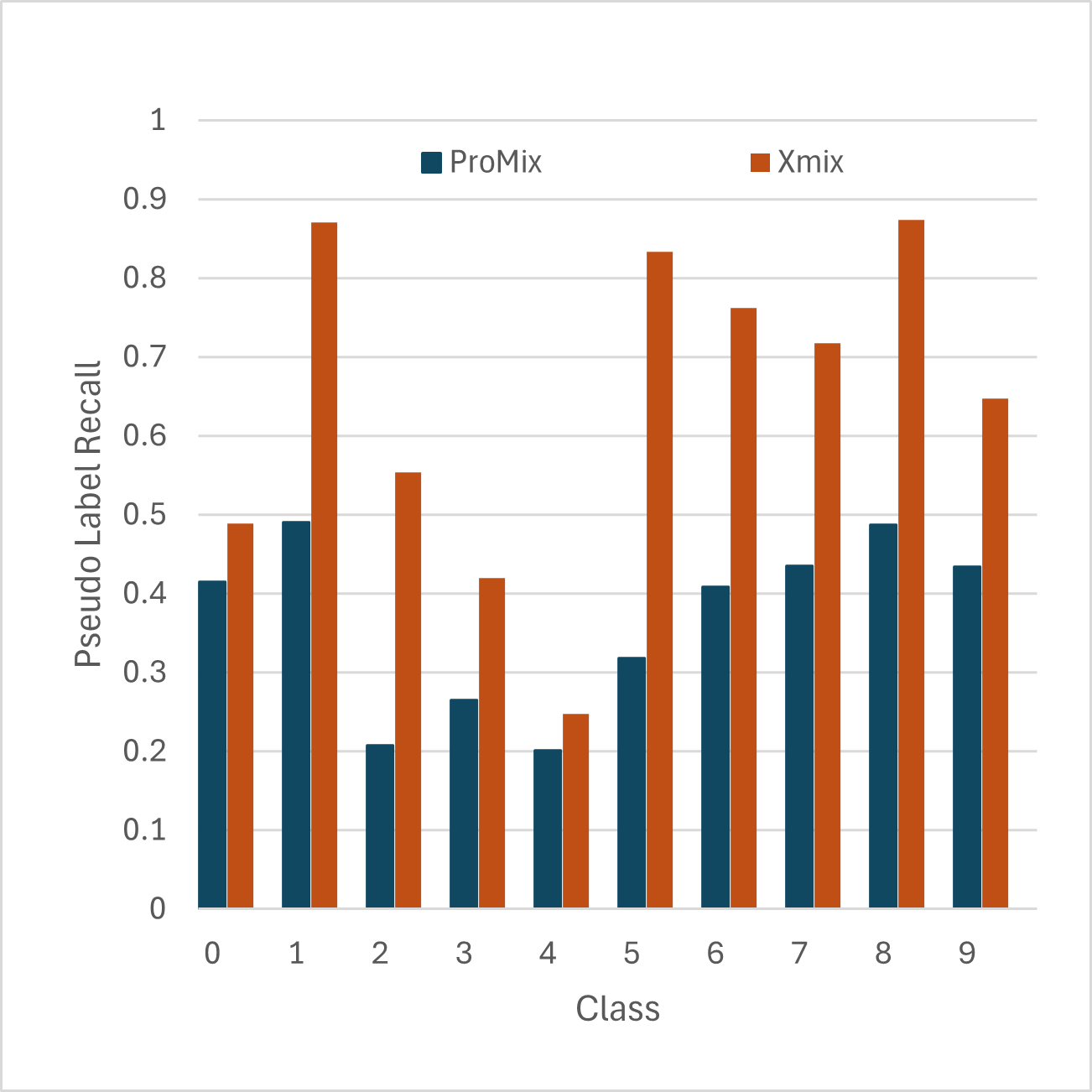}}
    \centerline{(c) 80\% Sym.}
\end{minipage}
\begin{minipage}[htb]{0.23\linewidth}
    \centering
    {\includegraphics[width=\linewidth]{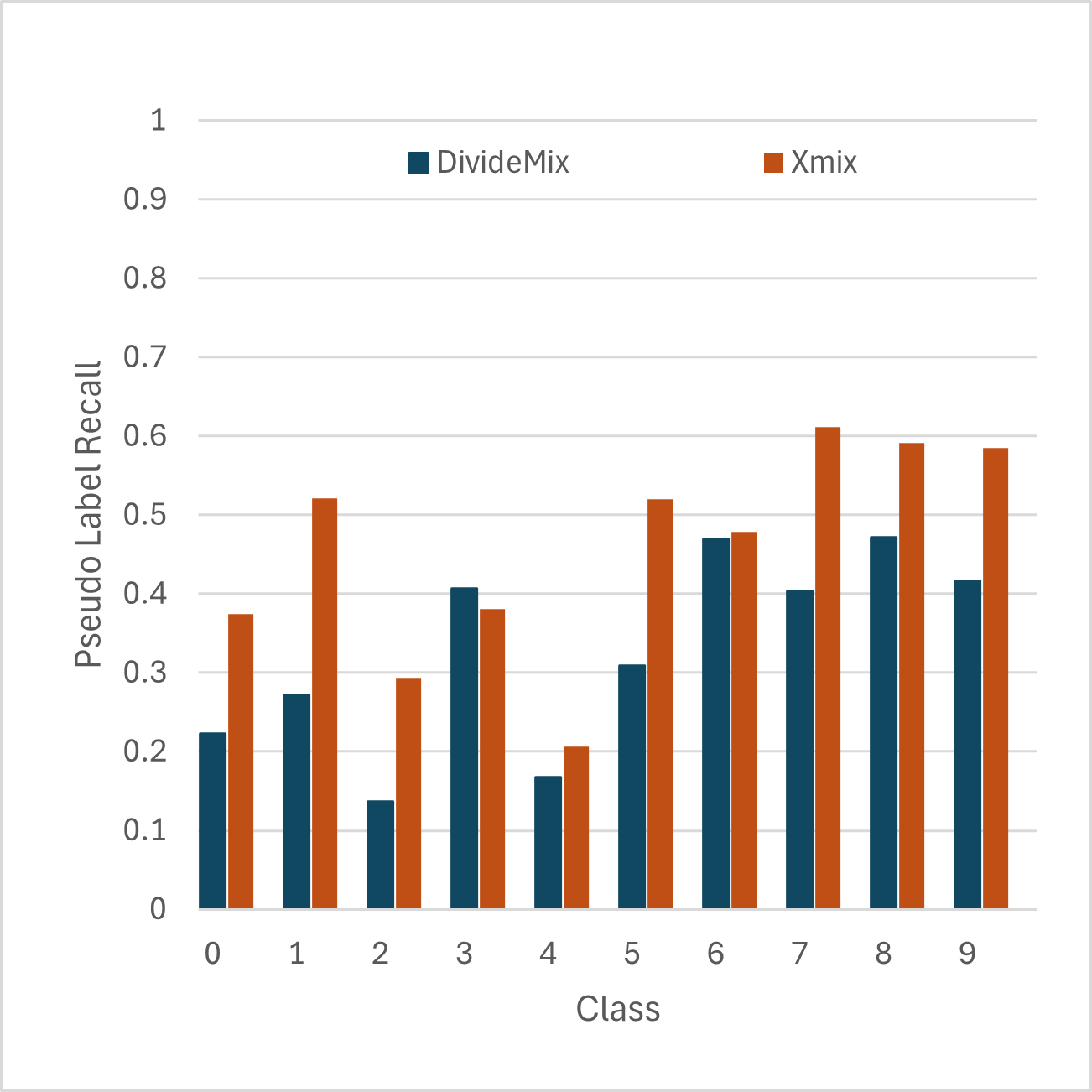}}
    {\includegraphics[width=\linewidth]{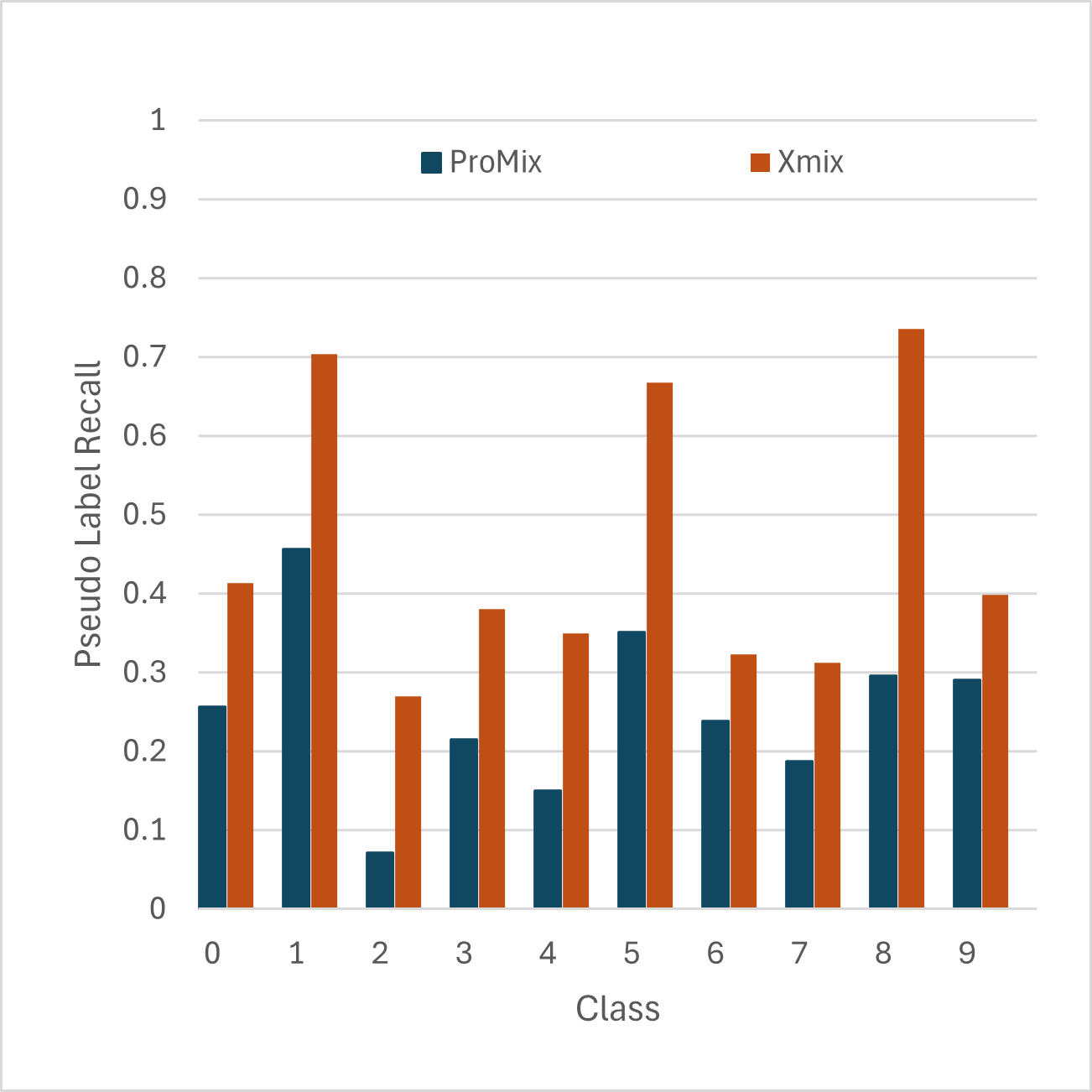}}
    \centerline{(d) 90\% Sym.}
\end{minipage}
\caption{Pseudo-label recall at an early epoch of training on the CIFAR-10 dataset with different noise settings. \textbf{Upper Row:} DivideMix vs.\ XMix (DivideMix+); \textbf{Bottom Row:} ProMix vs.\ XMix (ProMix+). Zoom in for details.}
\label{fig:pseudo_label_recall}
\end{figure*}

\noindent \textbf{Pseudo-label quality.} Figure~\ref{fig:pseudo_label_recall} reports pseudo-label recall at an early epoch of training on CIFAR-10 across the same symmetric noise levels. Nearest neighbor pseudo-labeling consistently produces more accurate pseudo-labels for the unlabeled data than the base models, strengthening the semi-supervised learning phase.

\begin{figure*}[!htb]
\centering
\begin{minipage}[htb]{0.24\linewidth}
    \centering
    {\includegraphics[width=\linewidth]{images/dividemixvsxmix_20.png}}
    {\includegraphics[width=\linewidth]{images/promixvsxmix_20.png}}
    \centerline{(a) 20\% Sym}
\end{minipage}
\begin{minipage}[htb]{0.24\linewidth}
    \centering
    {\includegraphics[width=\linewidth]{images/dividemixvsxmix_50.png}}
    {\includegraphics[width=\linewidth]{images/promixvsxmix_50.png}}
    \centerline{(b) 50\% Sym}
\end{minipage}
\begin{minipage}[htb]{0.24\linewidth}
    \centering
    {\includegraphics[width=\linewidth]{images/dividemixvsxmix_80.png}}
    {\includegraphics[width=\linewidth]{images/promixvsxmix_80.png}}
    \centerline{(c) 80\% Sym}
\end{minipage}
\begin{minipage}[htb]{0.24\linewidth}
    \centering
    {\includegraphics[width=\linewidth]{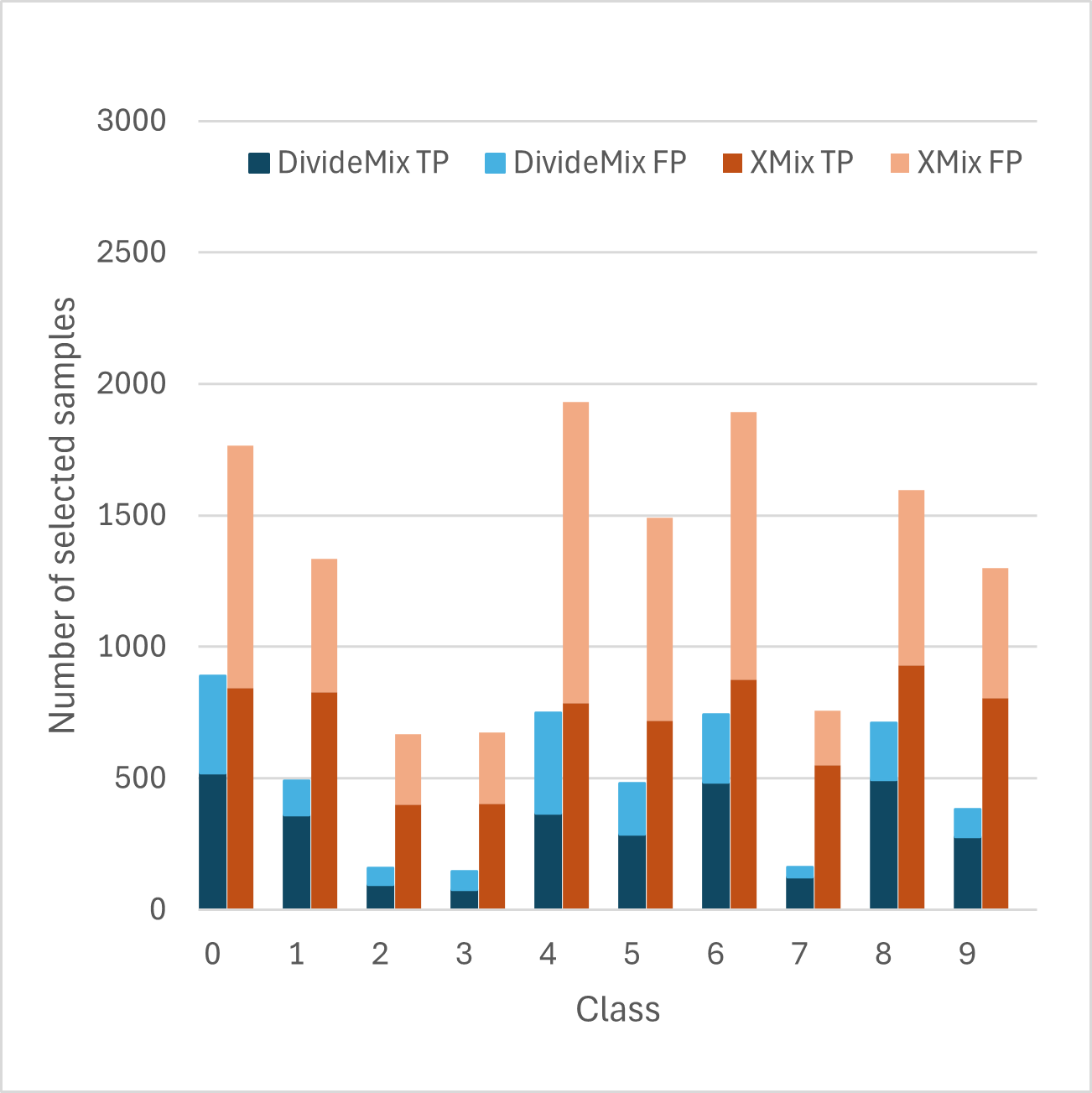}}
    {\includegraphics[width=\linewidth]{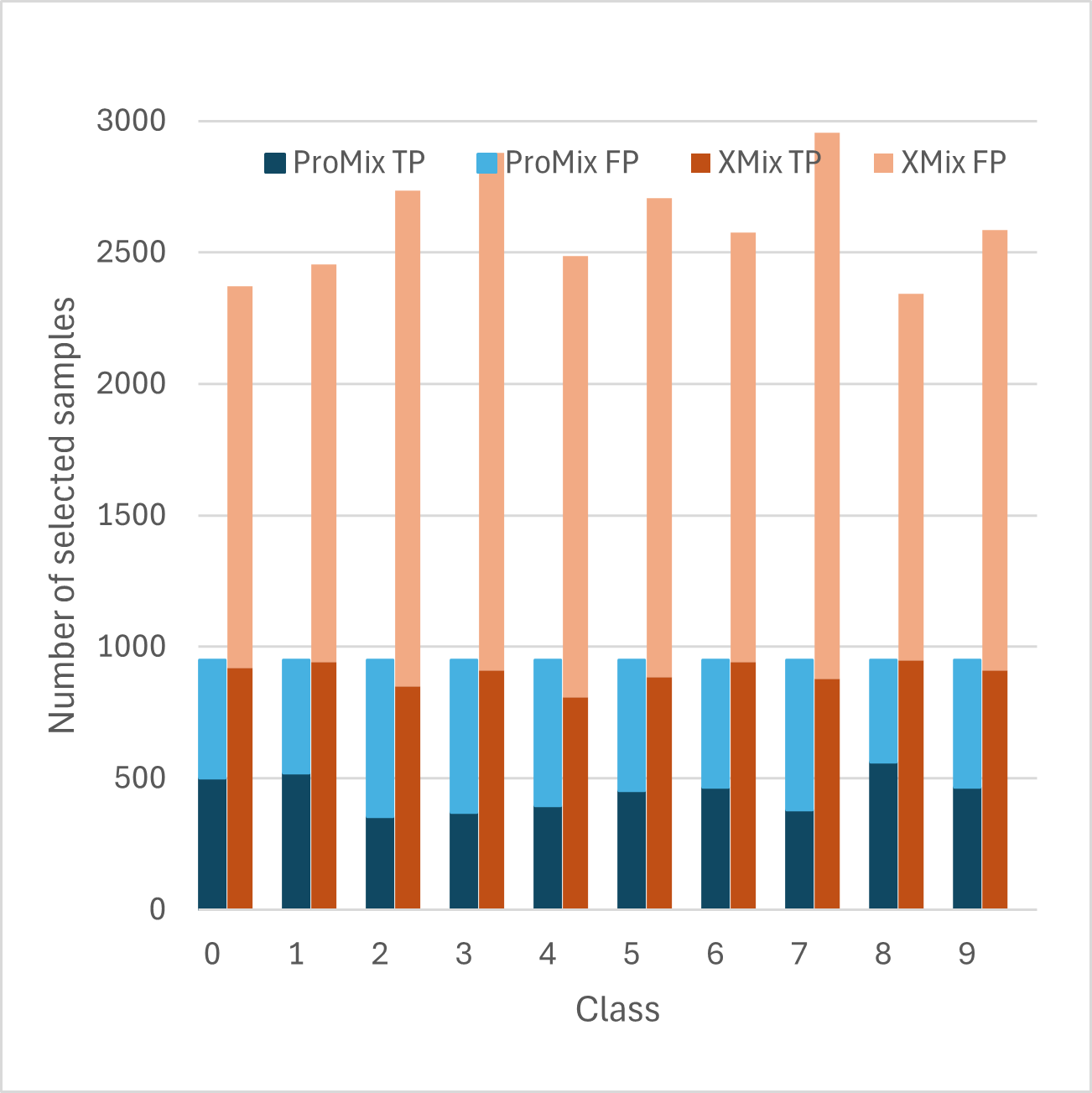}}
    \centerline{(d) 90\% Sym}
\end{minipage}
\caption{Selected samples for each class at an early epoch of training on the CIFAR-10 dataset with different symmetric label noise levels. XMix selects noticeably more true clean samples as noise increases. \textbf{Upper Row:} DivideMix vs.\ XMix; \textbf{Bottom Row:} ProMix vs.\ XMix. TP/FP denotes true and false clean samples, shown in light and dark shades, respectively. Zoom in for details.}
\label{fig:selected_samples_noise_full}
\end{figure*}

\begin{figure*}[!htb]
\centering
\begin{minipage}[htb]{0.24\linewidth}
    \centering
    {\includegraphics[width=\linewidth]{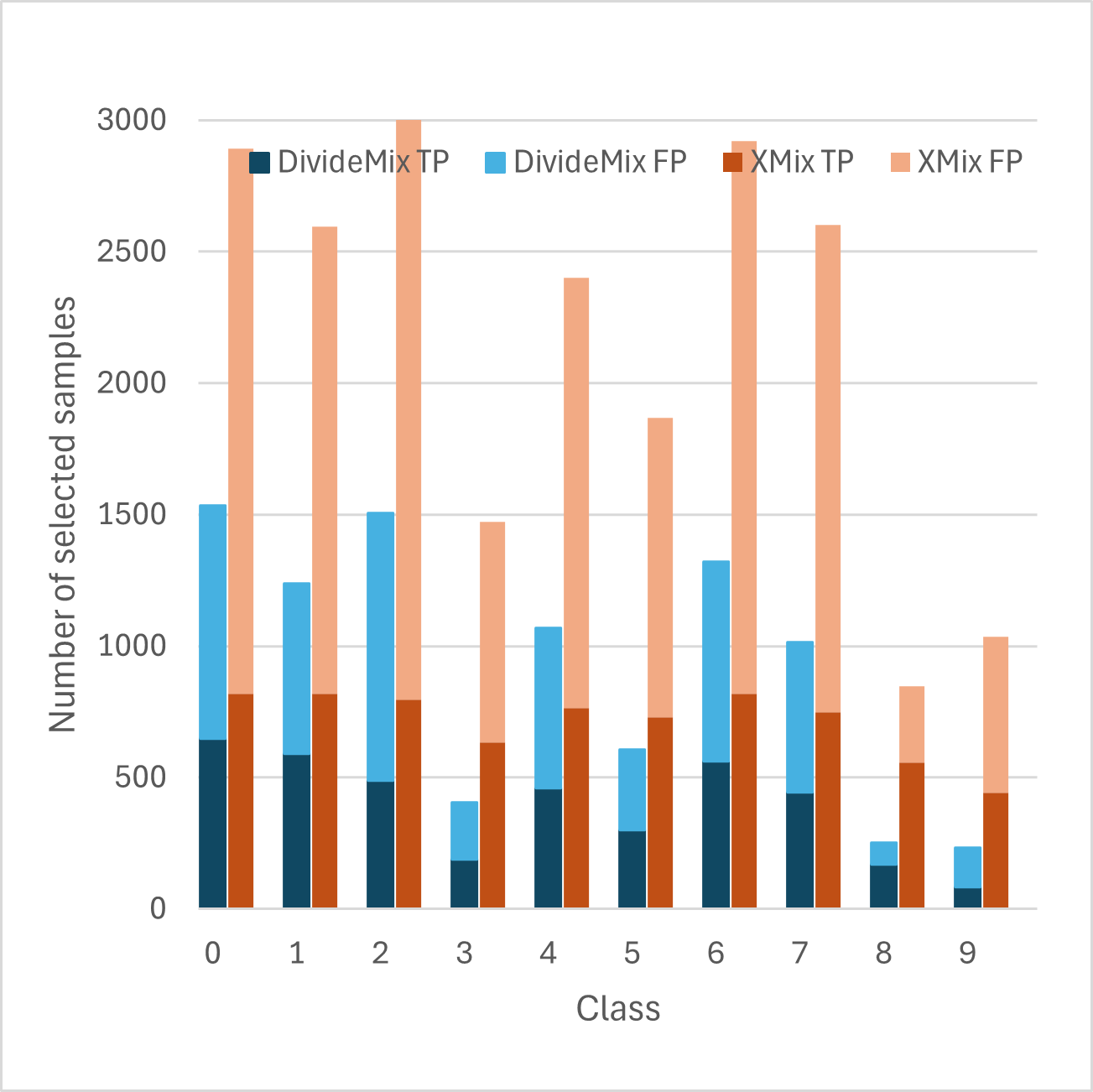}}
    {\includegraphics[width=\linewidth]{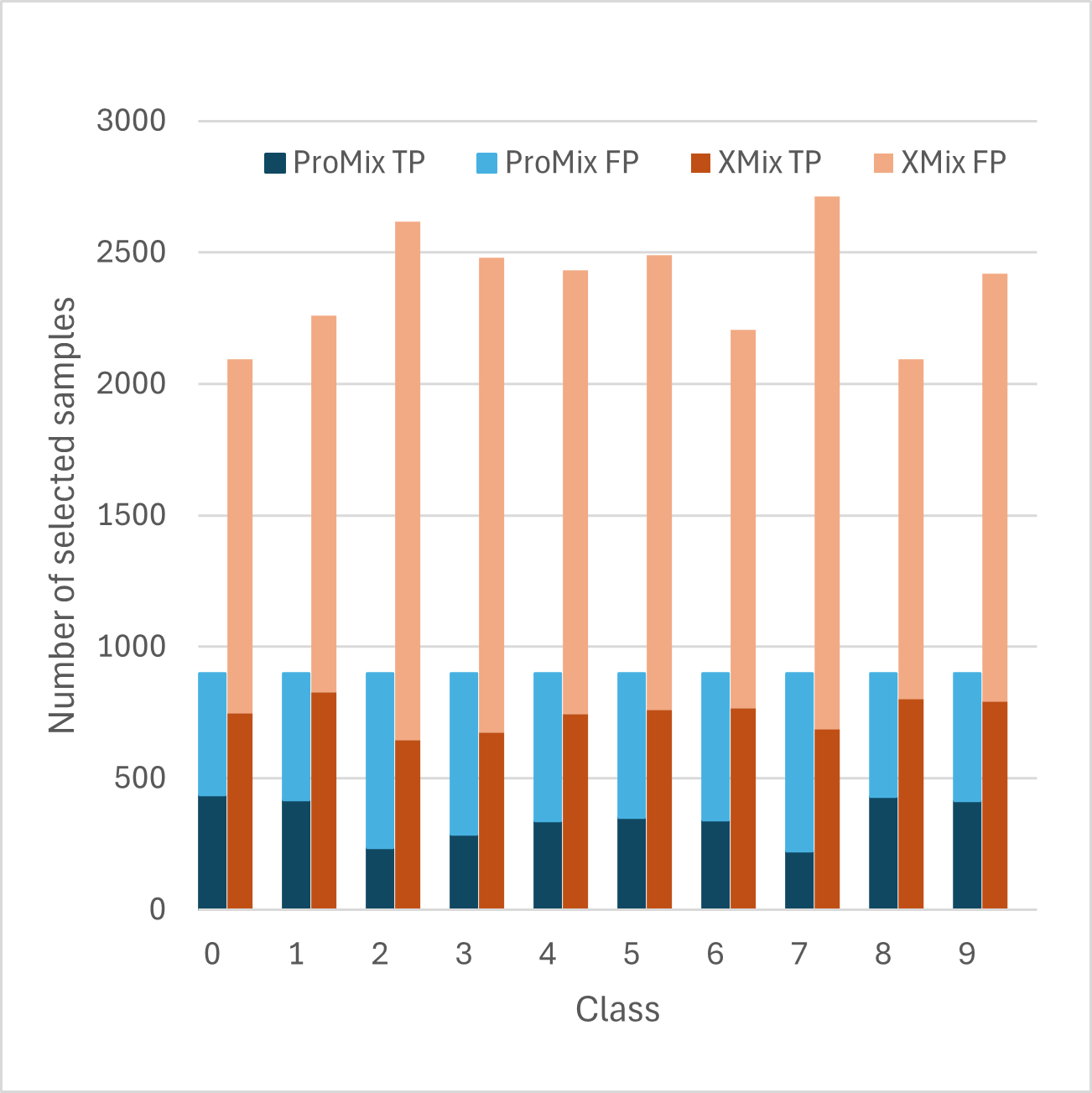}}
    \centerline{(e) 92\% Sym}
\end{minipage}
\begin{minipage}[htb]{0.24\linewidth}
    \centering
    {\includegraphics[width=\linewidth]{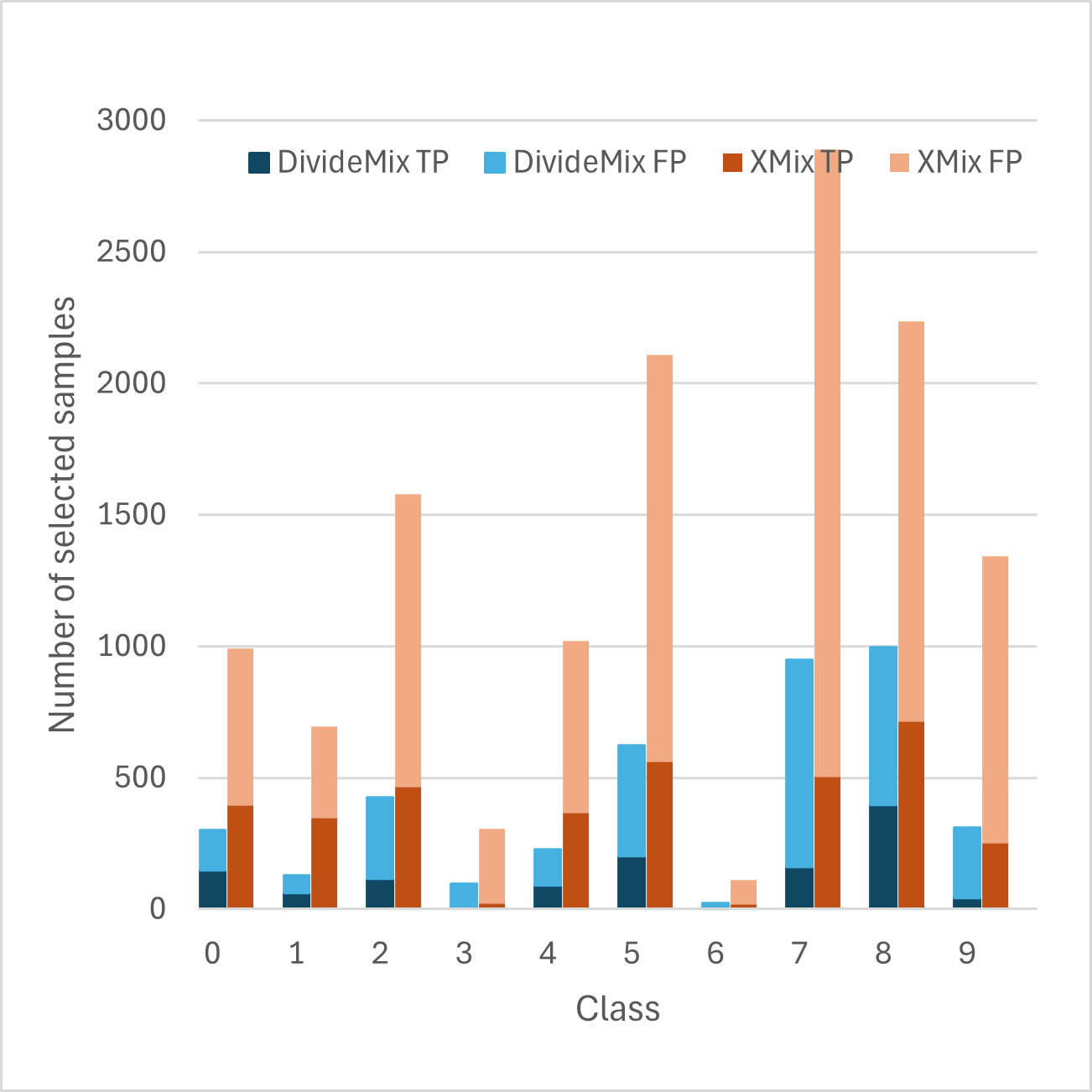}}
    {\includegraphics[width=\linewidth]{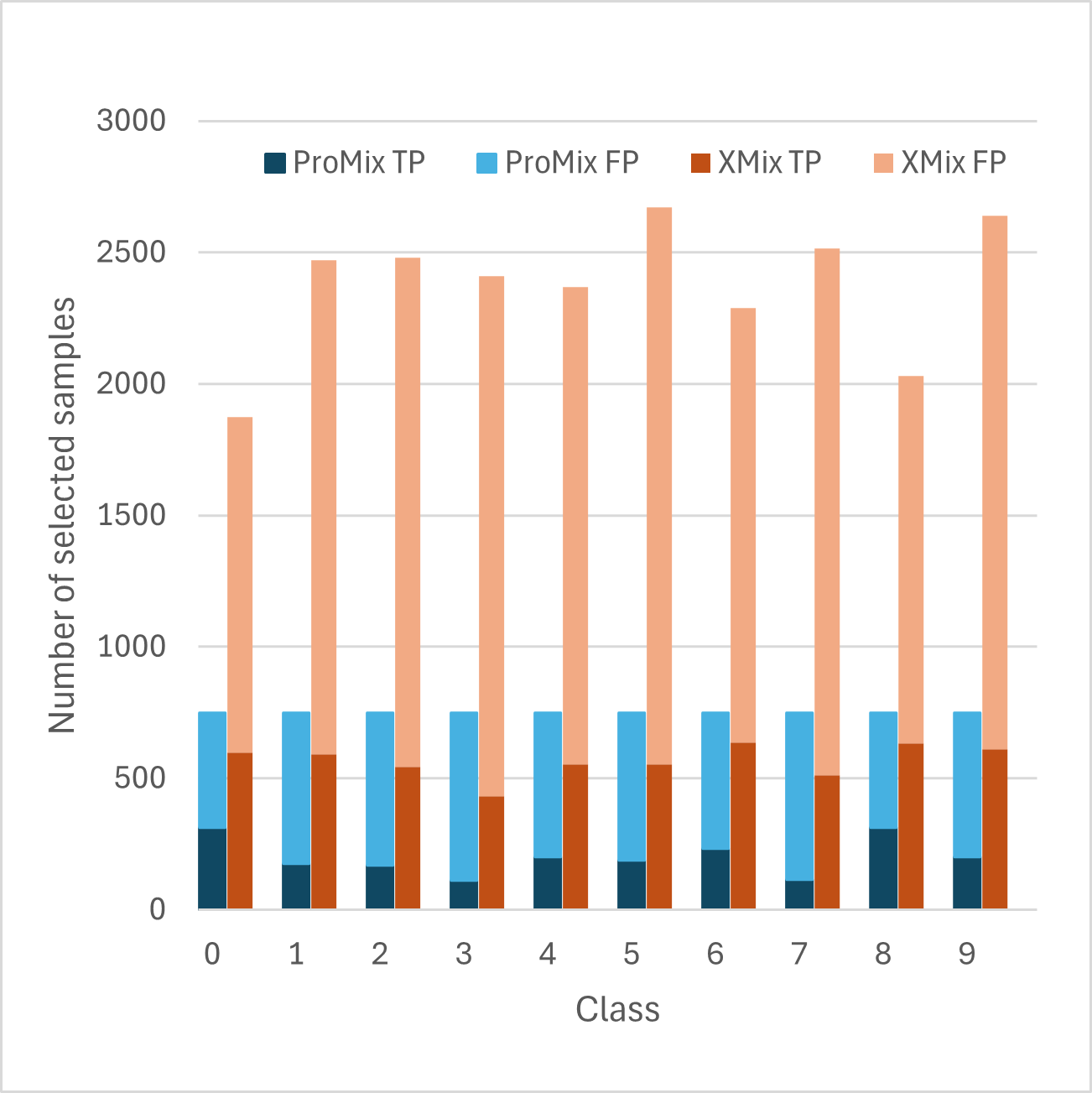}}
    \centerline{(f) 95\% Sym}
\end{minipage}
\begin{minipage}[htb]{0.24\linewidth}
    \centering
    {\includegraphics[width=\linewidth]{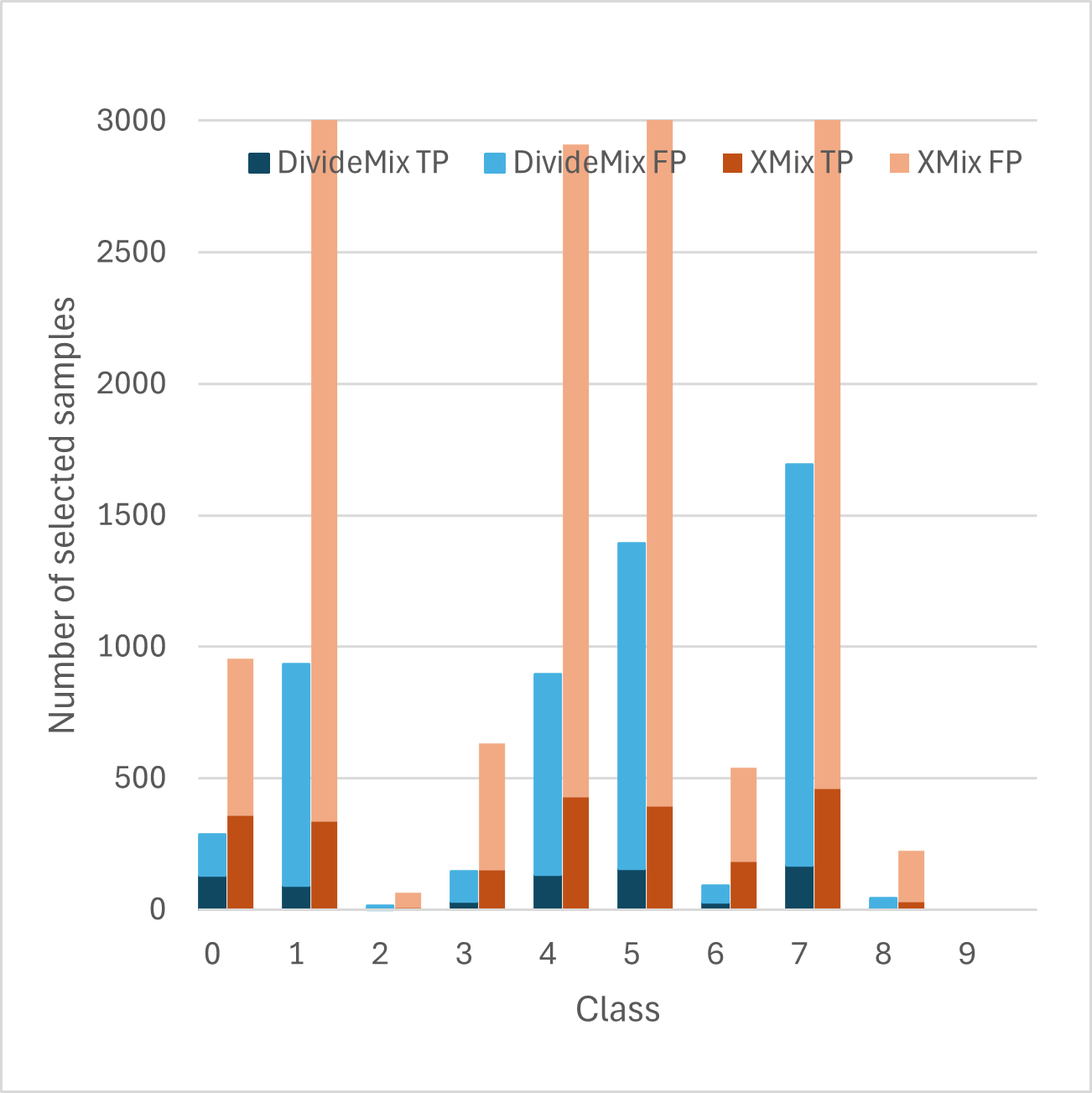}}
    {\includegraphics[width=\linewidth]{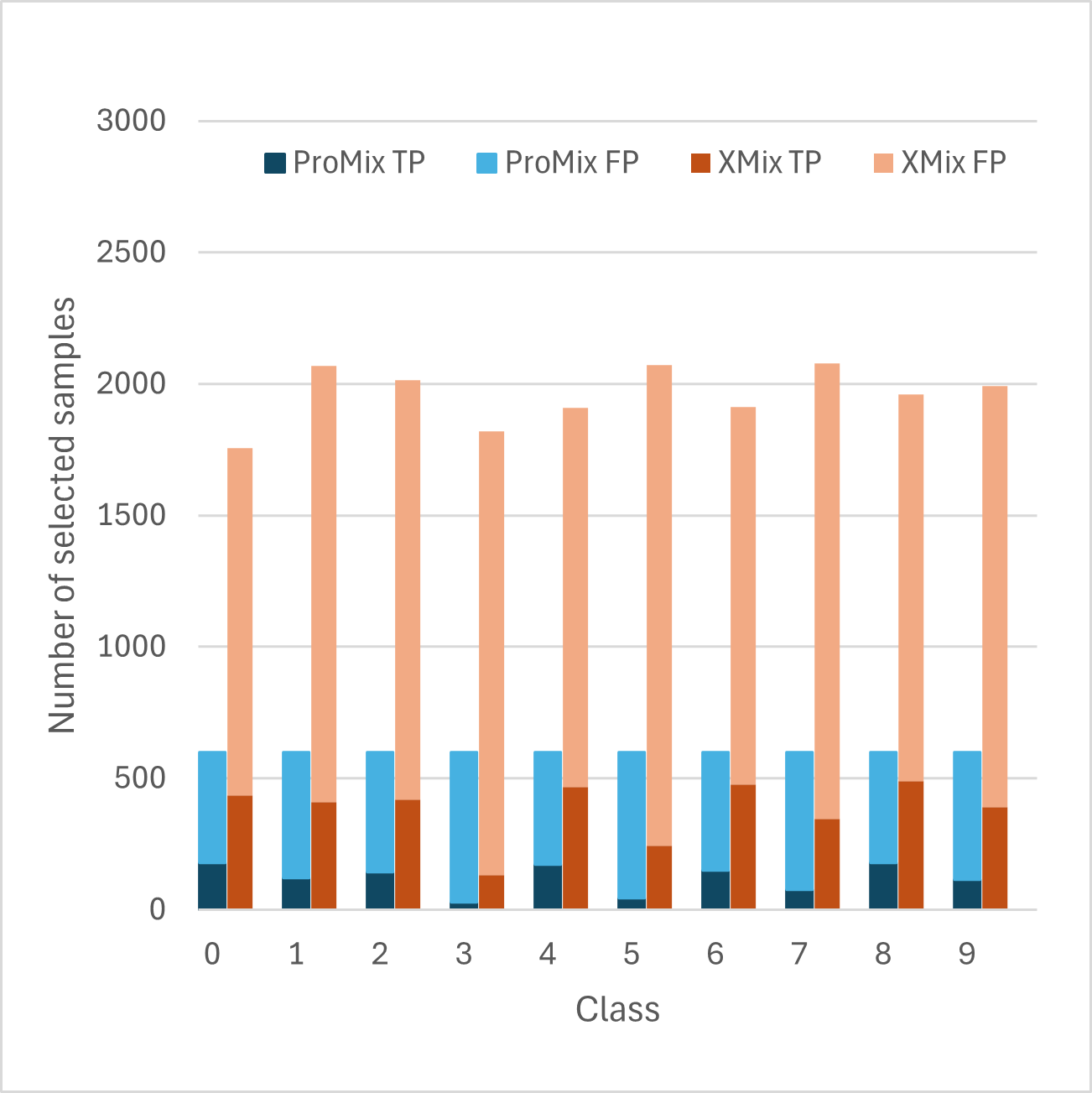}}
    \centerline{(g) 98\% Sym}
\end{minipage}
\caption{Continuation of Figure~\ref{fig:selected_samples_noise_full} at extreme symmetric noise levels. \textbf{Upper Row:} DivideMix vs.\ XMix; \textbf{Bottom Row:} ProMix vs.\ XMix. TP/FP denotes true and false clean samples, shown in light and dark shades, respectively. Zoom in for details.}
\label{fig:selected_samples_noise_full_2}
\end{figure*}

\FloatBarrier

\section{Connection to Label Differential Privacy}
\label{app:ldp}
Beyond standard LNL benchmarks, the symmetric noise setting studied in the main text also connects to \emph{Label Differential Privacy}, where labels are deliberately and heavily corrupted to protect privacy. Building on the success of Differential Privacy (DP) \cite{dwork_algorithmic_2013}, Label Differential Privacy \cite{chaudhuri_sample_2011} is a refined definition that adds calibrated noise to labels only, narrowing the protection scope and improving utility relative to feature-level DP. We include this connection here because it explains why an LNL method robust to \emph{extreme} symmetric noise, the regime XMix targets, is directly useful for learning from privately labeled data.

\noindent \textbf{Label Differential Privacy.} The definition of Label Differential Privacy mirrors standard Differential Privacy \cite{dwork_algorithmic_2013} except for the notion of adjacency: two adjacent datasets differ in only one record's label and must be statistically indistinguishable when processed through the same randomized mechanism. Formally, let $\varepsilon, \delta \in \mathbb{R}^+$. A randomized mechanism $\mathcal{M}$ that maps a dataset to a privatized dataset is $(\varepsilon, \delta)$-Label Differentially Private if, for all possible outputs $\mathcal{S}$ and any two neighboring datasets $\mathbb{D}=\{(\vecbold{x}_1, \vecbold{y}_1), \dots, (\vecbold{x}_i, \vecbold{y}_i), \dots, (\vecbold{x}_N, \vecbold{y}_N)\}$ and $\mathbb{D}'=\{(\vecbold{x}_1, \vecbold{y}_1), \dots, (\vecbold{x}_i, \vecbold{y}_i^{\prime}), \dots, (\vecbold{x}_N, \vecbold{y}_N)\}$ that differ only in a single label, $\vecbold{y}_i \neq \vecbold{y}_i^{\prime}$ for arbitrary $i$,
\begin{equation}
\operatorname{Pr}[\mathcal{M}(\mathbb{D}) \in \mathcal{S}] \leq \exp (\varepsilon) \operatorname{Pr}[\mathcal{M}(\mathbb{D}^{\prime}) \in \mathcal{S}] + \delta,
\end{equation}
where $\varepsilon$ is the privacy parameter and $\delta$ is the probability of privacy breach; when $\delta=0$, the mechanism achieves $\varepsilon$-Label Differential Privacy \cite{ghazi_deep_2021}.

\noindent \textbf{$k$-Randomized Response.} A common mechanism for generating private labels that satisfy $\varepsilon$-Label Differential Privacy is the $k$-Randomized Response mechanism \cite{kairouz_discrete_2016}, which extends the classic Randomized Response mechanism \cite{dwork_algorithmic_2013} to $k$-ary categorical data such as classification labels. Let the randomized response $\mathcal{M}_{KRR}$ be a mechanism over a categorical response $[k] \rightarrow [k]$:
\begin{equation}
\mathcal{M}_{KRR}(i) =
\begin{cases}
i, & \text{w.p. } \frac{\exp (\varepsilon)}{\exp (\varepsilon)+k-1}\\
j, & \text{w.p. } \frac{1}{\exp (\varepsilon)+k-1}
\end{cases}
\end{equation}
where $j \in [k] \setminus \{i\}$. This mechanism returns the true label with a probability that grows with $\varepsilon$, and otherwise distributes an incorrect label uniformly over the remaining $k-1$ categories. Its privacy guarantee follows directly from the ratio of the highest to lowest output probabilities,
\begin{equation}
\frac{\operatorname{Pr}[\mathcal{M}(\mathbb{D}) \in \mathcal{S}]}{\operatorname{Pr}[\mathcal{M}(\mathbb{D}^{\prime}) \in \mathcal{S}]} \leq \frac{\frac{\exp (\varepsilon)}{\exp (\varepsilon)+k-1}}{\frac{1}{\exp (\varepsilon)+k-1}} = \exp (\varepsilon),
\end{equation}
confirming that $\mathcal{M}_{KRR}$ is $\varepsilon$-Label Differentially Private.

\noindent \textbf{Relation to symmetric label noise.} Comparing the $k$-Randomized Response mechanism with the symmetric noise model in Section \ref{section: background} reveals a direct correspondence: learning from labels that satisfy $\varepsilon$-Label Differential Privacy is equivalent to learning under symmetric label noise at level $\eta = \frac{C}{\exp (\varepsilon) + C - 1}$. Lower values of $\varepsilon$ indicate stronger privacy, with $\varepsilon=0.5$ being a common setting for deep learning models; for a ten-class private classification task, this corresponds to an equivalent symmetric noise level of $94\%$, squarely within the extreme-noise regime XMix is designed for. Table \ref{tab:relation_ldp_sym} lists common privacy budgets used in practice together with their equivalent noise levels and expected fraction of correctly labeled samples.

\begin{table}[h]
\centering
\caption{Relation between Label Differential Privacy and symmetric label noise.}
\begin{tabular}{llll}
\toprule
\multicolumn{1}{l}{Dataset} & $\varepsilon$ & $\eta$ & Expected clean labels \\
\midrule
\multirow{4}{*}{CIFAR-10}  & 0.5 & 93.9\% & 15.5\% \\
                           & 1   & 85.3\% & 23.2\% \\
                           & 2   & 61.0\% & 45.1\% \\
                           & 4   & 15.7\% & 85.8\% \\
\midrule
\multirow{4}{*}{CIFAR-100} & 1   & 98.3\% & 2.7\%  \\
                           & 2   & 94.0\% & 6.9\%  \\
                           & 4   & 65.1\% & 35.5\% \\
                           & 6   & 19.9\% & 80.3\% \\
\bottomrule
\end{tabular}
\label{tab:relation_ldp_sym}
\end{table}

Because Label Differential Privacy at practical privacy budgets induces exactly the extreme symmetric noise regime that motivates this work, XMix's balanced-and-expanded selection and nearest-neighbor pseudo-labeling are directly applicable to learning from privately labeled data, without any change to the method itself.

\FloatBarrier

\section{Earlier Self-Supervised Clustering Variant}
\label{app:earlier-variant}
The neighbor-based formulation in Section \ref{section: method} evolved from an earlier version of XMix that used global representation clustering rather than local feature-space neighbors. In that earlier design, a self-supervised encoder was trained with contrastive learning (SimCLR \cite{chen_simple_2020}) directly on the target dataset, and $K$-means clustering was applied to the resulting representations to partition the training data into $K$ clusters. Samples sharing the same observed label as a low-loss (clean) sample within the same cluster were then added to the clean subset, in place of the $k$-nearest-neighbor criterion used in the main text. This clustering-based variant established the core idea that self-supervised, label-agnostic structure in feature space can recover additional clean samples beyond the low-loss criterion, and it is the basis of the ablation in Table~\ref{tab:ablation} that replaces the DINOv2 encoder with a SimCLR encoder. However, fixed clusters computed once over the whole dataset are coarser than per-sample neighborhoods, do not naturally support the per-class rate estimation described in Section \ref{section: method}, and are less able to adapt to local variations in class boundaries. The $k$-nearest-neighbor formulation used throughout this paper addresses these limitations while preserving the same underlying smoothness assumption.

\end{document}